\documentclass[journal]{IEEEtran} % use the `journal` option for ITherm conference style
\pdfoutput=1
\IEEEoverridecommandlockouts
\usepackage{cite}

\usepackage{amsmath,amssymb,amsfonts}
\usepackage{algorithm}
\usepackage{algpseudocode}
\usepackage{graphicx}
\usepackage{textcomp}
\usepackage{xcolor}
\usepackage{tabularx}
\usepackage{tablefootnote}
\usepackage{float}
\usepackage{subfigure}
\usepackage{multicol}
\usepackage{listings}
\usepackage{mdframed}
\usepackage{pdfpages}
\usepackage{caption}
\usepackage{titlesec} % For customizing section titles
\usepackage{graphicx} % For including images
\usepackage{multirow} 
\usepackage{hyperref}
\usepackage{multirow} 
\usepackage{boldline}
\usepackage{longtable,booktabs,array}
\newcommand{\subsectionimage}[3][width=3.5in]{%
  \subsection*{\centering#2}
  \includegraphics[#1]{#3}
  \addcontentsline{toc}{subsection}{#2}
}

\def\BibTeX{{\rm B\kern-.05em{\sc i\kern-.025em b}\kern-.08em
    T\kern-.1667em\lower.7ex\hbox{E}\kern-.125emX}}
\begin{document}

\title{ROSGPT\_Vision: Commanding Robots Using Only Language Models' Prompts %Making Robots Understand What They See

% delete or comment-out the following line before submission
\thanks{}
}

\author{%%%% author names
    \IEEEauthorblockN{Bilel Benjdira, Anis Koubaa, Anas M. Ali}% first author

    %%%% author affiliations
    \IEEEauthorblockA{\textit{Robotics and Internet of Things Lab (RIOTU Lab), Prince Sultan University, Saudi Arabia}}
    %\IEEEauthorblockA{\textit{Research Centre in Real-Time and Embedded Computing Systems, Polytechnic of Porto, Porto, Portugal}}% first affiliation
    %\IEEEauthorblockA{\textit{dept. name of organization (of Aff.), City, Country if needed}}\\% delete this line if not needed
    % duplicate the line above as many times as needed to list all affiliations
    %%%% corresponding author contact details
    
    \IEEEauthorblockA{bbenjdira@psu.edu.sa; akoubaa@psu.edu.sa; aaboessa@psu.edu.sa}
}

\maketitle

\begin{abstract}
In this paper, we argue that the next generation of robots can be commanded using only Language Models' prompts. Every prompt interrogates separately a specific Robotic Modality via its Modality Language Model (MLM). A central Task Modality mediates the whole communication to execute the robotic mission via a Large Language Model (LLM). This paper gives this new robotic design pattern the name of: Prompting Robotic Modalities (PRM). Moreover, this paper applies this PRM design pattern in building a new robotic framework named ROSGPT\_Vision. ROSGPT\_Vision allows the execution of a robotic task using only two prompts: a Visual and an LLM prompt. The Visual Prompt extracts, in natural language, the visual semantic features related to the task under consideration (Visual Robotic Modality). Meanwhile, the LLM Prompt regulates the robotic reaction to the visual description (Task Modality). The framework automates all the mechanisms behind these two prompts. The framework enables the robot to address complex real-world scenarios by processing visual data, making informed decisions, and carrying out actions automatically. The framework comprises one generic vision module and two independent ROS nodes. As a test application, we used ROSGPT\_Vision to develop CarMate, which monitors the driver's distraction on the roads and makes real-time vocal notifications to the driver. We showed how ROSGPT\_Vision significantly reduced the development cost compared to traditional methods. We demonstrated how to improve the quality of the application by optimizing the prompting strategies, without delving into technical details. ROSGPT\_Vision is shared with the community \footnote{link: \href{https://github.com/bilel-bj/ROSGPT\_Vision}{https://github.com/bilel-bj/ROSGPT\_Vision}} to advance robotic research in this direction and to build more robotic frameworks that implement the PRM design pattern and enables controlling robots using only prompts.

%This framework helps to enhance the capabilities of robotic systems in addressing intricate real-world scenarios. This pioneering approach expands upon ROSGPT by accommodating natural human commands, necessitating the robot to execute computer vision tasks, process visual data, make informed decisions, and carry out actions. To substantiate the efficacy of our approach, we employ cutting-edge computer vision models, 

%specifically Segment-Anything and Caption-Anything, which facilitate the extraction of unstructured textual information from visual data. Subsequently, we utilize ChatGPT, an advanced language model, to transmute the extracted textual information into structured robotic commands that the robot can efficiently interpret and execute.
%Using prompt engineering and zero/few-shot learning techniques ensures the seamless conversion of textual information into robotic commands. 

%Furthermore, we classify the robot's mission into three distinct tasks: (1) Action - encompassing the actuation and control of the robot's movements and activities, and (2) Communication - involving context-specific interactions with third parties. This comprehensive categorization enables a clear understanding of the robot's core functionalities, thereby providing a solid foundation for future advancements in the field of robotics. 
\end{abstract}

%We apply this concept to Robot Operating System (ROS) robotic framework, which is widely recognized as the standard and versatile development framework for robotic systems. 
\begin{IEEEkeywords}
    Robotic Design Patterns, Robotic Modalities, Prompting Robotic Modalities, Large Language Models, LLMs, Vision Language Models, VLMs, Robotic Operating System, ROS, ROS2, Robotic Prompt Engineering, Visual prompt, LLM prompt
\end{IEEEkeywords}

\section{Introduction}
Recent advances in Large Language Models (LLMs), Robotics, and Computer Vision have developed new research trends to bridge the gap between visual and sensor data interpretation, language modeling, and robotic actions. This research is inspired by ROSGPT \cite{koubaa2023rosgpt}, which built the first framework to convert human language into explicit robotic commands using the ChatGPT LLM, helping to interact with robotics in natural language. However, ROSGPT \cite{koubaa2023rosgpt}, did not consider multimode interaction patterns and thus limited to the textual modality. The current study enlarged the scope by suggesting a new robotic design pattern and a new robotic framework. We proposed the Prompting Robotic Modalities (PRM) design pattern to conceive a common ground for future developments in this field. Also, we proposed the ROSGPT\_Vision framework based on PRM to make seamless robotic interaction with visual data by the intermediate of Language Models. In the following subsections, a closer look has been made at the state of the art of Language Models, robotics, and visual data interpretation. This helps to deduce the current gaps in the literature targeted by this work. 

\subsection{ Large Language Models (LLMs) and their impact on Robotics}
In recent years, the development of large language models (LLMs) has revolutionized the field of natural language processing (NLP) with impressive capabilities, such as generating human-like text, understanding context, answering questions, and even translating languages \cite{GPT3-paper, radford2018improving}. %These advancements in LLMs have led to a growing interest in exploring their potential impact on robotics.
LLMs has brought about significant advancements with a wide range of applications spanning multiple domains \cite{zhao2023survey}. Notable examples include OpenAI's GPT-3 \cite{GPT3-paper} and ChatGPT \cite{koubaa2023exploring}, which have demonstrated remarkable language processing and understanding capabilities. These developments have created new opportunities for enhancing human-robot interaction and communication, as well as other aspects of robotics.

The research community has shown a growing interest in harnessing the potential of LLMs to improve various aspects of robotics. For instance, LLMs have been used to facilitate natural language understanding in robots, enabling them to process and respond to human commands more effectively \cite{thomason2019vision}. Additionally, LLMs have been employed in human-robot collaboration, where robots can generate context-aware responses to human instructions, promoting seamless cooperation \cite{singh2019towards}.

LLMs have been proven to be highly useful in several applications due to their remarkable ability to learn new communication patterns with either \textit{zero-shot} or \textit{few-shot learning}. In zero-shot learning, the LLM can generate accurate responses for tasks it has never been trained on, while in few-shot learning, it can effectively adapt to new tasks with only a few training examples. This adaptability is a key advantage of LLMs, allowing them to learn quickly and improve in various contexts.

The exceptional adaptability of LLMs stems from prompt engineering techniques, which equip them to tackle intricate natural language processing and comprehension tasks. By supplying LLMs with precise prompts or instructions, they can produce highly accurate and pertinent responses to input text. This versatility has rendered LLMs invaluable across diverse applications, encompassing language translation, text summarization, chatbots, and human-robot interaction.
Moreover, the potential of LLMs in the area of explainable AI (XAI) for robotics has been explored, with LLMs generating human-readable explanations for robotic decision-making processes, enhancing trust and understanding between humans and machines \cite{chakraborti2020explainable}. LLMs have also been employed to improve robotic learning from demonstrations, where they generate natural language descriptions of expert demonstrations to teach robots new tasks through imitation learning \cite{waytowich2019leveraging}.

As the capabilities of LLMs continue to advance, it is anticipated that their integration with robotic systems will lead to increasingly sophisticated and capable machines that can interact seamlessly with humans and their environment. The exploration of LLMs in robotics is still in its early stages, and it is expected that future research will uncover new applications and techniques for leveraging the power of LLMs in the field of robotics.

Large language models (LLMs) possess remarkable capabilities to comprehend human language inputs and generate contextually relevant responses across diverse applications. Their inherent versatility allows them to be fine-tuned for specialized tasks, such as language translation or text summarization. Renowned examples of LLMs include OpenAI's GPT-3 \cite{GPT3-paper} and GPT-4 \cite{gpt4, koubaa2023gpt}, Google's BERT \cite{devlin2018bert}, and T5 \cite{raffel2019exploring}. These models exhibit exceptional abilities that set them apart from smaller pre-trained language models (PLMs).

Despite the remarkable performance demonstrated by LLMs in tackling intricate tasks, their underlying capabilities remain an active area of research. The scientific community continues to explore and investigate the full extent of LLMs' intrinsic properties to gain a deeper understanding of their potential \cite{zhao2023survey}.

A key challenge in robotics is developing systems that can effectively communicate and collaborate with humans. To address this challenge, researchers have begun to integrate LLMs into robotic systems to facilitate natural language understanding and generation. For instance, \cite{thomason2019vision} demonstrated the use of LLMs to enable a robot to understand and execute natural language commands for tasks such as object manipulation and navigation. Similarly, \cite{singh2019towards} proposed a method to combine LLMs with robotics to improve human-robot collaboration by allowing robots to generate context-aware responses to human instructions.

%Another area where LLMs have been applied to robotics is in the development of explainable AI (XAI) systems. The ability of LLMs to generate human-like text has been leveraged to create explanations for the decision-making processes of robotic systems \cite{chakraborti2020explainable}. By providing interpretable justifications for their actions, robots can enhance trust and understanding between humans and machines.

Furthermore, LLMs have been used to improve robotic learning from demonstrations. In \cite{waytowich2019leveraging}, the authors employed LLMs to generate natural language descriptions of expert demonstrations, which were then used to teach robots new tasks through imitation learning. This approach enables robots to acquire new skills more efficiently by leveraging the vast knowledge encoded in LLMs.

In summary, the impact of LLMs on robotics is multifaceted, with potential applications in human-robot communication, explainable AI, and robotic learning from demonstrations. As LLMs continue to advance, it is expected that their integration with robotic systems will lead to increasingly sophisticated and capable machines that can seamlessly interact with humans and their environment.

\subsection{Image Understanding for Robotic Manipulation}

Image understanding plays a crucial role in advancing the field of robotic manipulation, as it allows robots to perceive and interact with their surrounding environment more effectively \cite{levine2018learning, zeng2017robotic}. The integration of computer vision and machine learning techniques has greatly improved the performance of robotic systems, enabling them to autonomously recognize, grasp, and manipulate objects in various scenarios \cite{saxena2008robotic, mahler2017dex}.

Deep learning, in particular, has been instrumental in driving progress in this domain \cite{sunderhauf2018limits, karoly2020deep}. Convolutional Neural Networks (CNNs) and other advanced neural network architectures have demonstrated remarkable success in learning to detect and localize objects, predict grasps, and plan manipulations from raw image data \cite{lenz2015deep, redmon2016you}. These advances have facilitated more natural and efficient human-robot interaction, as well as enabled robots to handle complex tasks in diverse environments \cite{liu2020deep, shao2019multimodal}.

Despite the significant progress made in recent years, numerous challenges still exist in the field of image understanding for robotic manipulation. These include the need for robust object recognition under varying conditions, generalization across different object categories and environments, and real-time processing for fast and adaptive robotic control \cite{bohg2017interactive, pinto2016supersizing}. Addressing these challenges will pave the way for more sophisticated robotic systems capable of performing a wide range of tasks autonomously and in cooperation with humans.

\subsection{Recent Advances in Vision-Language Models}

Vision-language models (VLMs) are a type of artificial intelligence (AI) that can comprehend and generate both visual and textual information \cite{lu2019vilbert, li2020oscar}. Trained on enormous datasets of images and text, they have been utilized for a range of tasks such as image captioning, visual question answering, and visual reasoning \cite{li2022visionlanguage, zhang2023vision, anderson2018bottom, tan2019lxmert}.

With their potential to transform the way we interact with computers, there has been an upsurge of interest in VLMs in recent years \cite{su2020vl}. VLMs can create more intuitive and engaging user interfaces and automate tasks that are currently performed by humans, such as image captioning and visual question answering.

Some of the notable advances in VLMs include:

\begin{itemize}
    \item \textbf{Larger and more complex models:} VLMs have grown increasingly large and complex, due to the availability of larger and more diverse datasets and the development of more powerful LLMs \cite{GPT3-paper}. For instance, the GPT-3 language model has 175 billion parameters, while the ViLBERT vision-language model boasts 340 million parameters \cite{lu2019vilbert}.

    \item \textbf{Multimodal learning:} The capability of VLMs to learn from both visual and textual data has been realized using multimodal neural networks that can merge information from both modalities \cite{zhang2023vision}.

    \item \textbf{Transfer learning:} Knowledge gained from Efficient Vision encoders is used to enhance the performance of the VLM models, which can be considered as a form of multimodal transfer learning \cite{zhang2023vision}. For instance, the MiniGPT-4 model \cite{zhu2023minigpt} used the efficient Q-former \cite{zhang2023visionQformer} architecture as an image feature extractor in their bi-modal network. 
\end{itemize}

These advances have contributed to significant improvements in the performance of VLMs on various tasks. For example, the MiniGPT-4 \cite{zhu2023minigpt} and LLava \cite{liu2023llava} have demonstrated satisfying capabilities in reasoning about image data in natural human language.

The future of VLMs appears very promising. As they continue to grow in size and complexity, VLMs will be capable of performing increasingly complex tasks. This progress is expected to facilitate the development of innovative applications such as virtual assistants that understand and respond in natural language and robots that can perceive and understand their environment \cite{zhang2023vision}.

VLMs, however, do face a few challenges:

\begin{itemize}
    \item \textbf{Data scarcity:}One major challenge for VLMs is the lack of data. For effective learning from both visual and textual data, VLMs require large and diverse datasets, which are often costly and time-consuming to collect \cite{long2022visionandlanguage}.

    \item \textbf{Bias:} Since VLMs are trained on data created by humans, they can reflect human bias in their outputs, leading to inaccurate or unfair results \cite{mehrabi2021survey}.

    \item \textbf{Interpretability:} Due to their complex training datasets and multifactorial decision-making processes, VLMs as well as many other deep learning models are often hard to interpret \cite{ribeiro2016should}. This can make it difficult to understand why a VLM makes a specific prediction or decision.
\end{itemize}

Despite these challenges, VLMs are a promising new technology with the potential to revolutionize our interaction with computers. As improvements continue, VLMs are expected to perform an increasingly complex array of tasks, leading to the development of new and innovative applications.

\subsection{Novelty of IRM and ROSGPT\_Vision}
As shown above, the robotic community needs to profit from the  development of LLMs and VLMs to design new methods that facilitate the execution of robotic tasks. This paper introduces a new Robotic Design Pattern: the PRM (Prompting Robotic Modalities). It presents a novel approach to robotic design, emphasizing the distinct querying of individual sensory and interaction channels. This innovative pattern suggests that future robots should be equipped to address each modality, such as vision or audition, through specific "prompts." These prompts, tailored commands or inquiries, allow for precise data gathering from each sensory input, ensuring that every modality operates as an independent data source. Every Modality should be associated with its own Modality Language Model (MLM). The MLM helps to inquire the data in a promptable manner. Form example, Vision-Language Model (VLM) are the MLM specific the Visual Modality. It used to be used to prompt visual data using textual input. However, to synthesize this data for cohesive robot action, the central "Task Modality" acts as a mediator. It collates information from each prompt, guiding the robot's overarching decisions and actions. The  robots actively probe each modality rather than passively receiving data. the Task Modality is connected to aLarge Language Model that should have advanced eliciting capability. In essence, the PRM design opt for a modular approach, offering robots greater flexibility and precision. By treating each sensory channel as a separate module and having a central system for decision-making, robots can adapt better to their environment, paving the way for more responsive and adaptable robotic systems in the future. 

To prove the validity of this new design pattern, we apply it for the task that needs only Vision Modality such as tasks that need to take informative decision about visual content detected by the Robotic camera. We applied the PRM Design pattern to develop the ROSGPT\_Vision framework. The input image given to the robot is converted into natural textual language with enhanced reasoning and eliciting features by the intermediate of the interrogating visual prompt. By doing so, we aim to bridge the gap between visual perception and natural language understanding in the context of human-robot interaction. This approach enables robots not only to process and analyze visual information but also to understand and act upon human language-based commands and instructions. Furthermore, the incorporation of enhanced reasoning and eliciting features in the generated textual language allows robots to interpret complex instructions better, ask relevant questions for clarification, and provide more informative feedback to humans. Ultimately, this method has the potential to facilitate more intuitive, efficient, and effective communication between humans and robots, leading to improved collaboration and performance in various tasks and environments.
In the rapidly evolving field of robotics, the integration of language models with computer vision capabilities has become a significant area of interest. This paper explain the PRM design pattern and introduces a new robotics framework that helps the robots interact with images in natural language: the ROSGPT\_Vision. This framework provides an intersection of language understanding and image perception to extend the capabilities of robots, enabling them not only to perceive their environment visually but also to interpret and interact with it in a meaningful way. ROSGPT\_Vision is an introductory step forward in the quest for sophisticated, autonomous, and intelligent robotic systems. The utility of the PRM pattern and the ROSGPT\_Vision framework is vast, with potential applications ranging from domestic helper robots to advanced manufacturing systems, autonomous vehicles, and more.

The name ROSGPT\_Vision is given to reflect the inherent integration of LLMs (Mostly based on GPT architecture) with Robot Operating System ROS, and Vision Language Models (VLMs). In the context of this study, we use the abbreviation ROS to refer to ROS 1 and ROS 2. With ROSGPT\_Vision, the robot can translate image data into natural human language commands using task-designed visual prompts. The textual output will be converted into LLMs to acquire more reasoning capabilities to deduce the best action to take following the context of the robot task. Both this work and ROSGPT \cite{koubaa2023rosgpt}  target to improve the state of the art in the robotic field by making seamless interpretation of sensor data in natural human language. ROSGPT \cite{koubaa2023rosgpt} makes a broker that converts human commands given in natural language to explicit robotic commands, while ROSGPT\_Vision makes a broker to convert image data into natural human language to benefit from the power of LLMs and to convert it into more customizable robot commands. 

The most near research works to our study are \cite{PromptCraft} and \cite{He2023}. In \cite{PromptCraft}, the authors used ChatGPT to enable natural language-based control of many robotic platforms such as drones, robot arms, and home assistant robots. They made design principles to help language models solve robotic tasks. They introduced an AirSim simulation environment integrated with ChatGPT. They also developed PromptCraft, a collaborative open-source platform that collects some prompting strategies for different robotics tasks. In \cite{He2023}, the author introduced robotGPT, by presenting ChatGPT's principles and discussing how to enhance robotic intelligence through the medium of ChatGPT.

Distinct from the approaches presented above, ROSGPT\_Vision introduces various innovative elements that significantly advance the field of robot interaction with visual data. Firstly, ROSGPT\_Vision divides the robot interaction with visual data into many modules explained in the next section. Second, ROSGPT\_Vision makes a distinction between the extraction of image semantics and the robot ontology associated with one given task. Third, it completely separates the prompting logic from the whole architecture, so that the end user will focus mostly on developing the most accurate prompts for his task in the YAML files associated with it. Finally, the shared ROS package is designed following a clear and extensible architecture that could be used and understood by normal users.
%Up to our knowledge, ROSGPT\_Vision is the first research work that incorporated LLMs with Image data analysis together inside the ROS system. The framework of this integration is described and detailed in this study. The ROS package code is shared with the community.
PRM and ROSGPT\_Vision tries to bridge the gap between Robotics, Natural Language Processing, and Computer Vision by orienting the next Robotic developers towards focusing on designing the best prompts for their tasks and to explicitly improving the clarity of the robot ontology behind these prompts. This work represents a stepping stone for the ROS, NLP, and Computer Vision communities to collaboratively advance this interdisciplinary research field.

\section{prompting Robotic Modalities (PRM) Design Pattern}
\begin{figure}[htbp]
    \centering
    \includegraphics[width=0.9\linewidth]{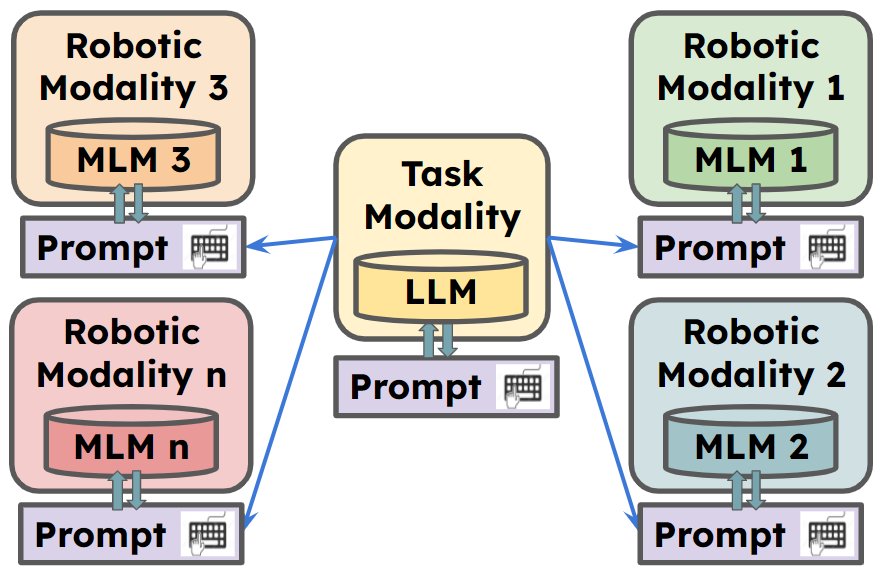}
    \caption{The robotic design pattern "Prompting Robotic Modalities (PRM)"}
    \label{fig_prm}
\end{figure}

The concept of "Prompting Robotic Modalities (PRM)" introduces a novel approach to the domain of robotic design, emphasizing modularity and targeted querying of individual sensory and interaction channels. The global idea is show in Figure \ref{fig_prm}. At the heart of this design pattern is the idea that robots, should be constructed with the capability to prompt each of their sensory or interaction modalities—be it vision, audition, or tactile sensing—separately and distinctly. This is done by incorporating every robotic modality inside a specific Modality Language Model (MLM). The MLM will be trained to interrogating the Modality data in textual manner giving more insights on it. As an example, the Visual Robotic Modality is incorporated in ROSGPT\_Vision (see next section) inside a VLM (Vision Language Model). The VLM helps to enquire the images captured by the robotic camera using textual inputs. 

The "prompt" in this context can be understood as a specific command or inquiry aimed at a particular modality. For instance, when the robot needs to gather visual data about its surroundings, a distinct prompt might be issued to its vision modality, asking it to identify objects within a certain range. Similarly, if the robot needs to ascertain ambient noise levels, a separate prompt might be directed at its auditory modality. The beauty of this approach lies in its precision; by addressing each modality with individualized prompts, the system ensures that every sensory input or output is treated as a unique and independent source of data, devoid of unnecessary cross-modality interference.

However, while the idea of querying each modality separately offers a high degree of precision, it also presents a challenge: how does one ensure that the data from these various modalities is synthesized effectively to guide the robot's overarching actions and decisions? This is where the concept of the "Task Modality" comes into play. Acting as the central  system of the robot, the Task Modality serves as a mediator or coordinator. Its primary role is to communicate with each of the robot's modalities, collate the information obtained from the distinct prompts, and then make informed decisions on the best course of action to ensure the successful execution of the robot's mission. Different from normal Robotic Modalities, The Task Modality is connected to a generic LLM not to an MLM. 

Instead of the robot passively receiving data from its various sensors, the PRM approach necessitates an active querying or probing of each modality. Such an active approach to data gathering might very well lead to the creation of robotic systems that are not only more responsive but also more adept at handling real-time challenges and changes in their environment.

In summary, the PRM design pattern is a testament to the potential benefits of a modular approach in robotic design. By treating each sensory or interaction channel as an independent promptable module, robots can achieve a higher degree of flexibility and precision in their operations. Moreover, the central Task Modality ensures that while each modality operates independently, there remains a cohesive system in place to guide the robot's overall behavior. Such a design could pave the way for the development of robots that can easily adapt to new tasks, integrate new modalities, and respond more efficiently to the demands of their environment. This paper applies the PRM design pattern to conceive the ROSGPT\_Vision framework, detailed in the next section.

\section{Conceptual Architecture of ROSGPT\_Vision}

\begin{figure*}[htbp]
    \centering
    \includegraphics[width=0.9\linewidth]{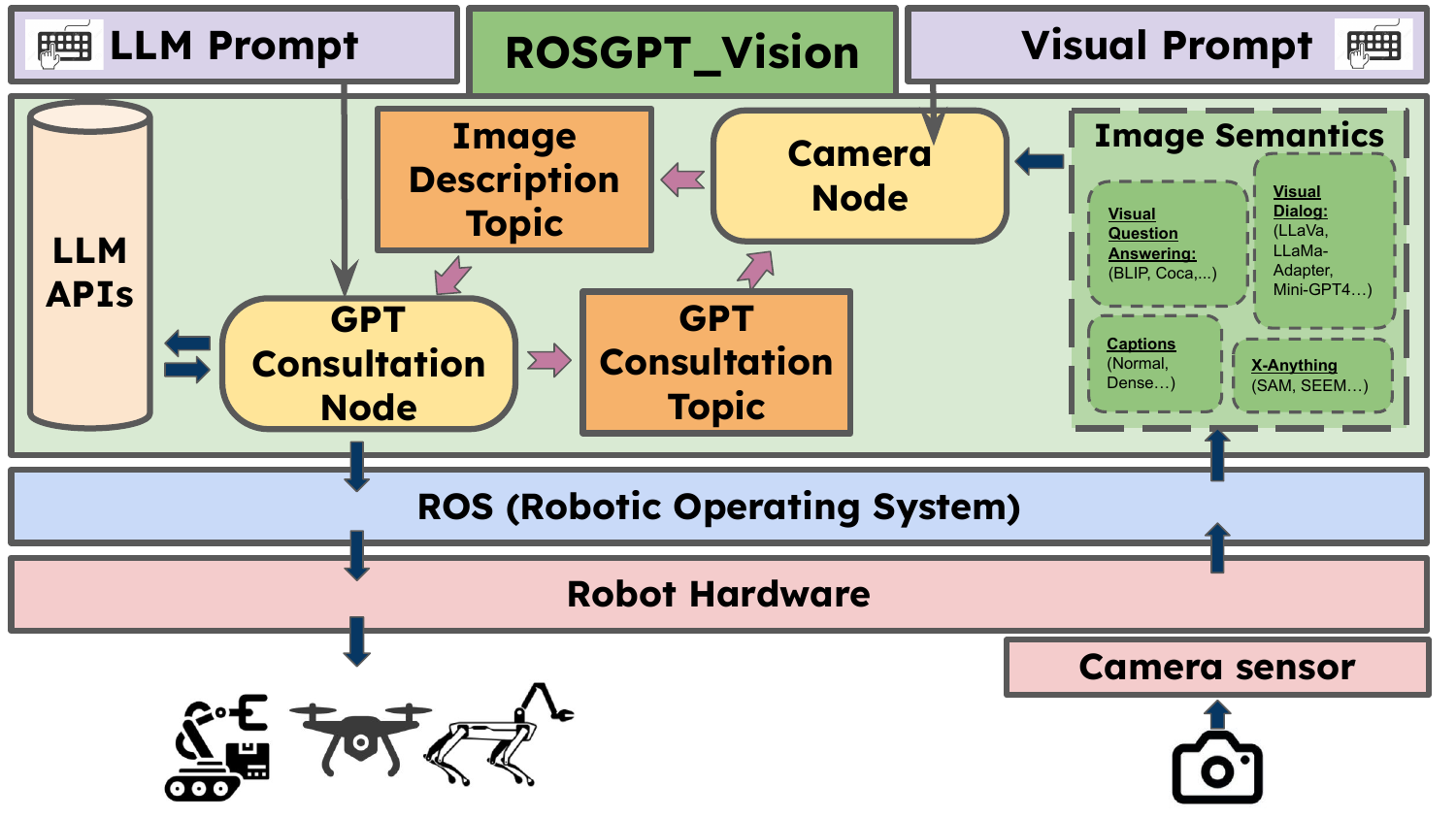}
    \caption{ROSGPT\_Vision Architecture for Natural Language Understanding of Visual data}
    \label{fig_rosgpt_vision_architecture_}
\end{figure*}

The ROSGPT\_Vision architecture is described in Figure \ref{fig_rosgpt_vision_architecture_}. Visual data is captured by the camera, transferred to the Image Semantics module to generate an accurate description guided by the Visual prompt given by the user. The Camera node always updates the Image Description topic with the given description of the image. This description is transferred to the GPT Consultation node, which interacts with the LLM to generate accurate decisions and instructions that should be made by the robot. The interaction with the LLM is also guided by an LLM prompt given by the user to optimize the robotic job. The prompting logic is separated from the whole architecture in order to let the end user focus on optimizing this critical part of the Robotic design. The framework is based on the PRM design pattern, where the GPT Consultation Node refers the Task Modality controlled via the LLM prompt. The Camera Node is the embodiment of the Visual Robotic Modality where the Visual Language Model (VLM) associated to this specific modality is included in the Image Semantics Module. We can see that the Camera Node helps to inspect the images captured from the robotic camera via a specific prompt (the Visual Prompt). In the next subsection, more details are given about every component of the framework. 

\subsection{\textbf{Image Semantics: Extracting meaning from Images}}
The Image Semantic module is the primary building block in ROSGPT\_Vision. It serves as a bridge between visual data and human language. This module's primary function is to extract and interpret meaning from images in a way that mirrors natural human language understanding. It uses different deep learning models to analyze the contents of an image in various ways. It acts by identifying objects, patterns, and relationships within the image, and then translates these visual elements into semantic descriptions in natural language. Moreover, it can use direct image captioning models, visual chatting models to achieve this goal. The output of any method contained in this module is to generate a descriptive narration for an image in a human-natural manner. This output will be guided for a specific robotic task by customized prompts.

It assembles many predefined methods for this topic, such as LLAVA model \cite{liu2023llava}, MiniGPT-4 \cite{zhu2023minigpt}, and SAM (Segment Anything) \cite{kirillov2023segment}:

\subsubsection{\textbf{LLaVA Model}}
The LLaVA (Large Language-and-Vision Assistant) model represents a novel end-to-end trained large multimodal model that combines a vision encoder and Vicuna for general-purpose visual and language understanding \cite{liu2023llava}. LLaVA connects a pre-trained CLIP ViT-L/14 visual encoder and a large language model, Vicuna, using a simple projection matrix. It employs a two-stage instruction-tuning procedure:
\begin{enumerate}
\item Pre-training for Feature Alignment: In this stage, only the projection matrix is updated, based on a subset of CC3M.
\item Fine-tuning End-to-End: In this stage, both the projection matrix and the large language model are updated. Two different use scenarios are considered:
    \begin{itemize}
    \item Visual Chat: LLaVA is fine-tuned on generated multimodal instruction-following data for daily user-oriented applications.
    \item Science QA: LLaVA is fine-tuned on a multimodal reasoning dataset for the science domain.
    \end{itemize}
\end{enumerate}
LLaVA has demonstrated impressive multimodal chat abilities and achieved a new state-of-the-art accuracy of 92.53\% when fine-tuned on Science QA, thus setting a benchmark in the domain of visual and language understanding \cite{liu2023llava}.

\subsubsection{\textbf{MiniGPT-4 Model}}

The MiniGPT-4 \cite{zhu2023minigpt} model, similarly to LLAVA, is an advancement in the field of vision-language understanding that leverages an advanced large language model (LLM), Vicuna, and demonstrates multi-modal abilities similar to the more complex GPT-4. MiniGPT-4 consists of a vision encoder with a pre-trained ViT and Q-Former, a single linear projection layer, and an advanced Vicuna large language model. The model only requires training the linear layer to align the visual features with the Vicuna \cite{zhu2023minigpt}.
MiniGPT-4 exhibits a wide range of capabilities such as detailed image description generation, website creation from hand-written drafts, writing stories and poems inspired by given images, providing solutions to problems shown in images, and teaching users how to cook based on food photos. It is also highly computationally efficient, training a projection layer utilizing approximately 5 million aligned image-text pairs \cite{zhu2023minigpt}.

\subsubsection{\textbf{The Segment Anything Model (SAM) model}}
The Segment Anything Model (SAM) \cite{kirillov2023segment} is a foundational model for image segmentation that is designed to be promptable, meaning it can be guided to perform specific tasks via prompts. The task of SAM is to return a valid segmentation mask given any segmentation prompt. A prompt can include spatial or text information identifying an object, and the output should be a reasonable mask for at least one of the objects specified by the prompt. SAM consists of three main components: an image encoder that computes an image embedding, a prompt encoder that embeds prompts, and a lightweight mask decoder that predicts segmentation masks based on the image and prompt embeddings. To train SAM, a diverse, large-scale source of data is needed. The authors built a "data engine" to collect a dataset of over 1 billion masks. This engine has three stages: assisted-manual, semi-automatic, and fully automatic. In the first stage, SAM assists annotators in annotating masks. In the second stage, SAM can automatically generate masks for a subset of objects, and annotators focus on annotating the remaining objects. In the final stage, SAM is prompted with a regular grid of foreground points, yielding an average of around 100 high-quality masks per image. The final dataset, SA-1B, includes more than 1 billion masks from 11 million licensed and privacy-preserving images.

\subsection{\textbf{The Camera Node and Image Description Topic}}
As represented in Figure \ref{fig_rosgpt_vision_architecture_}, ROSGPT\_Vision contains two ROS nodes: Camera Node and GPT Consultation Node. They interact using two ROS topics: Image\_Description and GPT\_Consultation. The Camera Node utilizes the Image Semantics module to extract detailed descriptions from an input image. This node is designed to handle real-time image streaming, processing each image frame as it comes in. It gets the description from the Image Semantics module, and then publishes it to a ROS topic named Image\_Description. The GPT Consultation Node is subscribed to this topic, allowing it to receive and utilize these image descriptions in real-time. 

%%%%%%%%%%%%%%%%%%%%%%%%%%%%%%%%%%%%%%%%%%%%%%%
\begin{algorithm}
\caption{Camera node}
\label{alg:image2text_}
\textbf{Inputs}:
\begin{enumerate}
    \item $DM$: The description model [image2text] [string]
    \item $WC$: The webcam used [True or False]
    \item $VP$: The video path, add path if $WC$ is False [string]
    \item $f$: The frame interval [integer]
\end{enumerate}

\begin{algorithmic}
\Procedure{}{}
    \State $f\_counter \gets 0$
    \If{$WC = \text{True}$}
        \While{$WC = \text{True}$}
            \If{$f\_counter \mod f = 0$}
                \State $frame \gets \text{camera.get\_image()}$
                \State $text \gets \text{DM}(frame)$
                \State $\text{send2publisher}(text, \text{'Image description'})$
                \State $f\_counter \gets f\_counter + 1$
            \EndIf
        \EndWhile
    \Else
        \While{$\text{video.is\_opened()}$}
            \If{$f\_counter \mod f = 0$}
                \State $frame \gets \text{video.get\_image()}$
                \State $text \gets \text{DM}(frame)$
                \State $\text{send2publisher}(text, \text{'Image description'})$
                \State $f\_counter \gets f\_counter + 1$
            \EndIf
        \EndWhile
    \EndIf
    \State $consultation \gets \text{subscribe}(\text{'GPT consultation'})$
\EndProcedure
\end{algorithmic}
\end{algorithm}

\begin{figure}[htbp]
    \centering
    \includegraphics[width=0.5\textwidth]{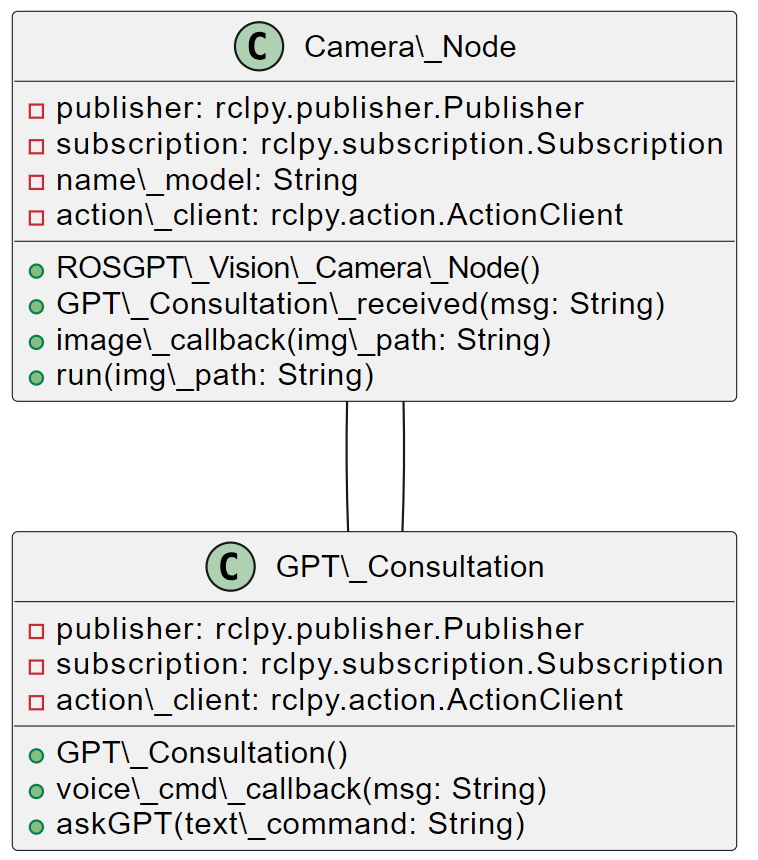}
    \caption{UML diagram}
    \label{fig:label}
\end{figure}

The Camera\_Node class converts the input image into descriptive representations, following a predetermined prompt according to Algorithm \ref{alg:image2text_}. It can use both video paths and webcam streams. Subsequently, the Camera\_Node class proceeds to transmit the resulting descriptions to the GPT\_Consultation class via the Image Description topic. Figure \ref{fig:label} shows the camera\_node and GPT Consultation classes diagrams following the the Unified Modeling Language (UML) framework.

\subsection{\textbf{The GPT Consultation Node and Topic}}
The two nodes (Camera and GPT Consultation) are completely separated. The task logic must be completely separated from the image description logic, following the PRM design pattern. The Camera node focuses on accurately describing the image without being concerned about the actions that should be taken based on it. The second node, the GPT Consultation node, is an Ontology-Based Action Planner Node. It subscribes to the Image\_Description topic published by the Camera node and uses this information to suggest appropriate actions for the robot. The node should explicitly emphasize a specially designed robot ontology behind the given task. A robot ontology is a formal representation of knowledge within the robotics domain. This ontology includes definitions of concepts, properties, and relationships pertaining to the robot and its environment, providing a structured and semantically rich framework for reasoning and decision-making.\par
The GPT Consultation Node uses the robot ontology to map the image descriptions to potential actions according to Algorithm \ref{alg:image2text_2}. The final feedback relevant to the task is published into the topic GPT\_Consultation. This name is chosen over the name GPT\_Action because it is more generic. Consultation refers to the real feedback from GPT, which can be just a general notification, a decision-aid estimation, or an action to be taken by other nodes. \par
%%%%%%%%%%%%%%%%%%%%%%%%%%%%%%%%%%%%%%%%%%%%%%%%%%%%%%%%%%%%%

\begin{algorithm}
\caption{GPT Consultation node}
\label{alg:image2text_2}
\textbf{Inputs}:
\begin{enumerate}
    \item $Key$: The OpenAI key [string]
    \item $Temp$: The ChatGPT temperature [float]
\end{enumerate}

\begin{algorithmic}
\Procedure{}{}
    \State $text \gets \text{subscribe}(\text{'Image description'})$
    \State $response \gets \text{askGPT}(Key, Temp, text)$
    \State $\text{send2publisher}(response, \text{'GPT consultation'})$
\EndProcedure
\end{algorithmic}
\end{algorithm}

The main component of the node is its interaction with the LLM, because it represents the Task Modality in the PRM design pattern. The framework contains several parameters collected in one YAML file. YAML file consists of the Task name and parameters of Camera\_node and GPT\_node and image description models. The main two parameters to change in the YAML file are the prompts: the LLM and the Visual Prompts. Other parameters can be let unchanged. 

\subsection{\textbf{Visual and LLM prompts}}
ROSGPT\_Vision is configured using a single YAML (Yet Another Markup Language) file which describes the robot task and saved under the folder cfg of the ROSGPT\_Vision folder. YAML files are commonly used in ROS for their human-readable data serialization format, allowing for easy configuration and parameter setting.  Notably, this file includes the configurations for both visual and language model (LLM) prompts. Visual prompts guide the image processing and semantic extraction capabilities of the node, while LLM prompts are used to instruct the language model to generate appropriate textual descriptions or actions based on the image semantics. This consolidated configuration approach allows for efficient setup and modification of the ROSGPT\_Vision node, facilitating its integration and use in various robotic applications. 

\begin{figure}[htbp]
    \centering
    \includegraphics[width=0.5\textwidth]{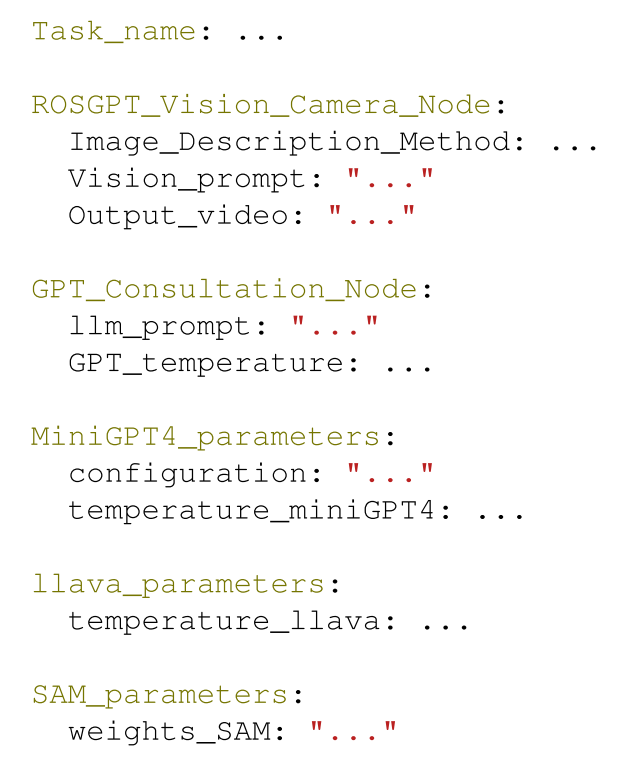}
    \caption{YAML file format in ROSGPT\_Vision}
    \label{fig:yaml-configuration}
\end{figure}

The structure of the YAML file is provided in the the Figure \ref{fig:yaml-configuration}. The YAML contains 6 main sections of configurations parameters:

\begin{itemize}
\item \texttt{Task\_name}: This field specifies the name of the task that the ROS system is configured to perform. 
\item \texttt{ROSGPT\_Vision\_Camera\_Node}: This section contains the configuration for the ROSGPT\_Vision\_Camera\_Node. 
  \begin{itemize}
  \item \texttt{Image\_Description\_Method}: This field specifies the method used by the node to generate descriptions from images. It can be one of the currently developed methods: MiniGPT4, LLaVA, or SAM. The configurations needed for everyone of them is put separately at the end of this file. 
  \item \texttt{Vision\_prompt}: This field specifies the prompt used to guide the image description process.
  \item \texttt{Output\_video}: This field specifies the path or the name of where to save the  output video file.
  \end{itemize}

\item \texttt{GPT\_Consultation\_Node}: This section contains the configuration for the GPT\_Consultation\_Node.
  \begin{itemize}
  \item \texttt{llm\_prompt}: This field specifies the prompt used to guide the language model.
  \item \texttt{GPT\_temperature}: This field specifies the temperature parameter for the GPT model, which controls the randomness of the model's output.
  \end{itemize}

\item \texttt{MiniGPT4\_parameters}: This section contains the configuration for the MiniGPT4 model. It should be clearly set if the model is used in this task, otherwise it could be empty. 
  \begin{itemize}
  \item \texttt{configuration}: This field specifies the path for the configuration file of MiniGPT4.
  \item \texttt{temperature\_miniGPT4}: This field specifies the temperature parameter for the MiniGPT4 model.
  \end{itemize}

\item \texttt{llava\_parameters}: This section contains the configuration for the llavA model (if used).
  \begin{itemize}
  \item \texttt{temperature\_llavA}: This field specifies the temperature parameter for the llavA model.
  \end{itemize}

\item \texttt{SAM\_parameters}: This section contains the configuration for the SAM model.
  \begin{itemize}
  \item \texttt{weights\_SAM}: This field specifies the weights used by the SAM model.
  \end{itemize}
\end{itemize}

\subsection{ROSGPT\_Vision Implementation on ROS2}
ROSGPT\_Vision code is shared with the community, under the link: \href{https://github.com/bilel-bj/ROSGPT\_Vision}{https://github.com/bilel-bj/ROSGPT\_Vision}. Our aim is to fosters further advancements in the realm of visual data comprehension and interaction, leveraging the capabilities of ROS in conjunction with the potential of Language and Vision Language Models (LLMs and VLMs). %This implementation serves as a stepping stone towards augmenting robotic systems with an enhanced understanding of visual data, facilitated by the application of natural language models. The project's repository is open to the public on GitHub, encouraging community contributions and enhancements.
\section{Proof-of-Concept: the CarMate Application}
\begin{figure*}[htbp]
    \centering
    \includegraphics[width=0.9\linewidth]{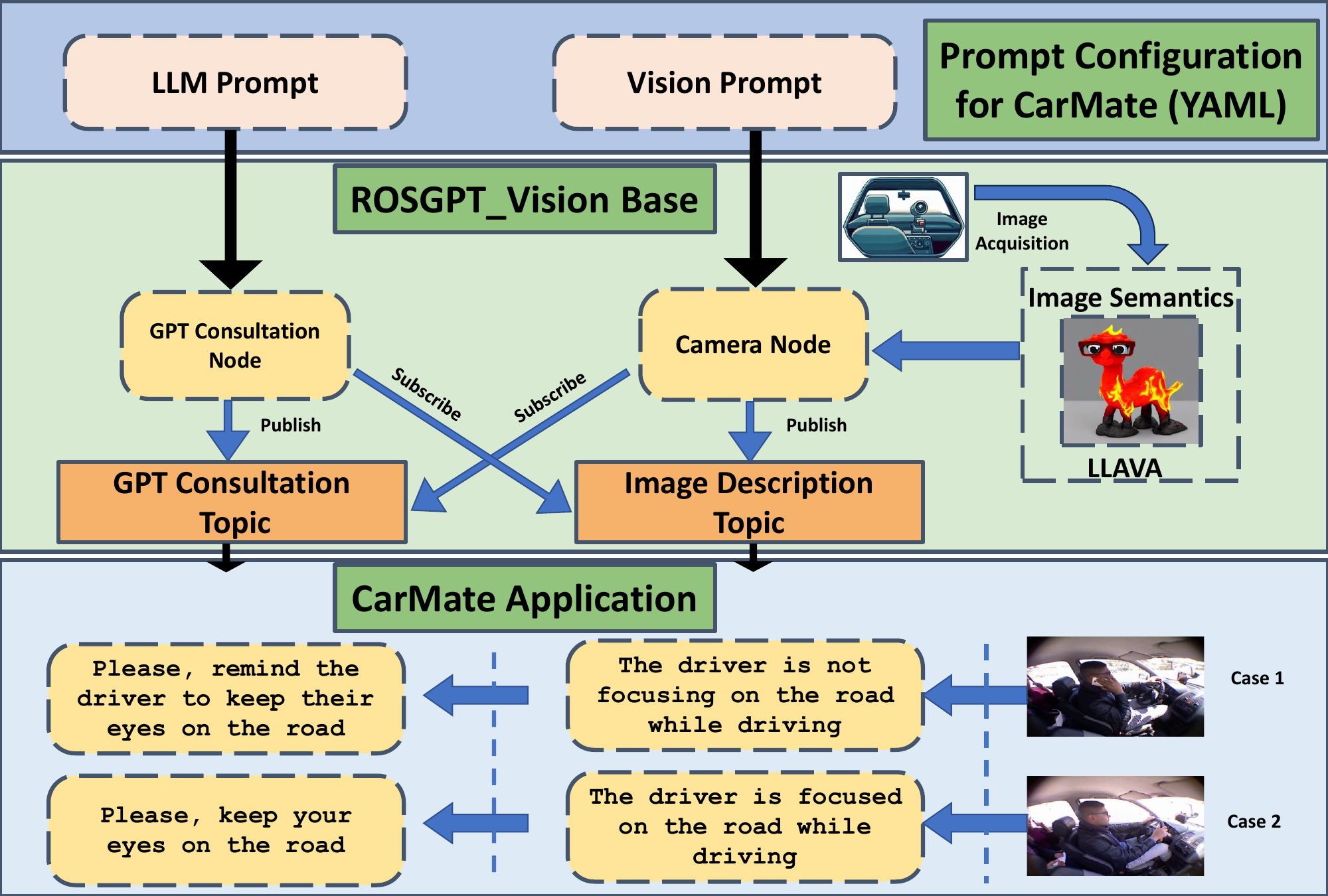}
    \caption{CarMate Application diagram.}
    \label{fig:carmate_architecture}
\end{figure*}
We present a software application named CarMate based on the ROSGPT\_Vision framework, and the PRM design pattern. CarMate is an advanced driver monitoring and assistance application designed to ensure safe and efficient driving experiences. CarMate is developed by just configuring the YAML file of the  ROSGPT\_Vision framework, and mainly by just setting the two prompts. Through a camera mounted inside the car, CarMate continuously monitors the driver's actions, expressions, and attention levels. The system is capable of detecting unsafe behaviors such as distracted driving, drowsiness, or aggressive driving patterns. When such behaviors are identified, CarMate issues timely alerts heard by the driver, prompting corrective action. 

In addition to real-time monitoring and alerts, CarMate helps generate personalized driving reports. These reports summarize the driver's behavior over a particular period, highlighting areas of strength and those requiring improvement. By providing clear, actionable insights, CarMate aids drivers in developing safer driving habits.

\subsection{\textbf{The Architecture of CarMate}}  
The CarMate application's architecture follows the ROSGPT\_Vision framework, as depicted in Figure \ref{fig:carmate_architecture}. Initially, the app captures driver images at five-second intervals. Subsequently, the image semantics model (LLAVA) processes the images by employing a conditional vision description prompt, as indicated in Figure \ref{fig:yaml-configuration}. The prompt used is ["Describe the driver’s current level of focus on driving based on the visual cues. Answer with one short sentence."]. Following this, the Camera node publishes the vision description to the image description topic. The GPT consultation node subscribes to the image description topic and forwards the vision description, along with an LLM prompt, to ChatGPT 3.5. The prompt employed is ["Consider the following ontology:
You must write your reply with one short sentence. Behave as a CarMate that surveys the driver and gives him advice and instruction to drive safely.
You will be given human language prompts describing an image. Your task is to provide appropriate instructions to the driver based on the description."]. Subsequently, ChatGPT 3.5 determines whether to caution the driver if they are at risk or encourage them to maintain focus on the road. The lower section of Figure \ref{fig:carmate_architecture} presents two examples: the first involves a driver speaking on a cell phone while driving, and the second depicts a driver maintaining focus on driving.
%%%%%%%%%%%%%%%%%%%%%%%%%%%%%%%%%%%%%%%%%%%%%%%%%%%%%%%

\subsection{\textbf{Unified prompting through the YAML file}} 
Presented below is the YAML configuration file used for CarMate:

\begin{itemize}
\item \texttt{Task\_name}: driver phone usage
\item \texttt{ROSGPT\_Vision\_Camera\_Node}: 
  \begin{itemize}
  \item \texttt{Image\_Description\_Method}: llava \# comment: you can choose between [llava, MiniGPT4, SAM]
  \item \texttt{Vision\_prompt}:"Describe the driver’s current level of focus on driving based on the visual cues, Answer with one short sentence."
  \item \texttt{Choose\_input}:"video" \# comment:  you can choose between [webcam, video]
  \item \texttt{Input\_video}:"Absolute path" \# comment: if you chose video 
  \item \texttt{Output\_video}:"Absolute path of output video demo"
  \end{itemize}

\item \texttt{GPT\_Consultation\_Node}:
  \begin{itemize}
  \item \texttt{llm\_prompt}:"Consider the following ontology:  
  You must write your Reply with one short sentence. Behave as a carmate that surveys the driver and gives him advice and instruction to drive safely. 
  You will be given human language prompts describing an image. Your task is to provide appropriate instructions to the driver based on the description."
  \item \texttt{GPT\_temperature}: 0.2
  \end{itemize}

\item \texttt{MiniGPT4\_parameters}: 
  \begin{itemize}
  \item \texttt{configuration}:"minigpt4\_eval.yaml" \# comment: absolute path for configuration of MiniGPT4 model
  \item \texttt{temperature\_miniGPT4}: 0.2
  \end{itemize}

\item \texttt{llava\_parameters}:
  \begin{itemize}
  \item \texttt{temperature\_llavA}: 0.2
  \item \texttt{llama\_version}:"13B" \# comment: you can choose between [7B, 13B]
  \end{itemize}

\item \texttt{SAM\_parameters}: 
  \begin{itemize}
  \item \texttt{weights\_SAM}:"sam\_vit\_h\_4b8939.pth" \# comment: absolute path for configuration of MiniGPT4 mode
  \end{itemize}
\end{itemize}
%%%%%%%%%%%%%%%%%%%%%%%%%%%%%%%%%%%%%%%%%%%%%%%%%%%%%%%%%%%%%%%%%%%%%%%%%%%%

\subsection{Prompting strategies}

Prompting strategies for the CarMate application can be designed to accurately describe the driver's behavior and provide appropriate consultations. Here are some strategies for both visual and language model (LLM) prompts:

\begin{itemize}
\item \textbf{Visual Prompts:}
\begin{itemize}
\item \textbf{Focused Description Prompts:} These prompts can be designed to specifically describe the driver's focus on driving. For example, "Describe the driver's current level of focus on driving based on the visual cues."
\item \textbf{Behavioral Description Prompts:} These prompts can be used to describe the driver's overall behavior. For example, "Describe the driver's overall behavior, including any distractions or signs of fatigue."
\item \textbf{Ontological Prompts:} These prompts can be designed to extract specific ontological entities from the visual data. For example, "Identify and describe the ontological entities related to driving focus in the current scene."
\end{itemize}

\item \textbf{LLM Prompts:}
\begin{itemize}
\item \textbf{Consultative Prompts:} These prompts can be used to generate appropriate consultations based on the visual descriptions. For example, "Based on the visual description, provide a consultation to the driver about their current level of focus on driving."
\item \textbf{Action-Oriented Prompts:} These prompts can be used to suggest actions that the driver should take based on the visual descriptions. For example, "Suggest actions the driver should take to improve their focus on driving based on the visual description."
\item \textbf{Ontological Prompts:} These prompts can be used to generate consultations based on specific ontological entities. For example, "Provide a consultation to the driver based on the identified ontological entities related to driving focus."
\end{itemize}
\end{itemize}

\begin{table}[!ht]
  \caption{The Best Prompts in ten cases}
  \label{best_prompts_1}
  \centering
  \begin{tabular}{p{0.9cm}p{0.9cm}p{0.9cm}p{0.8cm}p{0.8cm}p{0.9cm}p{0.8cm}}%{lllllll}
    \toprule
     & \multicolumn{3}{c}{Visual Prompts} & \multicolumn{3}{c}{LLM Prompts}    \\
     \cmidrule(l){2-7}
    Cases & \rotatebox{45}{Focused Description} & \rotatebox{45}{Behavioral Description} & \rotatebox{45}{Ontological} & \rotatebox{45}{Consultative} & \rotatebox{45}{Action-Oriented}  & \rotatebox{45}{Ontological} \\
    \midrule
    Case 1  &       &       & \checkmark &       & \checkmark&       \\
    Case 2  &       &       & \checkmark &\checkmark&       &       \\
    Case 3 &         & \checkmark &          & \checkmark &         &         \\
    Case 4 & \checkmark &         &          & \checkmark &         &         \\
    Case 5 & \checkmark &         &          & \checkmark &       &           \\
    Case 6 &         & \checkmark &          &         & \checkmark &         \\
    Case 7 &         & \checkmark &          & \checkmark &         &         \\
    Case 8 & \checkmark &         &          & \checkmark &         &      \\
    Case 9 &         & \checkmark &          &         & \checkmark &         \\
    Case 10 &         & \checkmark &          & \checkmark &         &         \\
    \bottomrule
  \end{tabular}
\end{table}

In this particular case, given the focus of the CarMate application on driver behavior monitoring and consultation, an optimal approach would involve a combination of focused description prompts for visual cues and consultative prompts for the language model (LLM). However, determining the most effective prompting strategy would ultimately depend on the specific requirements of the application and the capabilities of the models employed.

To evaluate the best prompts for visual cues and LLM, we devised ten distinct scenarios that involved distracting the driver's attention while driving. Table \ref{best_prompts_1} presents the optimal prompts for each scenario, and additional visual representations can be found in the appendix. Upon reviewing the appendix and Table \ref{best_prompts_1}, it becomes evident that Behavioral Description of Visual Prompts outperforms other prompt types due to its adaptability and ability to provide a clear description for each scenario. Conversely, Focused Description prompts, while exhibiting high efficacy in some cases, generally fall short in delivering a precise description due to their limited adaptability. Ontological prompts, on the other hand, offer excellent efficiency but provide a comprehensive image description without focusing on the driver's behavior during driving, thereby impacting decision-making from the GPT Consultation perspective.

Regarding LLM prompts, the Consultative prompt type exhibits superiority over other prompt types due to its exclusive focus on describing the driver's condition during driving, without offering any directives. This aligns with the expected responsibilities of the CarMate application. Conversely, Action-Oriented Prompts enable direct decision-making without describing the driver's behavior, while Ontological Prompts primarily offer assistance and advice to the driver in most cases without providing any description of their behavior.

Therefore, based on our analysis, a combination of focused description prompts for visual cues and consultative prompts for LLM is recommended for the CarMate application, taking into account the specific scenarios and requirements of the system.

\subsection{\textbf{Description of the Datasets used in CarMate}} 
In this paper, we utilized two publicly available datasets, namely MDAD \cite{jegham2019mdad} and 3MDAD \cite{jegham2020novel}, to address the challenges of driver action recognition and driver distraction monitoring. The MDAD dataset provides synchronized RGB and depth data from side and frontal views, allowing researchers to develop algorithms across multiple modalities and views under different lighting and weather conditions. The 3MDAD dataset includes synchronized data modalities from daytime and nighttime recordings, offering a comprehensive dataset for evaluating driver distraction monitoring methods. Both datasets enable an independent analysis of recognition results based on individual modalities as well as combinations of modalities. These datasets provide valuable resources for advancing research in the field of driver behavior analysis and improving road safety.
\section{Conclusion}

In this paper, we have introduced a new Robotic design pattern: the Prompting Robotic Modalities (PRM). It opts for designing robotic application as Robotic Modalities connected to Task Modality. Every Modality is commanded via a specific prompt. Every Robotic Modality incorporates a specific Modality Language Model (MLM) while the Task Modality is connected to a Large Language Model (LLM). The PRM design pattern is used to conceive a new robotic framework: the ROSGPT\_Vision, a novel robotics framework that connects language models to computer vision capabilities, enabling enhanced human-robot interaction and understanding. By integrating LLMs with ROS and VLMs, ROSGPT\_Vision tries to make the robots interact with visual data through natural language.  The Visual Robotic Modality of ROSGPT\_Vision lies in the Image Semantics module, which effectively extracts meaning from images and generates descriptive narrations in human-like language. It integrates advanced VLM models such as LLAVA, MiniGPT-4, and SAM.

This paper tested the ROSGPT\_Vision framework in the development of the CarMate application, a driver monitoring and assistance system designed to ensure safe and efficient driving experiences. By continuously monitoring driver actions, expressions, and attention levels, CarMate can identify unsafe behaviors and issue timely alerts to prompt corrective actions. Additionally, CarMate generates personalized driving reports, providing valuable insights to drivers for developing safer driving habits. The whole development of the CarMate is reduced remarkably due to the PRM and ROSGPT\_Vision, it needs mainly the setting of only two prompts in the YAML file.

To optimize the performance of CarMate, we have explored various prompting strategies for both visual and language model prompts. Through thorough evaluation and analysis, we have identified the most effective prompt types, enabling CarMate to deliver precise and contextually appropriate consultations to drivers.

In conclusion, PRM and ROSGPT\_Vision represents a new contribution to the realm of robotics. By releasing the open-source implementation of ROSGPT\_Vision, we invite the robotics community to collaborate and enhance this interdisciplinary research field further. The successful integration of LLMs with Image data analysis within the ROS system increases the potential for innovative applications, such as virtual assistants and intelligent robots that can seamlessly interact with humans and their environment.

\bibliographystyle{IEEEtran}
\bibliography{references}

%%%%%%%%%%%%%%%%%%%%%%%%%%%%%%%%%%%%%%%%%%%%%%%%%
\appendix
\section{Appendix}
\subsection{Prompting strategies}
%%%%%%%%%%%%%%%%%%%%%%%%%%%%%%%%%%%%%%%%%%%%%%%%%
\subsubsection{Case 1: Drinking Coffee during driving}

\paragraph{Visual Prompts}

\subsectionimage{Focused Description}{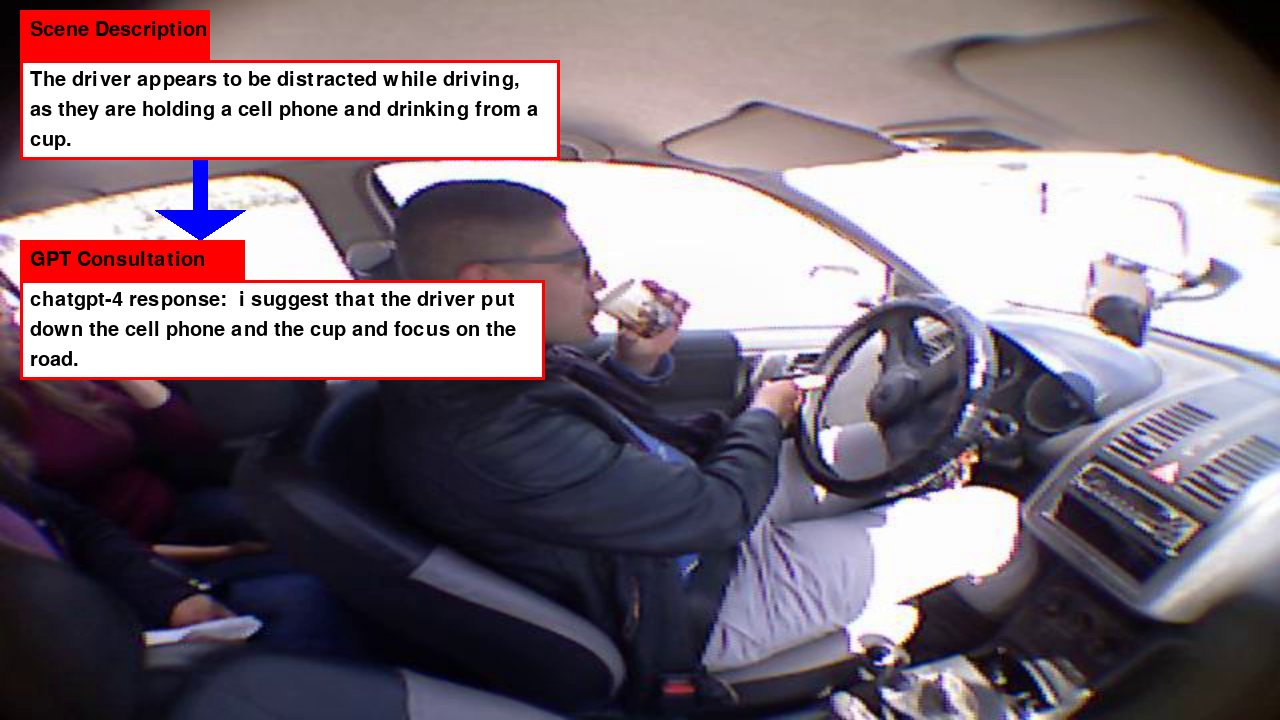}

\subsectionimage{Behavioral Description}{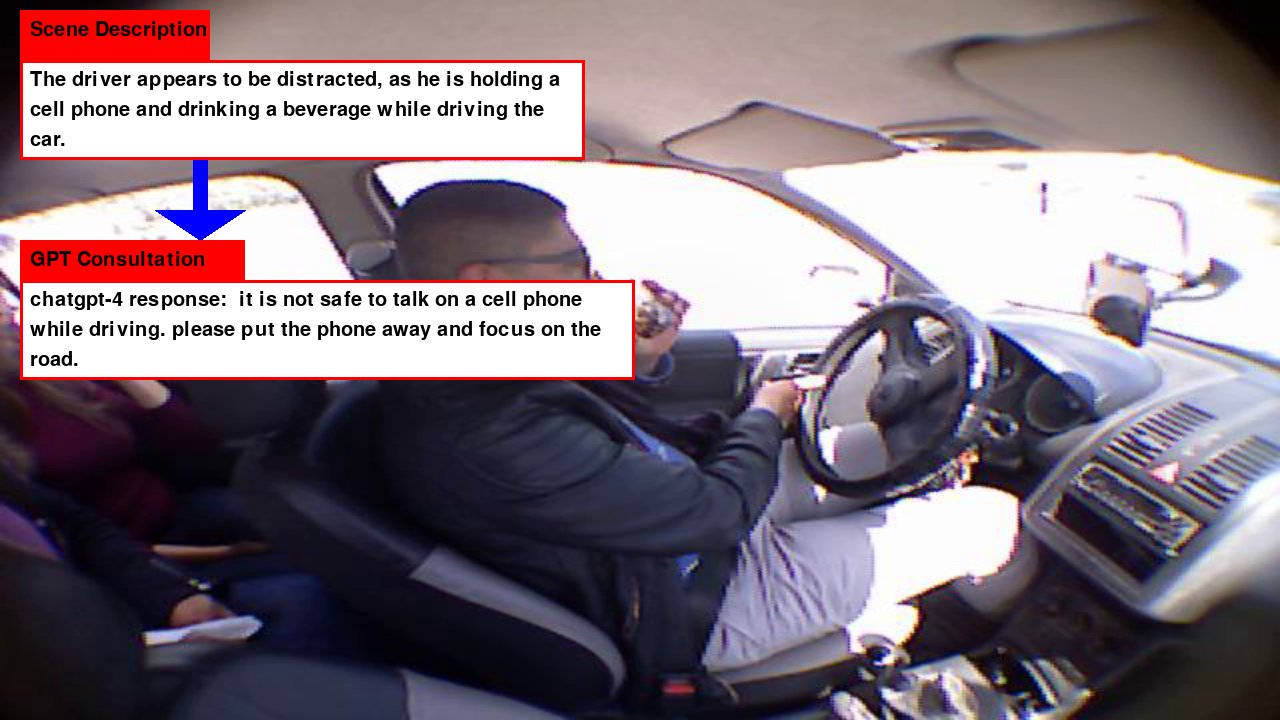}

\subsectionimage{Ontological}{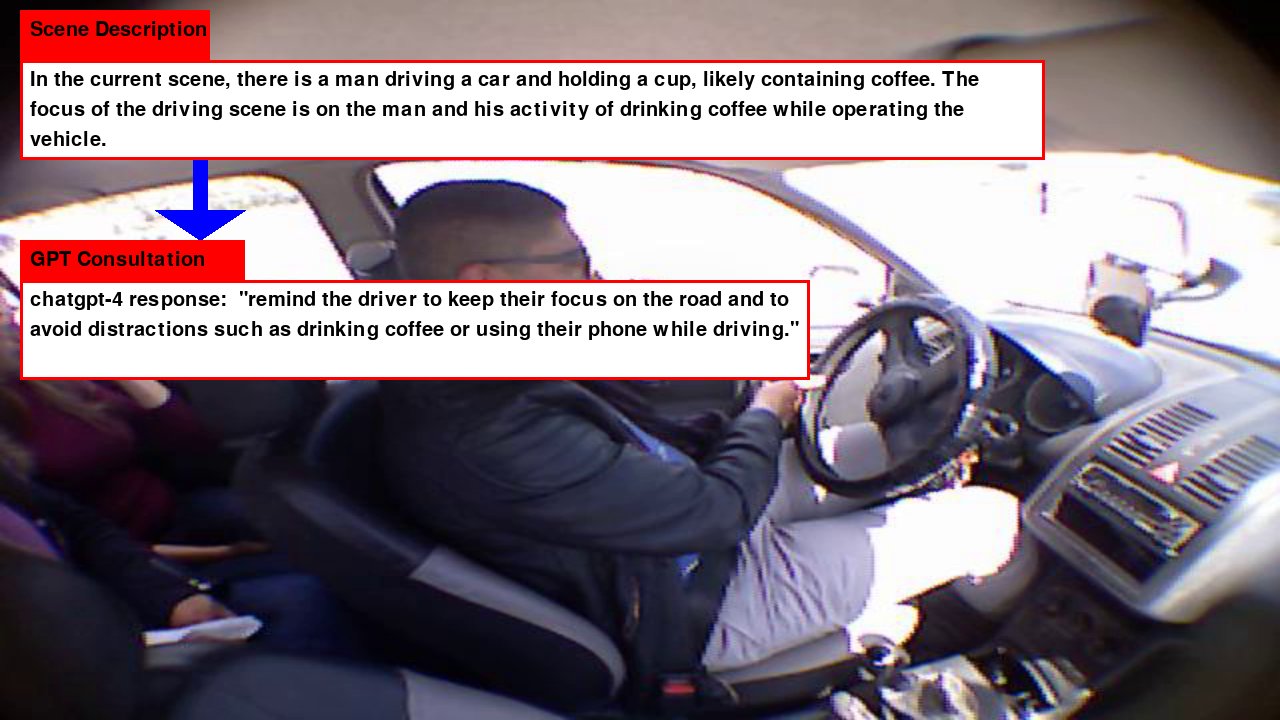}

\paragraph{LLM Prompts}

\subsectionimage{Consultative}{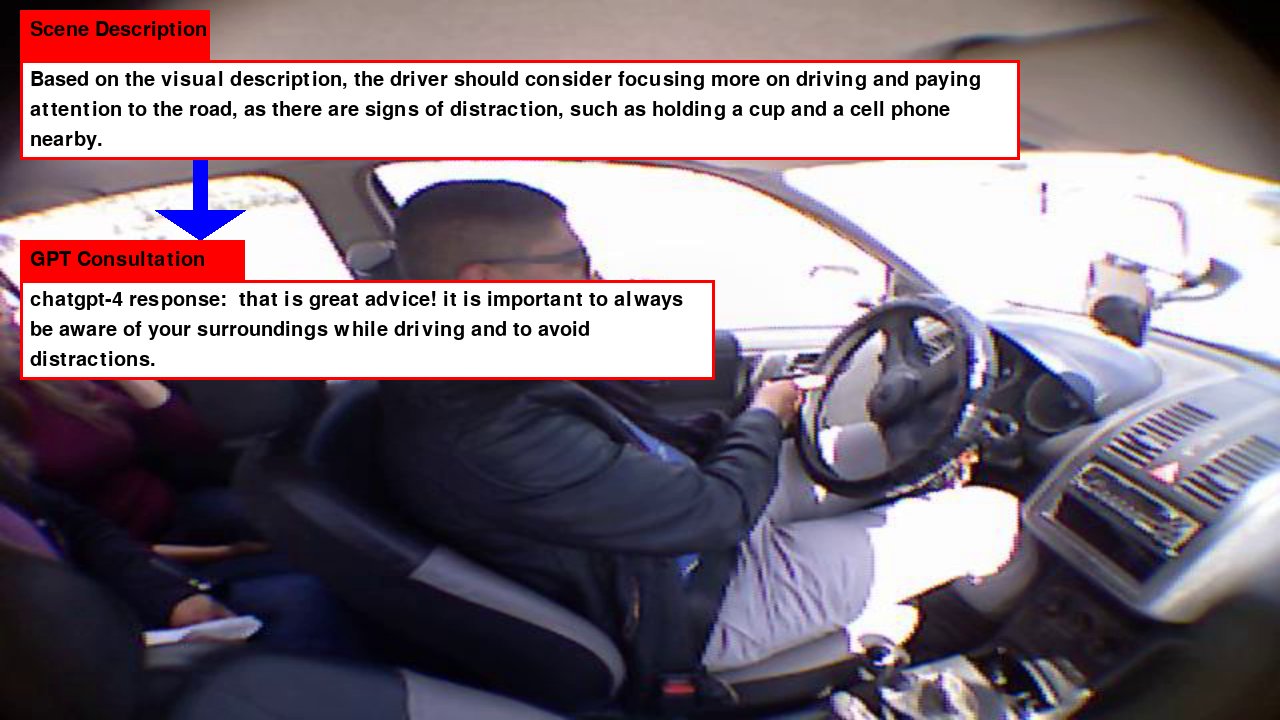}

\subsectionimage{Action-Oriented}{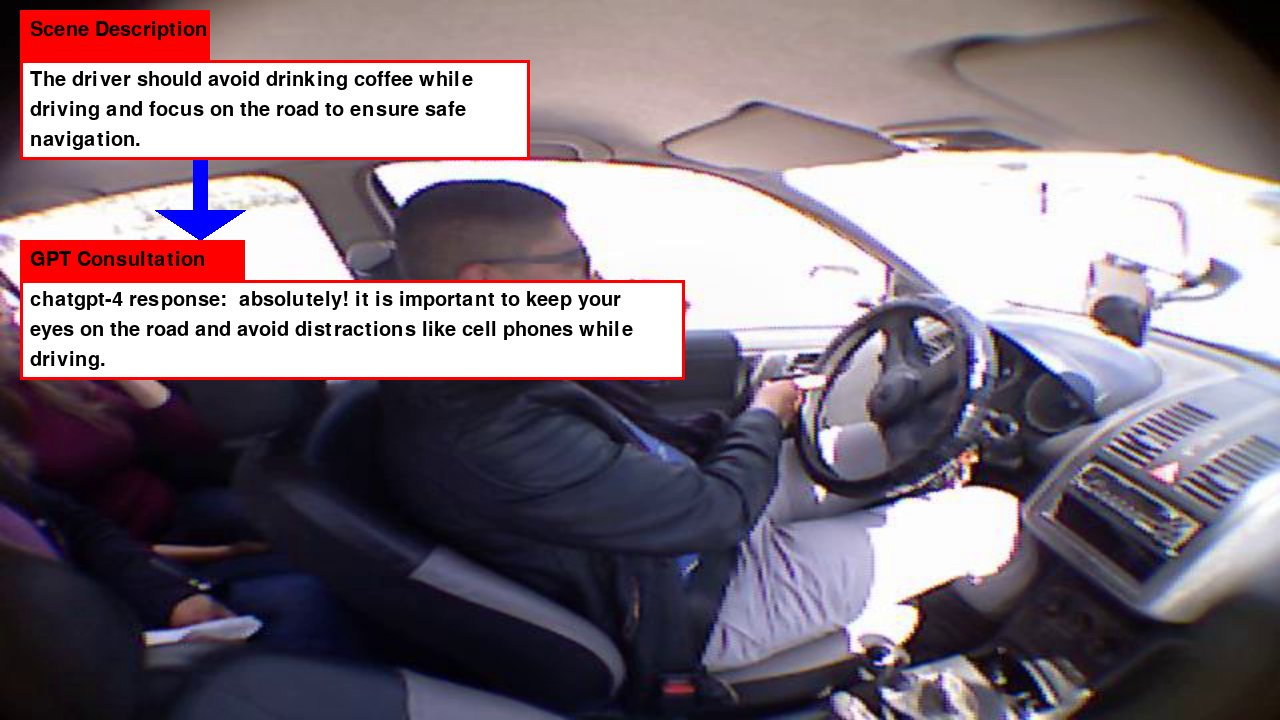}

\subsectionimage{Ontological}{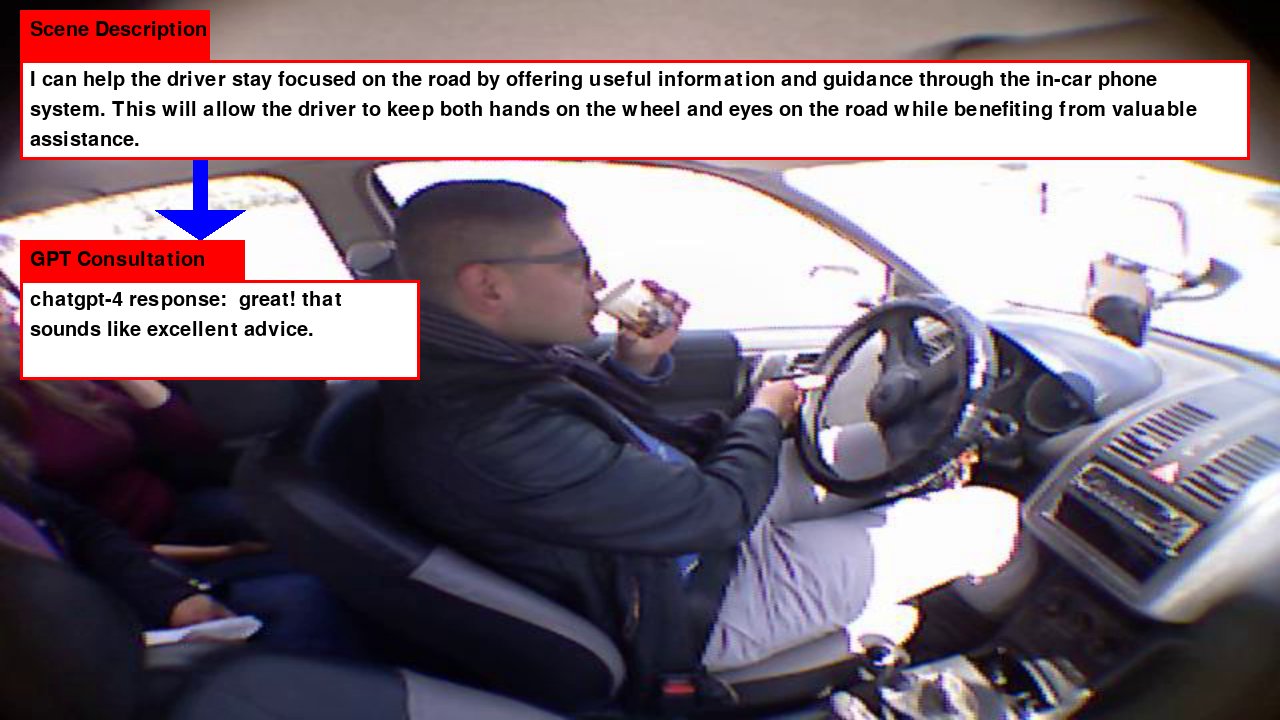}
%%%%%%%%%%%%%%%%%%%%%%%%%%%%%%%%%%%%%%%%%%%%%%%%%
\subsubsection{Case 2: Focus on the road during driving}

\paragraph{Visual Prompts}

\subsectionimage{Focused Description}{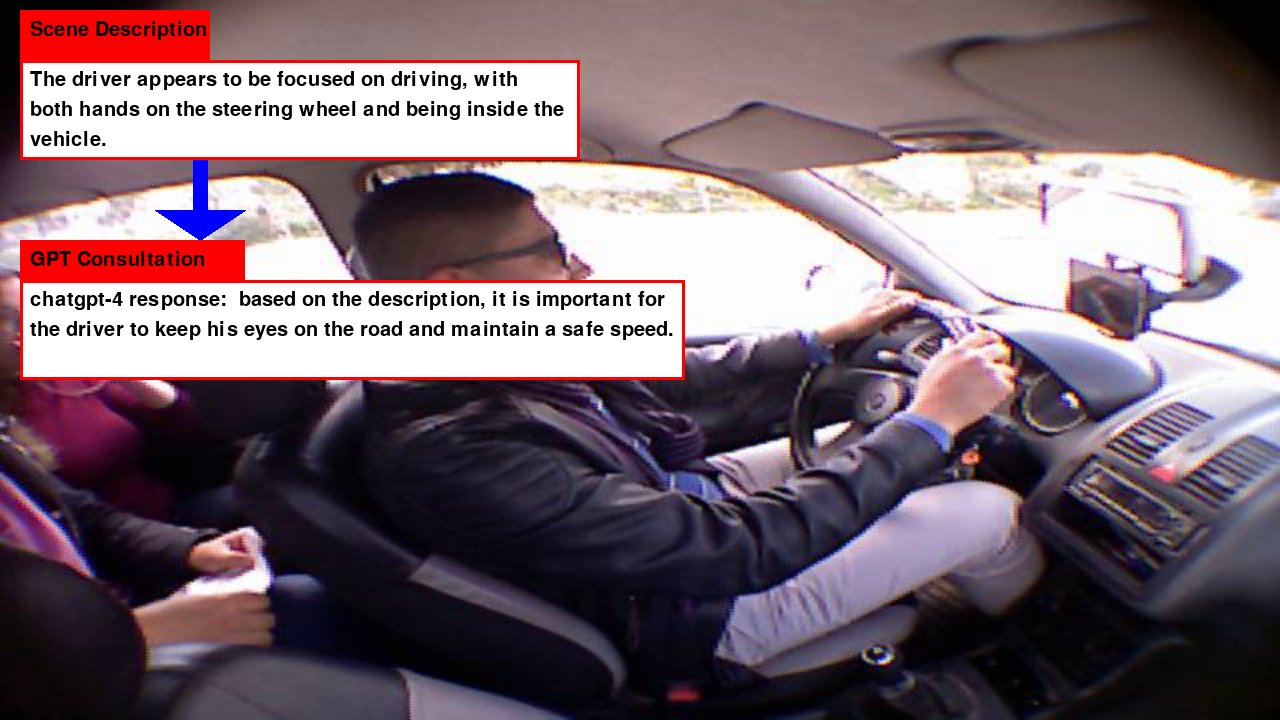}

\subsectionimage{Behavioral Description}{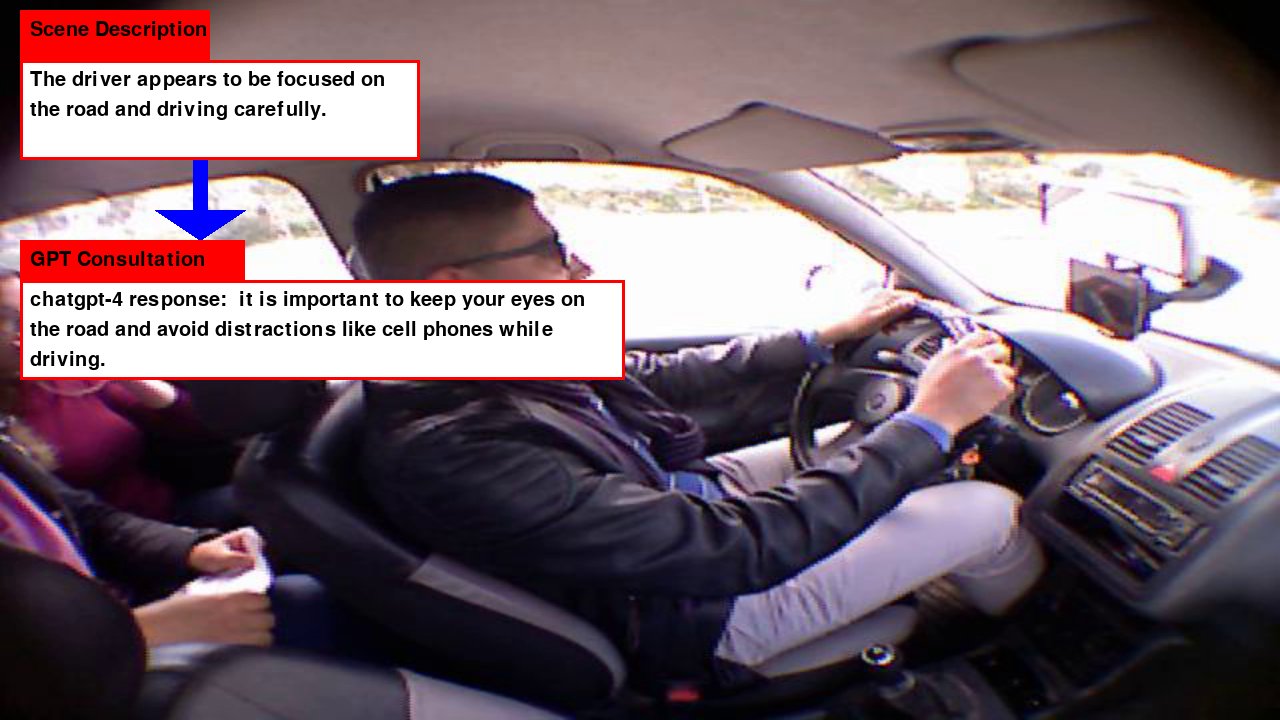}

\subsectionimage{Ontological}{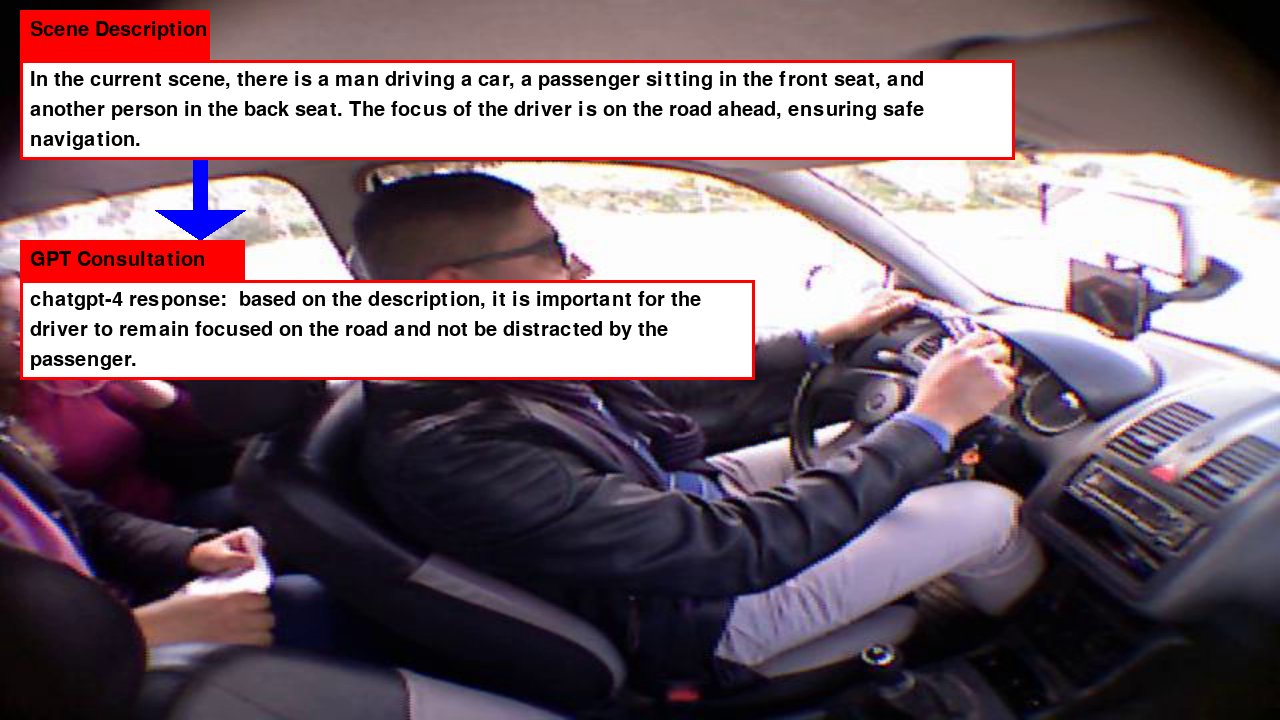}

\paragraph{LLM Prompts}

\subsectionimage{Consultative}{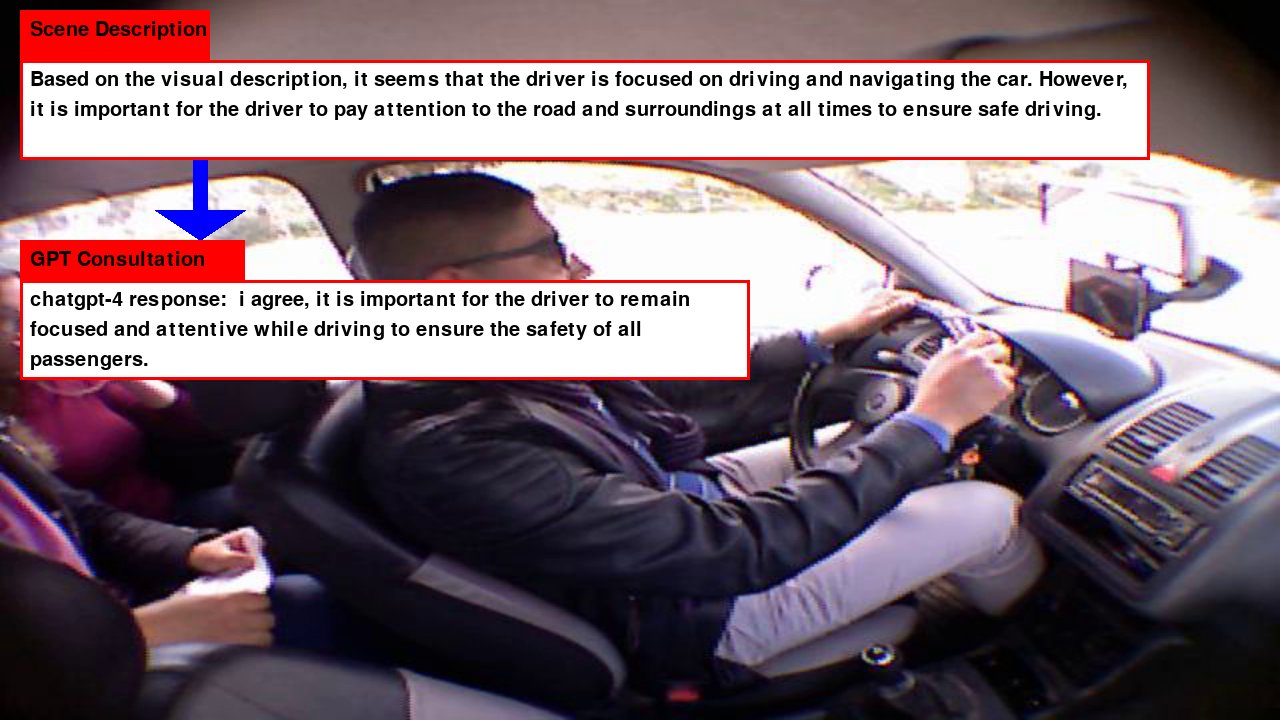}

\subsectionimage{Action-Oriented}{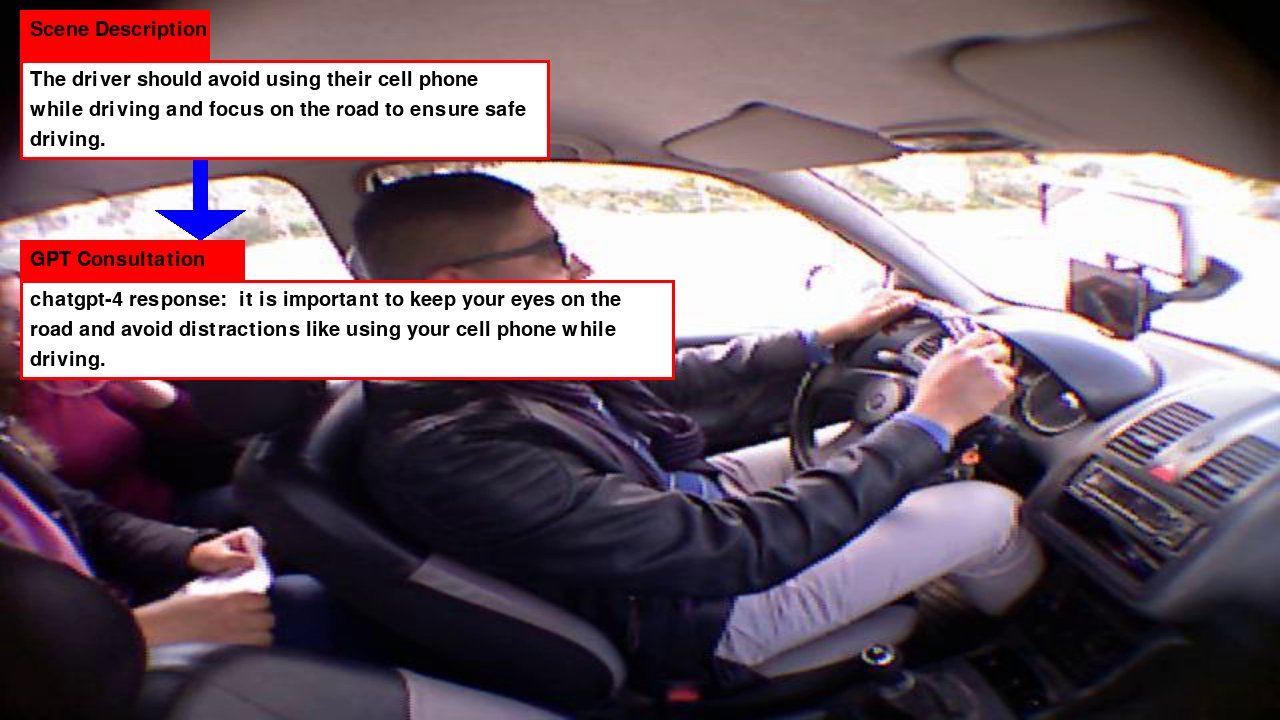}

\subsectionimage{Ontological}{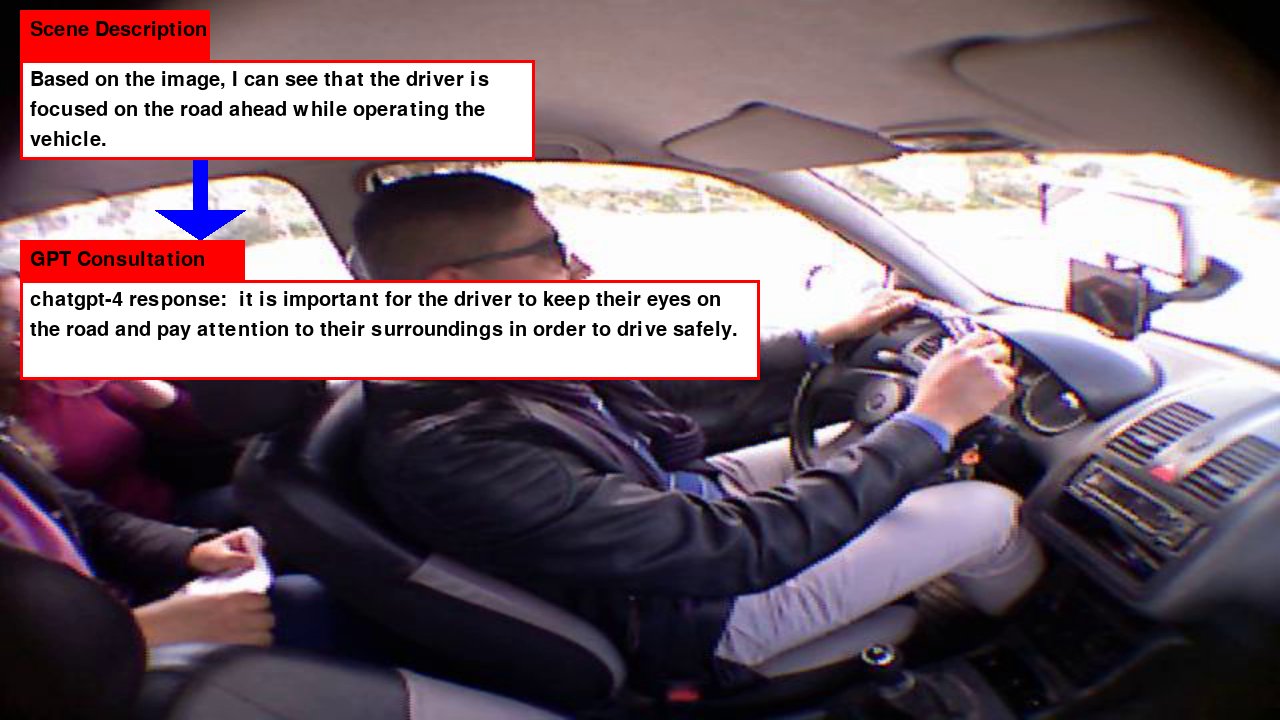}
%%%%%%%%%%%%%%%%%%%%%%%%%%%%%%%%%%%%%%%%%%%%%%%%%
\subsubsection{Case 3: holding a cup during driving}

\paragraph{Visual Prompts}

\subsectionimage{Focused Description}{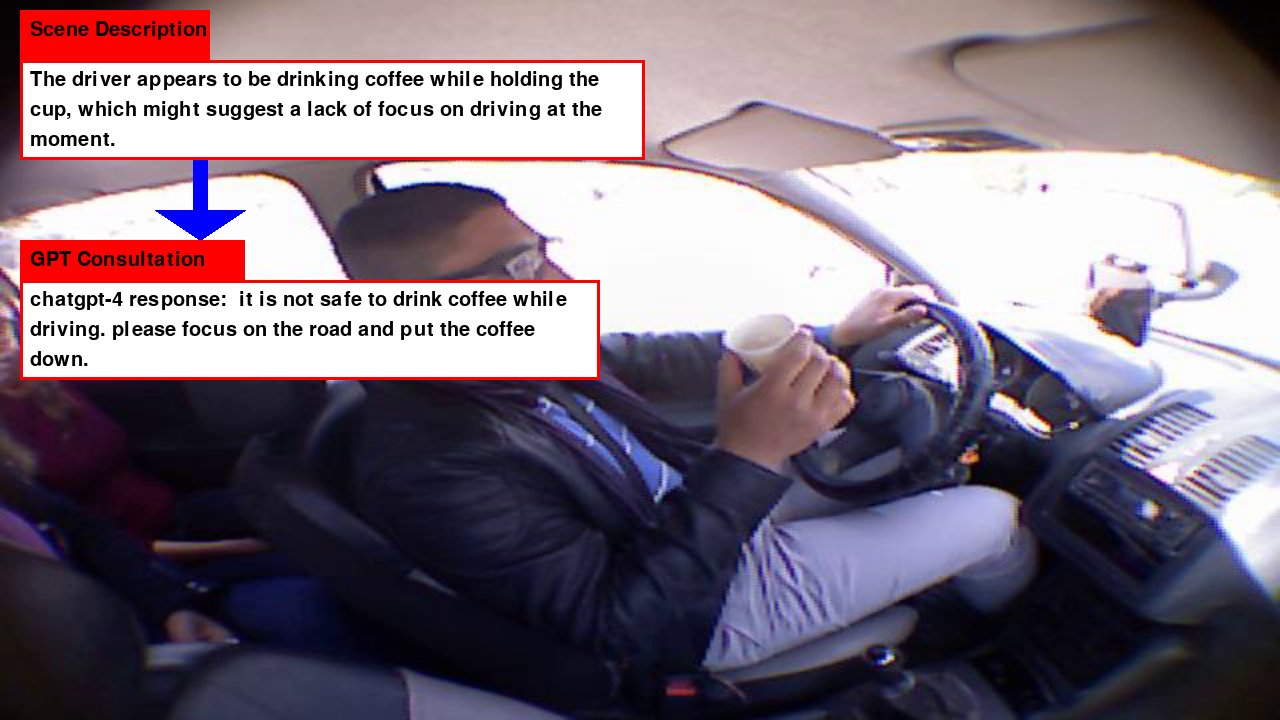}

\subsectionimage{Behavioral Description}{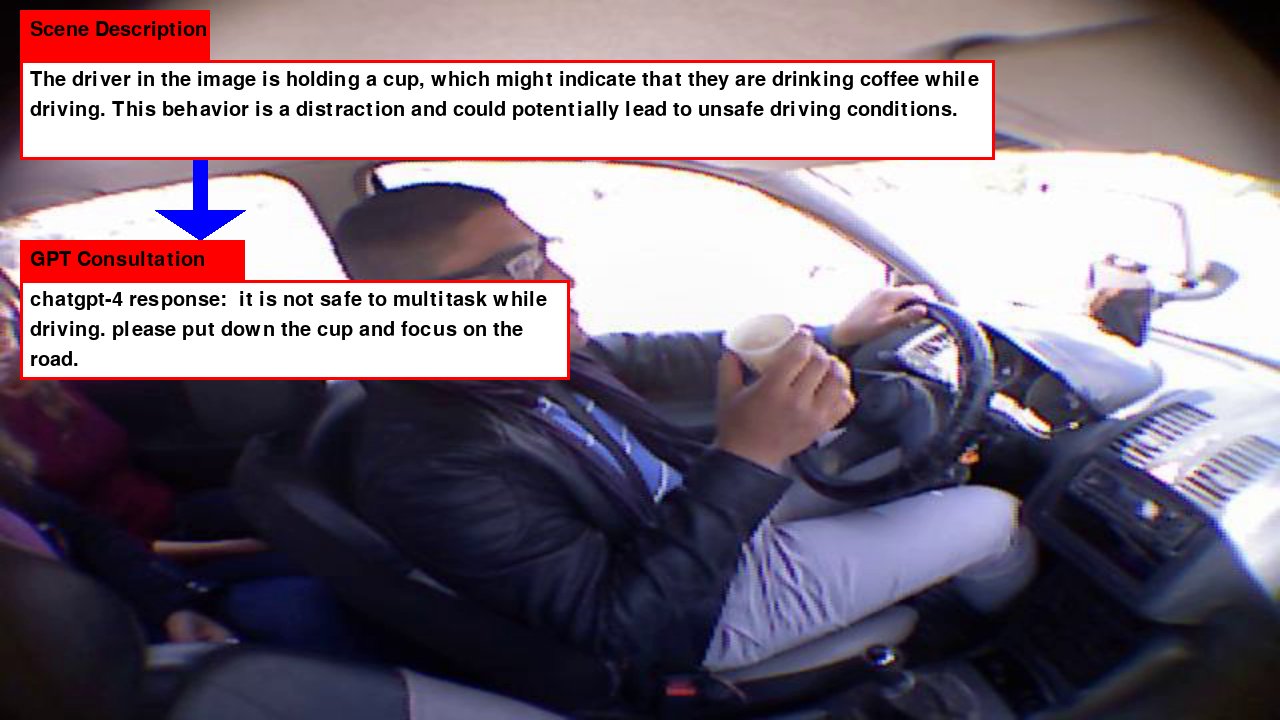}

\subsectionimage{Ontological}{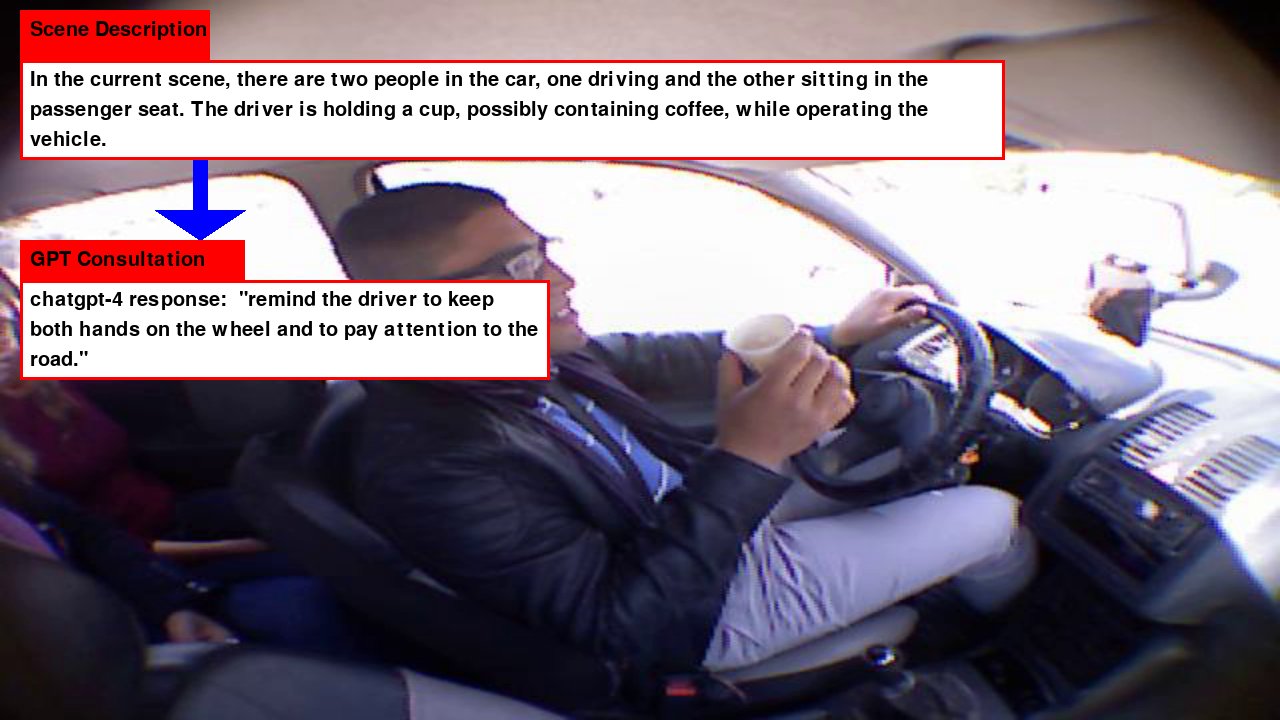}

\paragraph{LLM Prompts}

\subsectionimage{Consultative}{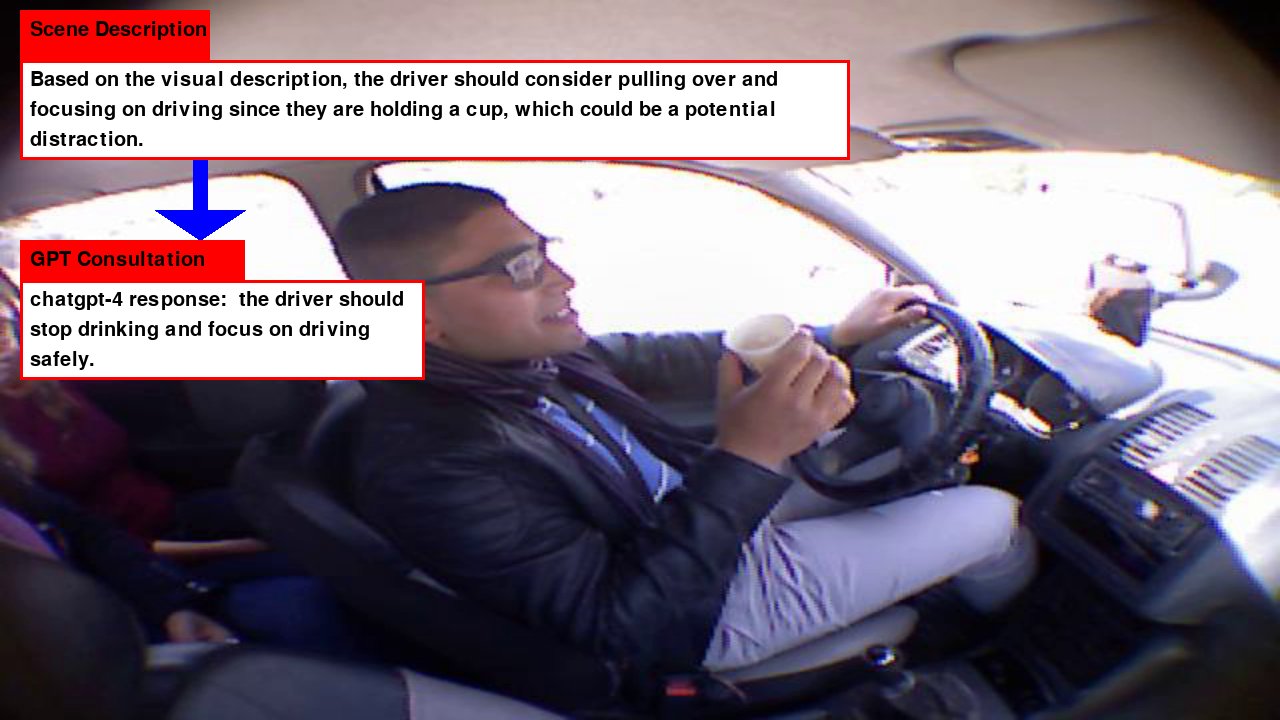}

\subsectionimage{Action-Oriented}{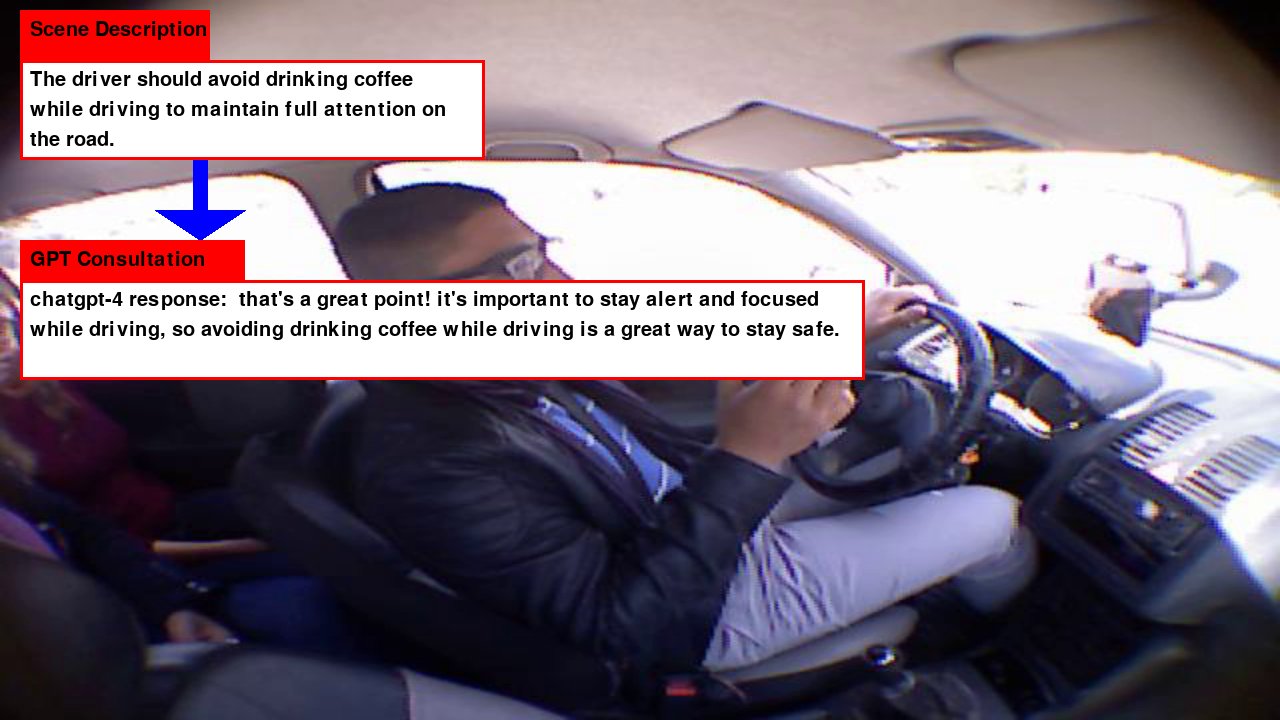}

\subsectionimage{Ontological}{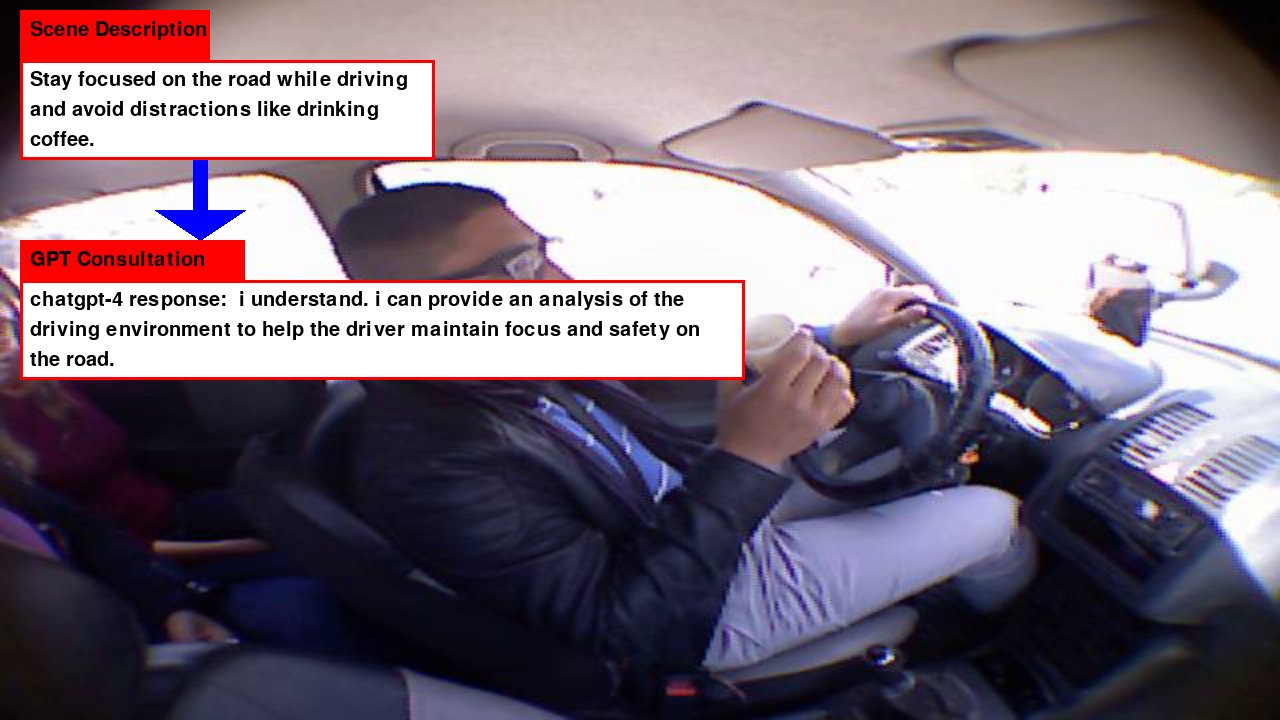}
%%%%%%%%%%%%%%%%%%%%%%%%%%%%%%%%%%%%%%%%%%%%%%%%%
\subsubsection{Case 4: looking down during driving}

\paragraph{Visual Prompts}

\subsectionimage{Focused Description}{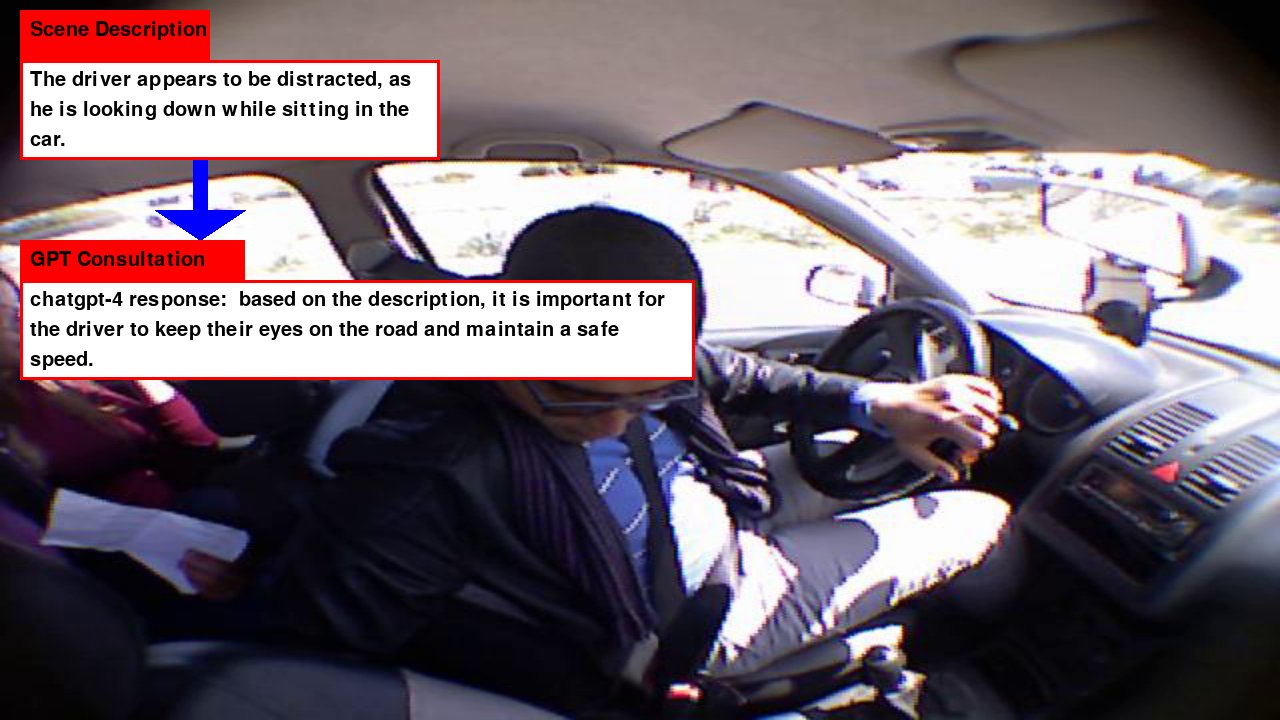}

\subsectionimage{Behavioral Description}{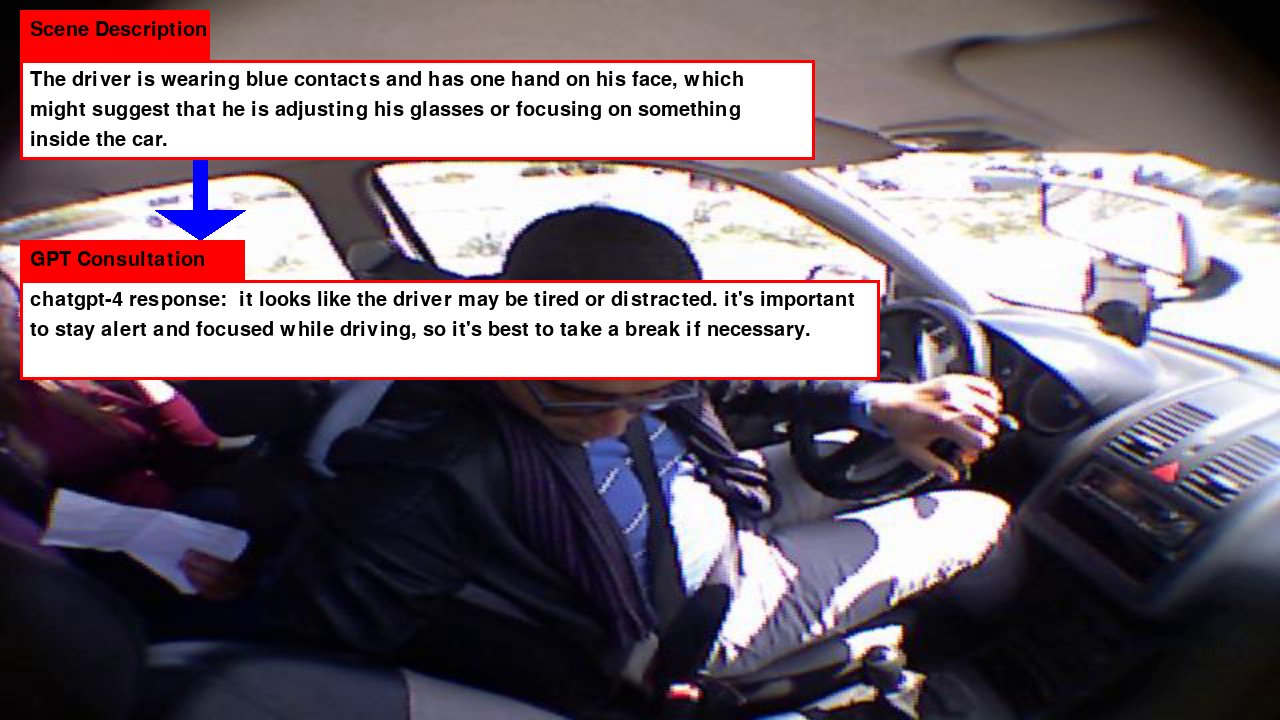}

\subsectionimage{Ontological}{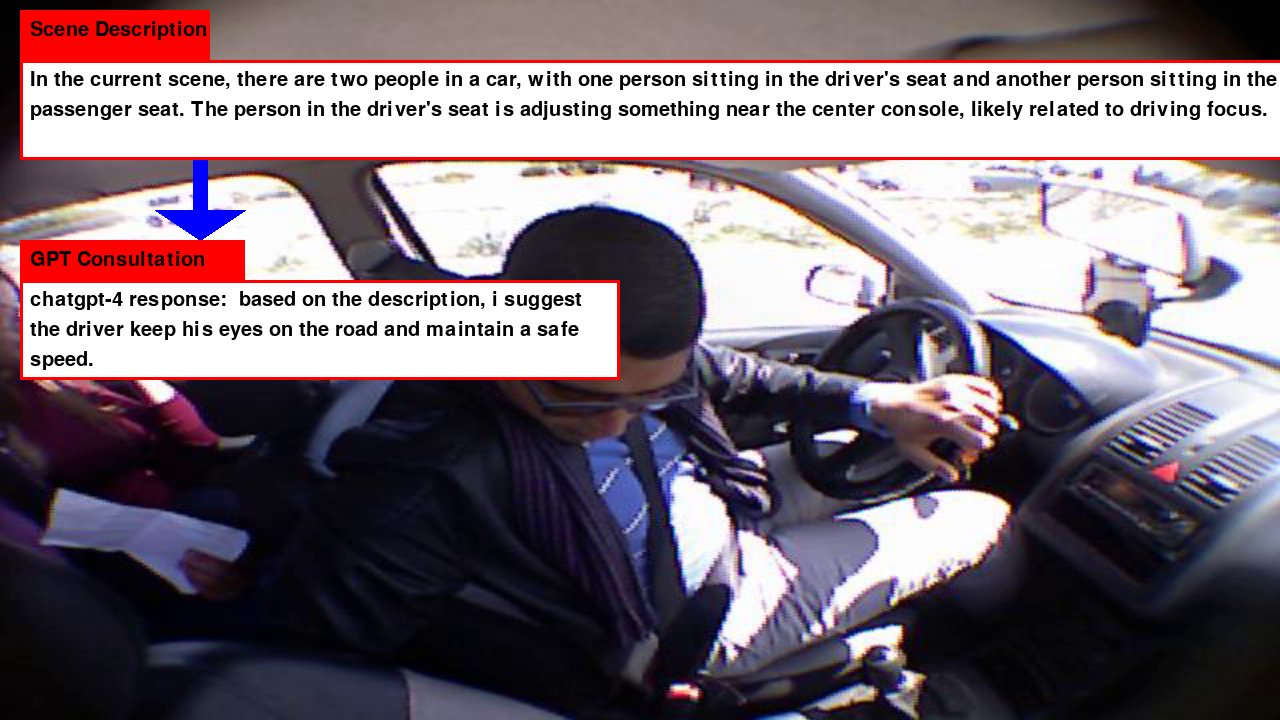}

\paragraph{LLM Prompts}

\subsectionimage{Consultative}{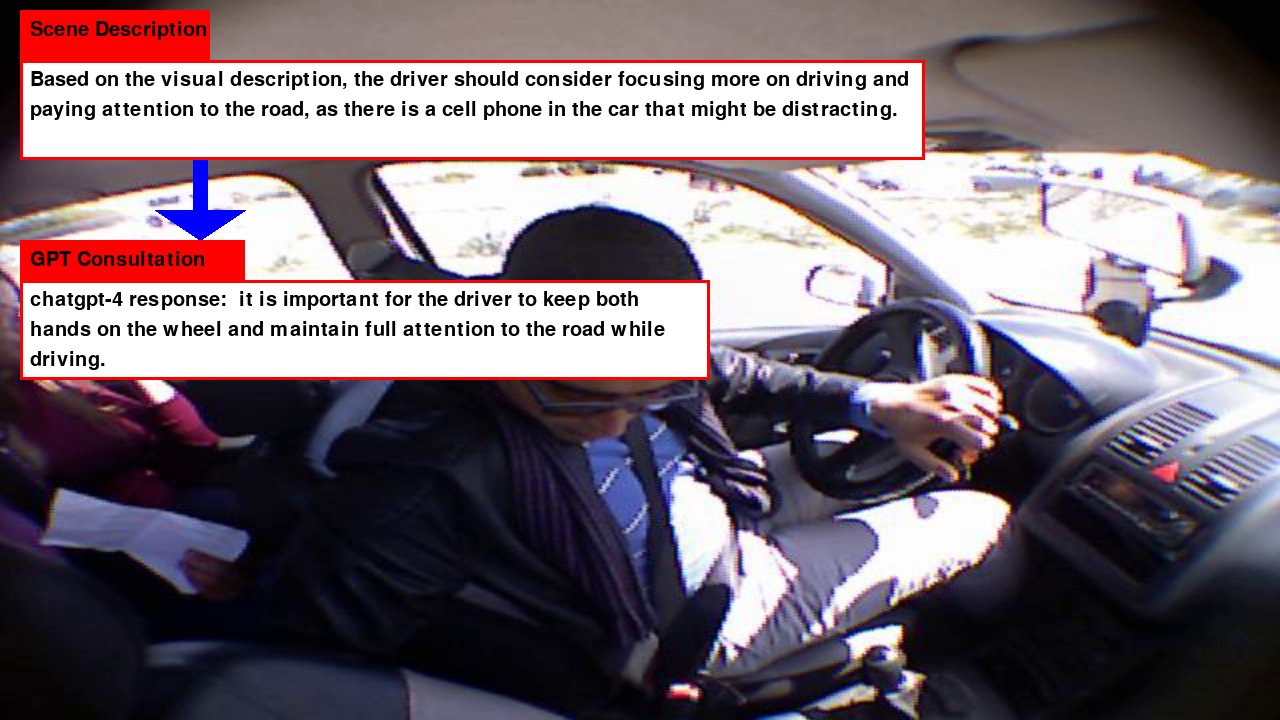}

\subsectionimage{Action-Oriented}{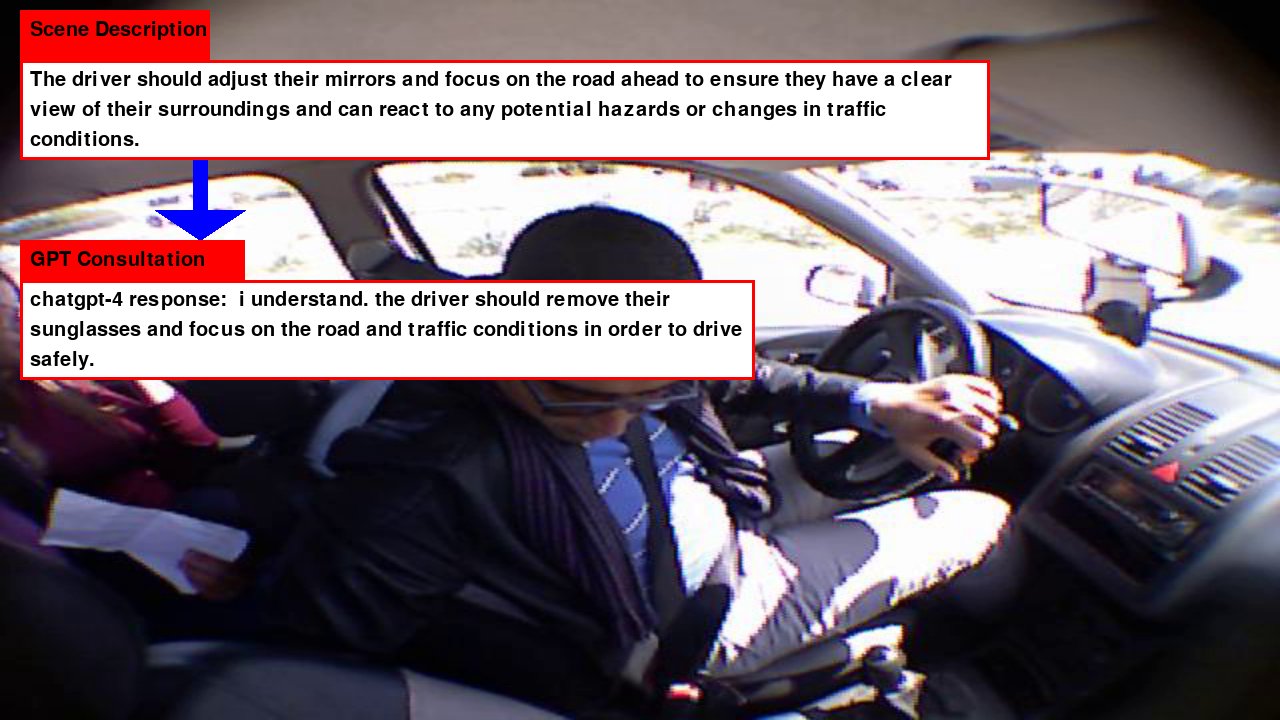}

\subsectionimage{Ontological}{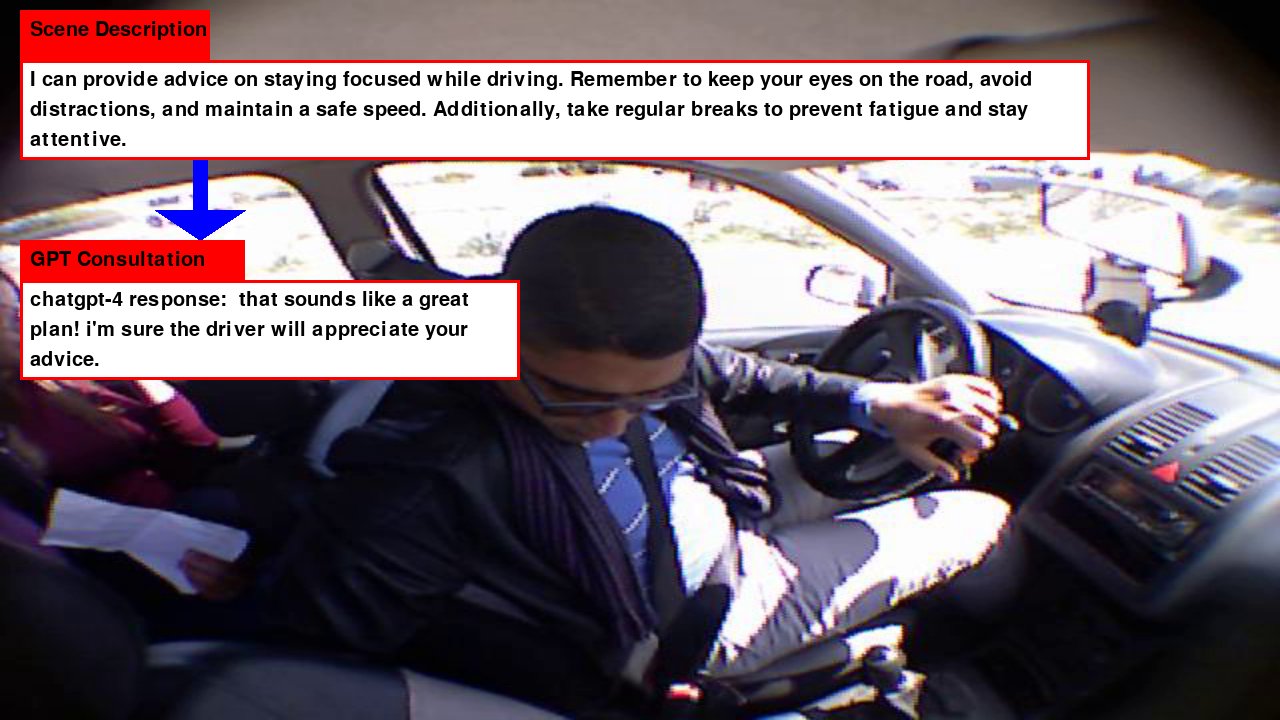}
%%%%%%%%%%%%%%%%%%%%%%%%%%%%%%%%%%%%%%%%%%%%%%%%%
\subsubsection{Case 5: looking to passenger during driving}

\paragraph{Visual Prompts}

\subsectionimage{Focused Description}{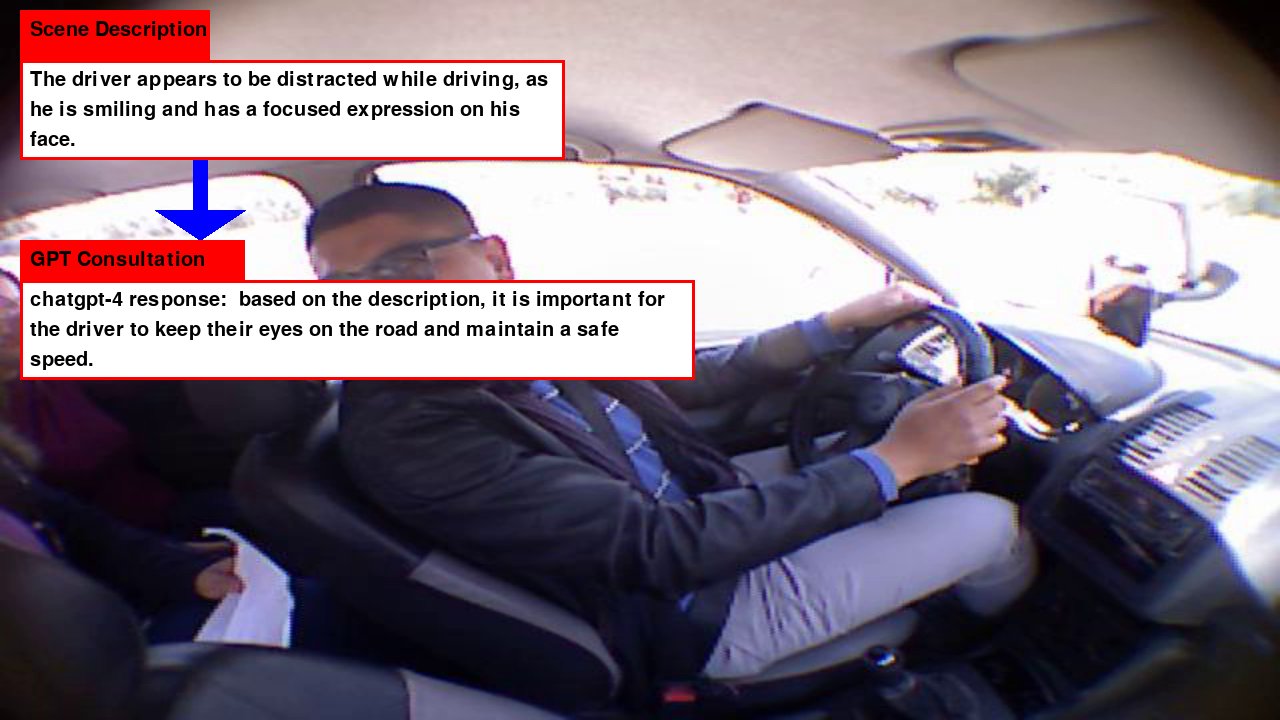}

\subsectionimage{Behavioral Description}{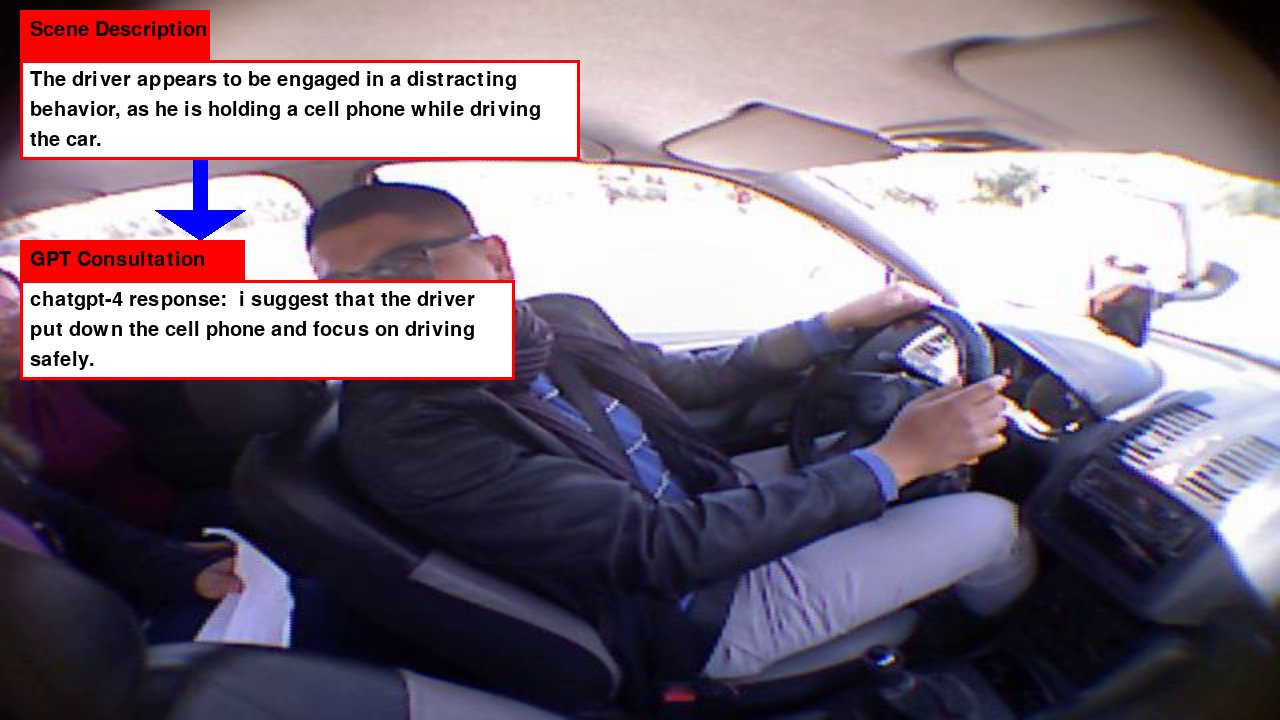}

\subsectionimage{Ontological}{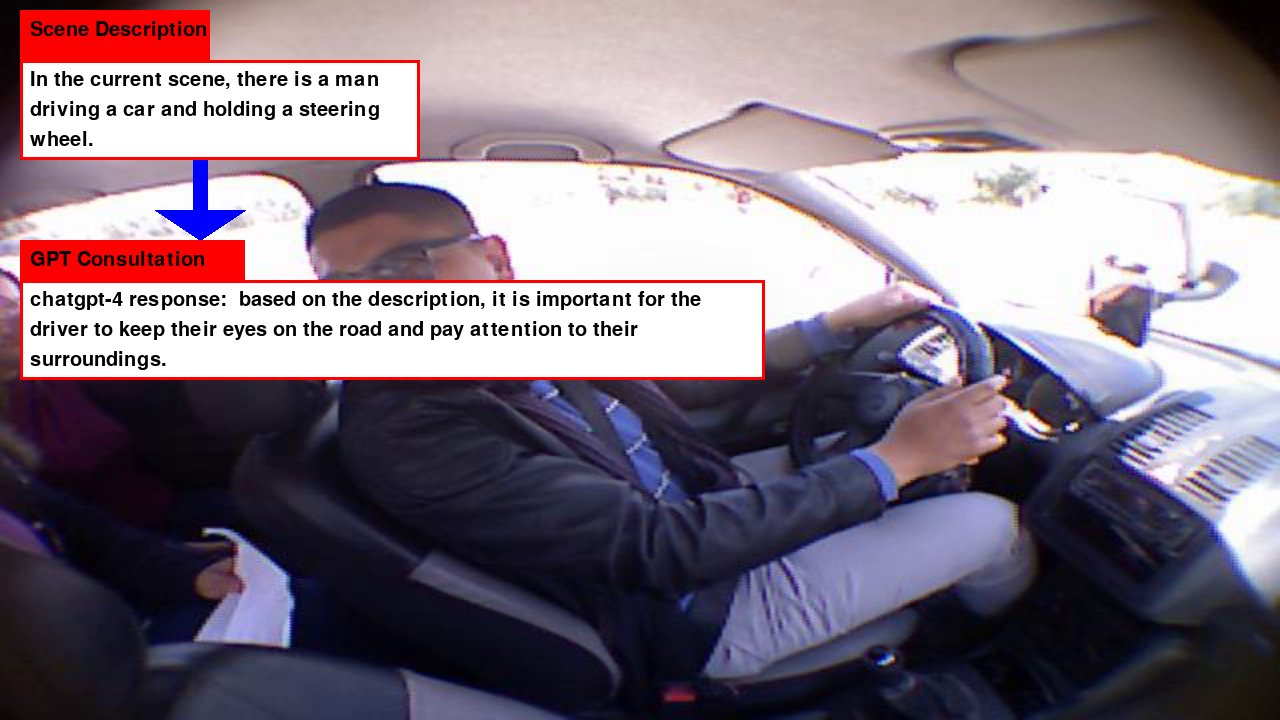}

\paragraph{LLM Prompts}

\subsectionimage{Consultative}{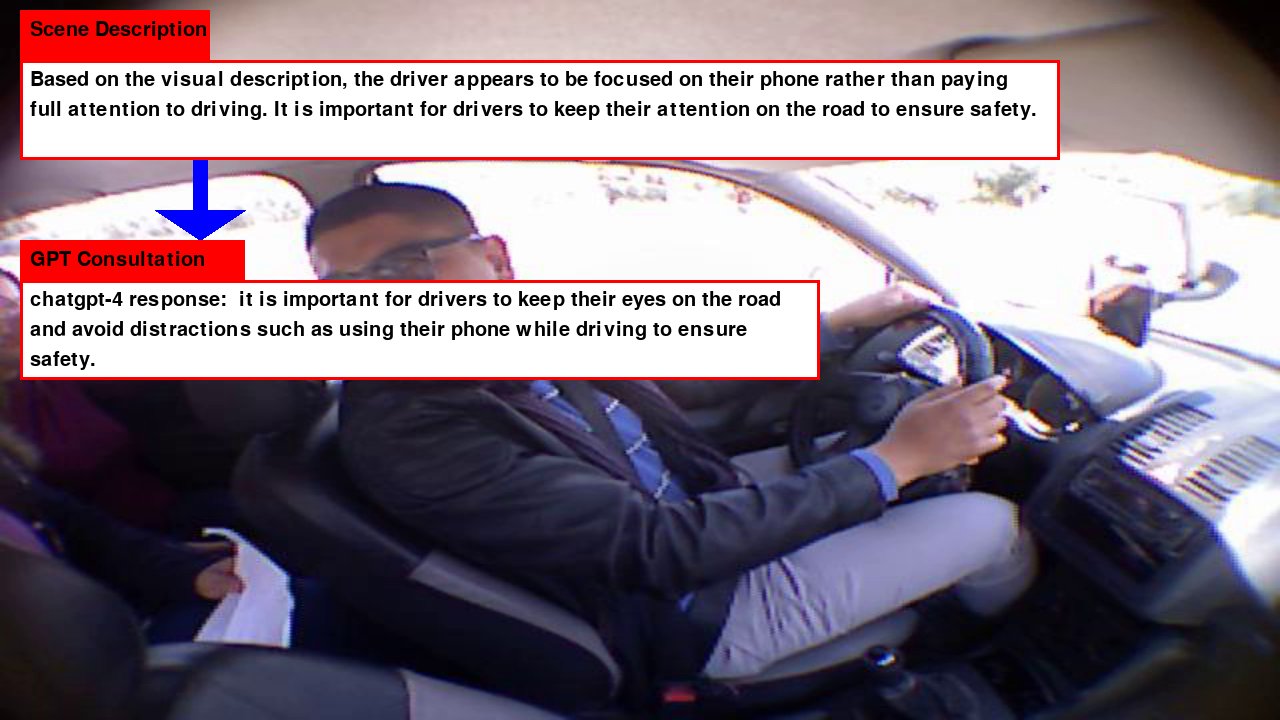}

\subsectionimage{Action-Oriented}{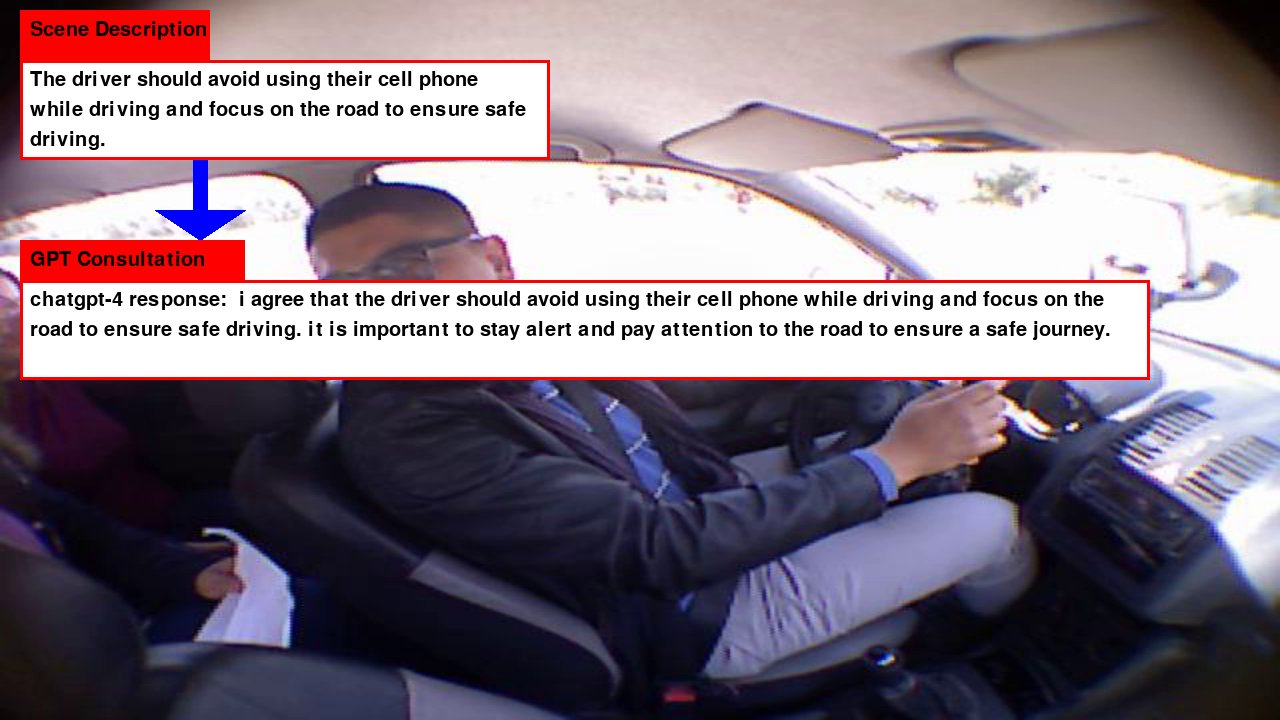}

\subsectionimage{Ontological}{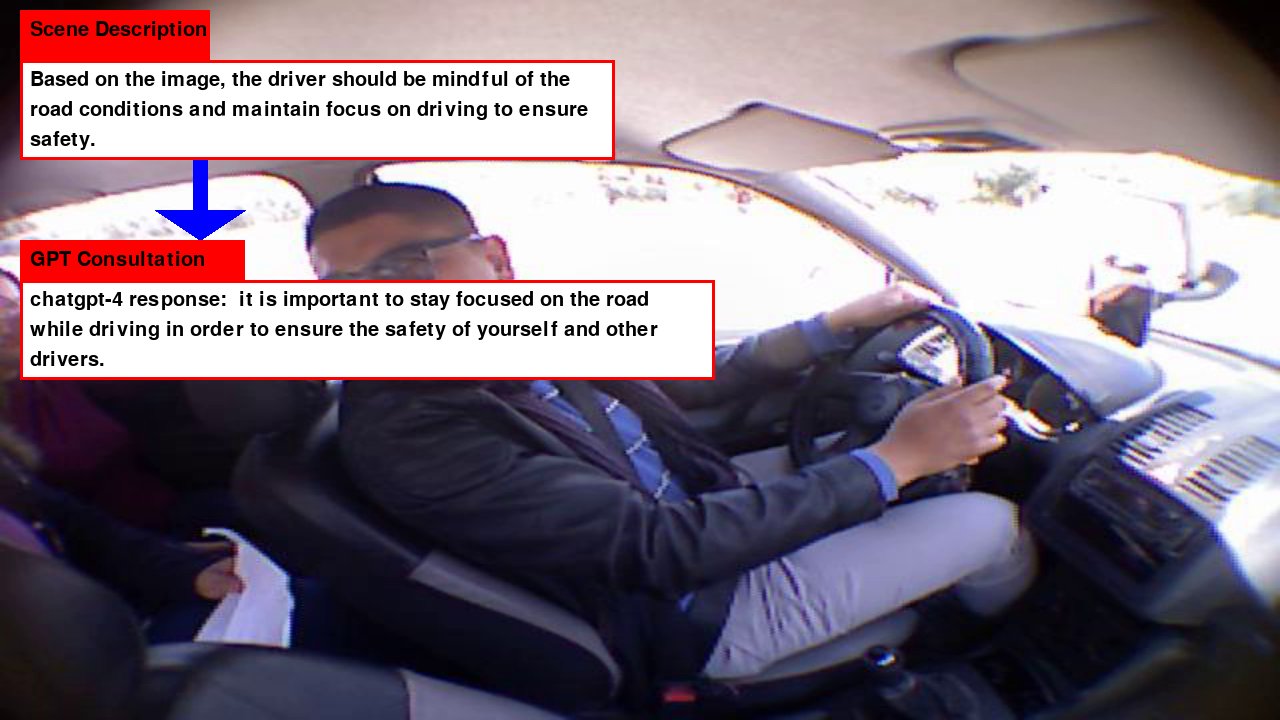}
%%%%%%%%%%%%%%%%%%%%%%%%%%%%%%%%%%%%%%%%%%%%%%%%%
\subsubsection{Case 6: using radio during driving}

\paragraph{Visual Prompts}

\subsectionimage{Focused Description}{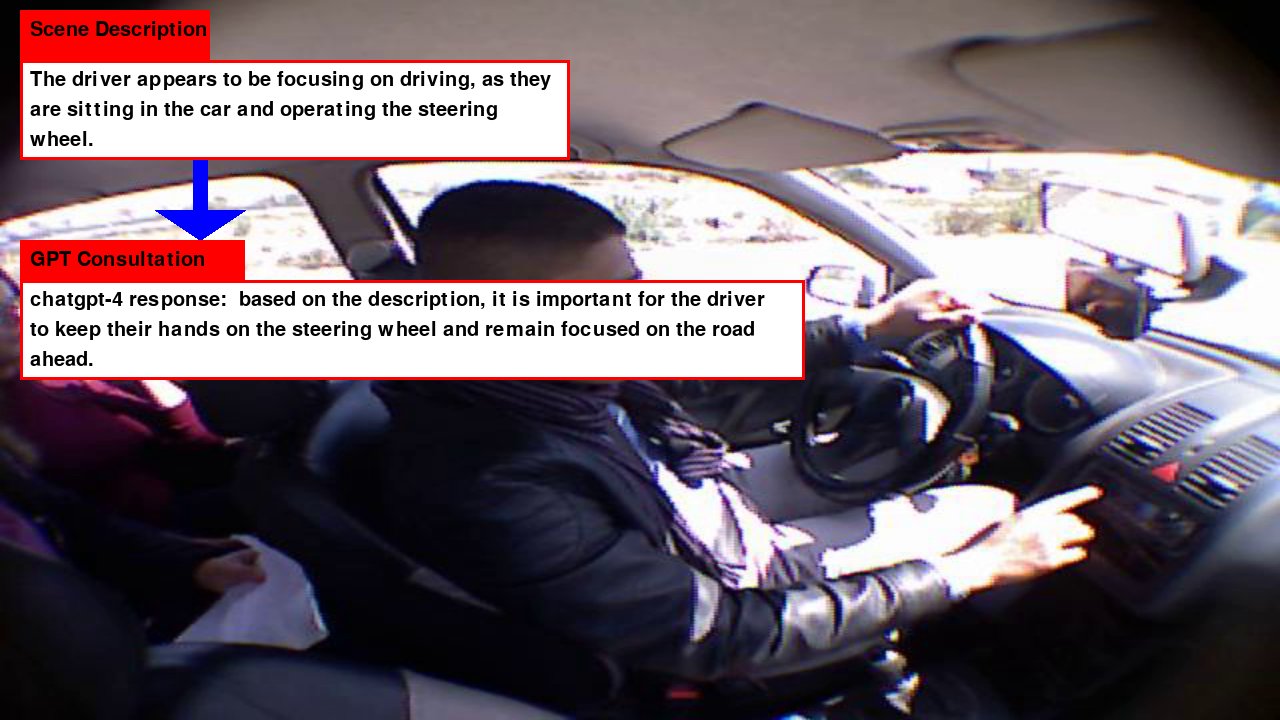}

\subsectionimage{Behavioral Description}{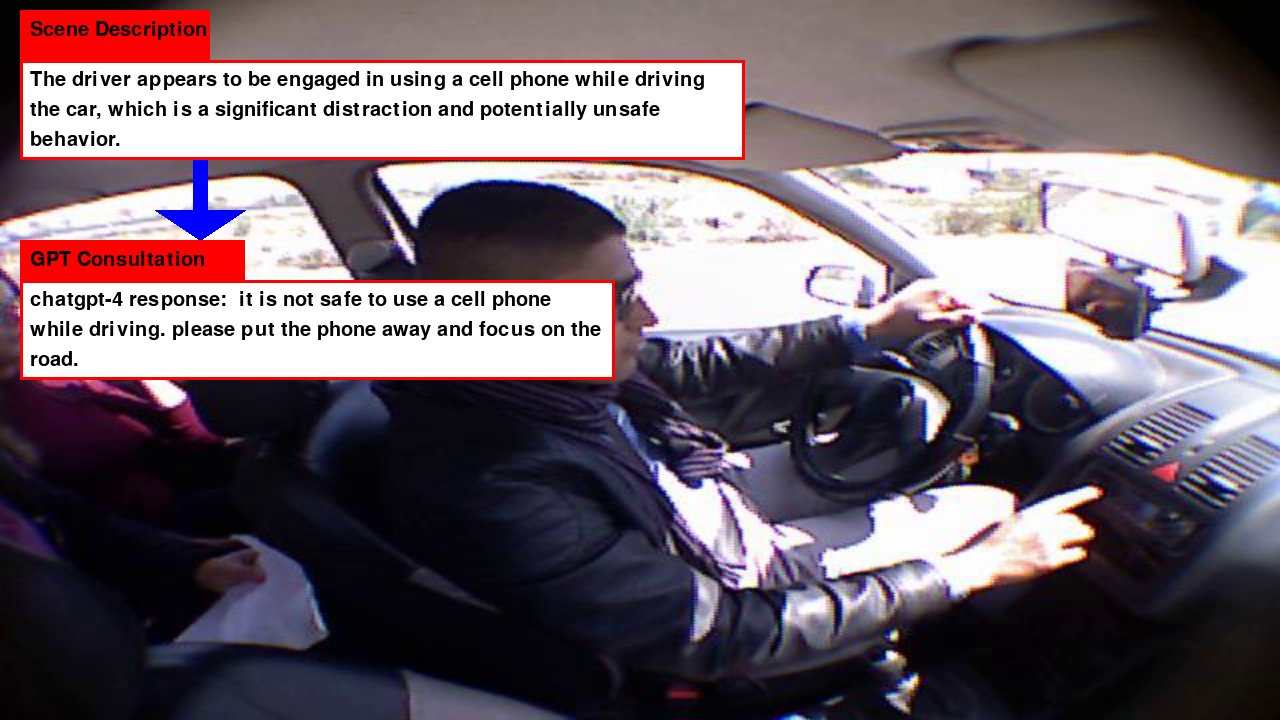}

\subsectionimage{Ontological}{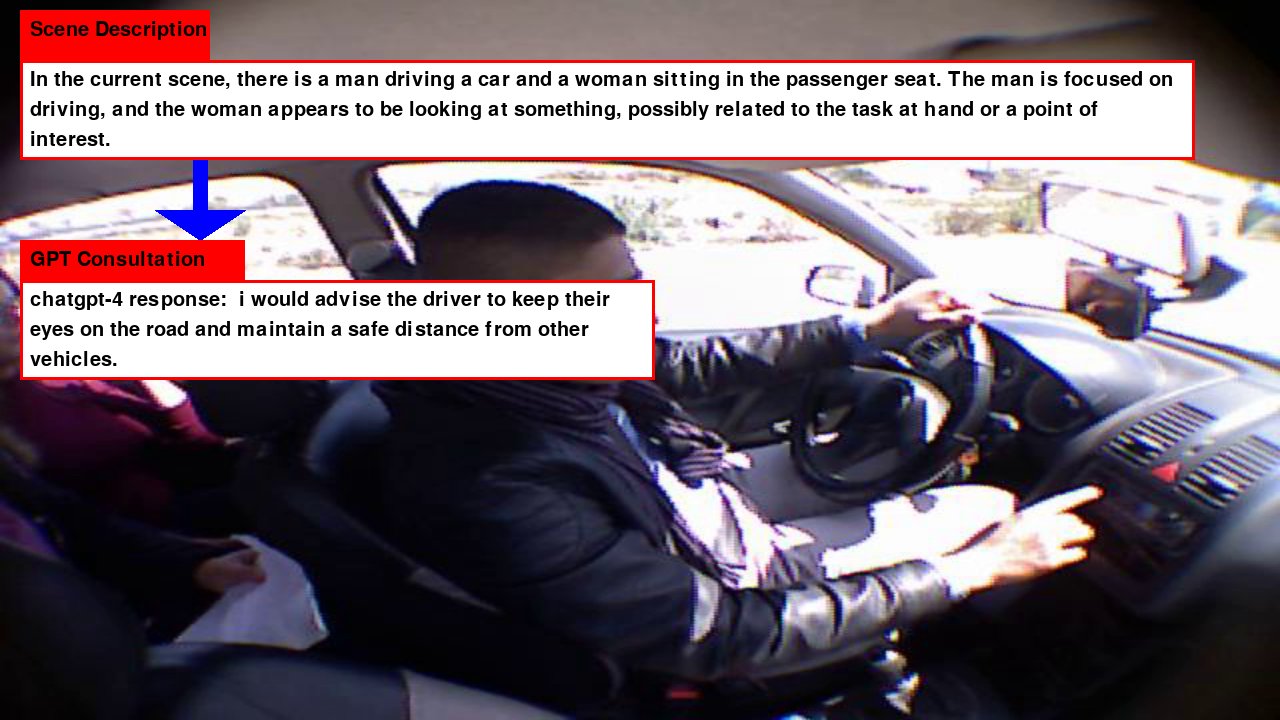}

\paragraph{LLM Prompts}

\subsectionimage{Consultative}{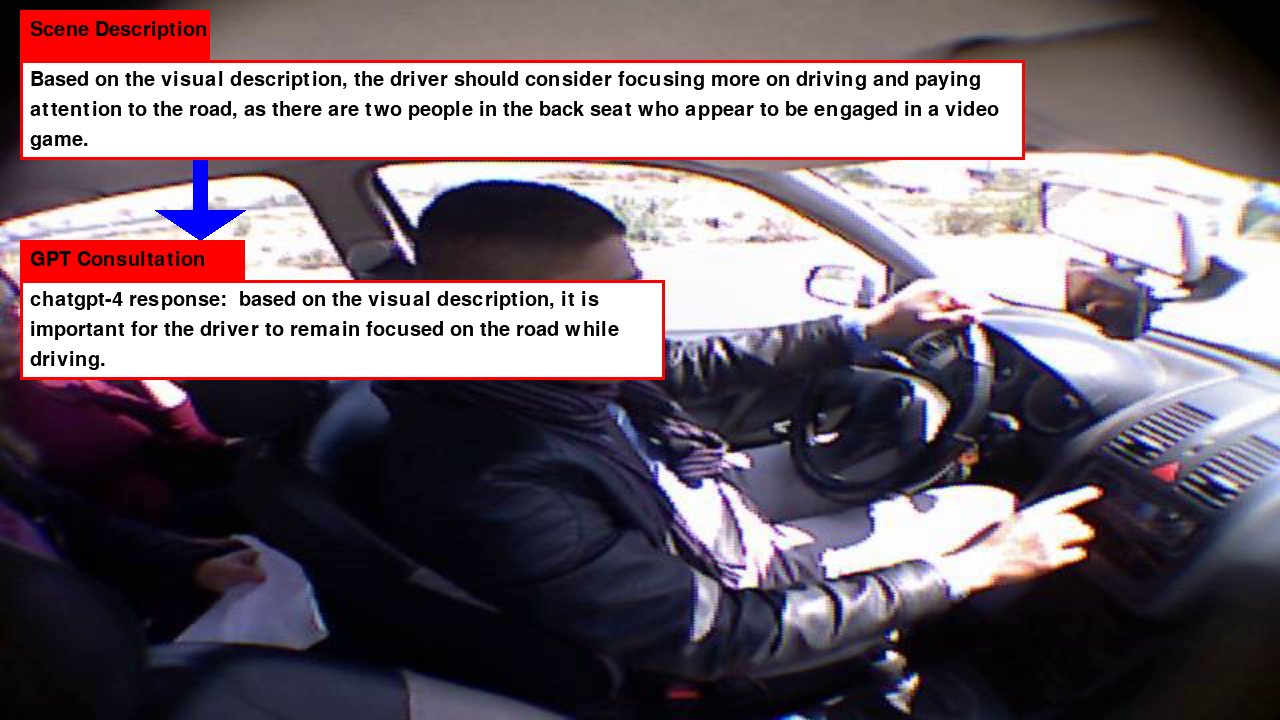}

\subsectionimage{Action-Oriented}{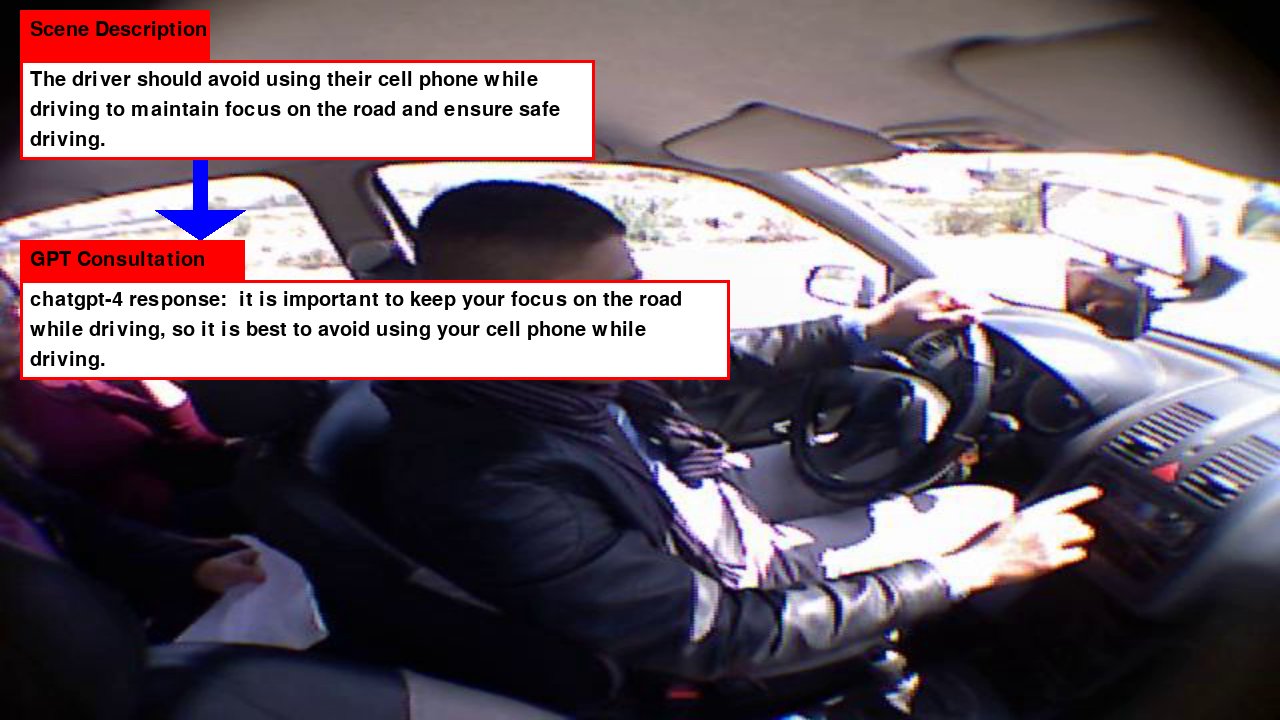}

\subsectionimage{Ontological}{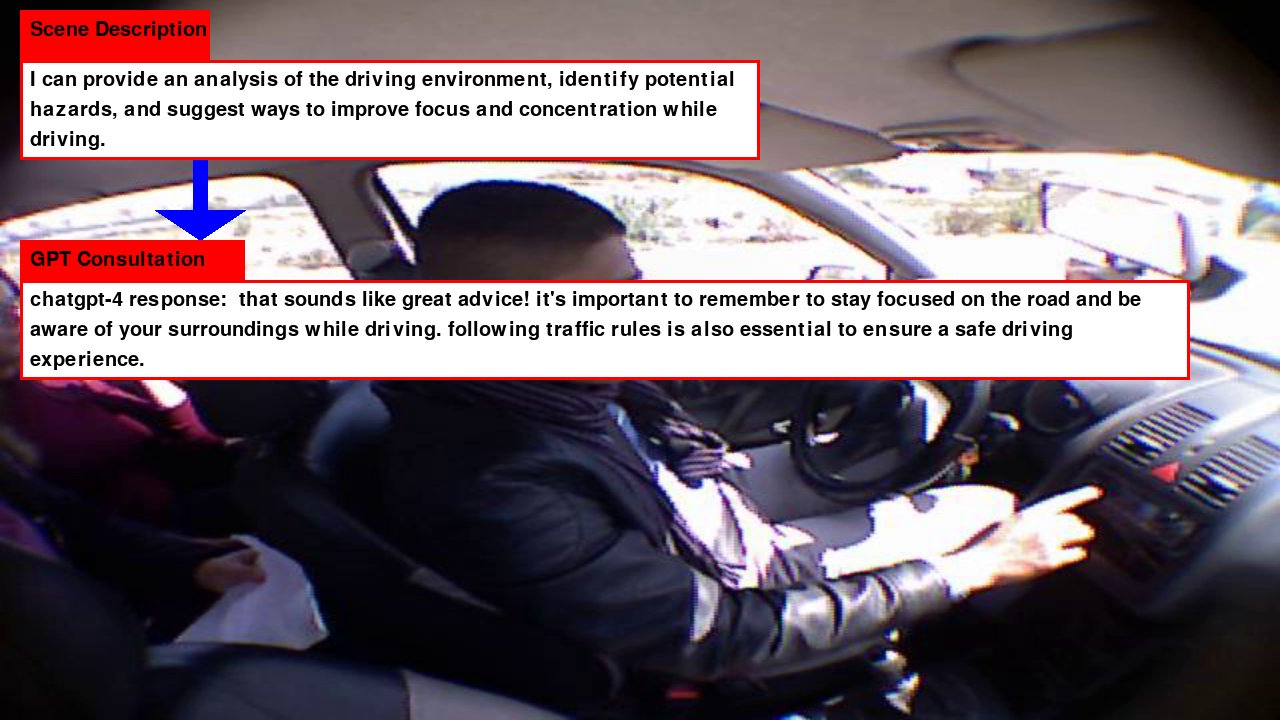}
%%%%%%%%%%%%%%%%%%%%%%%%%%%%%%%%%%%%%%%%%%%%%%%%%
\subsubsection{Case 7: using mobile during driving}

\paragraph{Visual Prompts}

\subsectionimage{Focused Description}{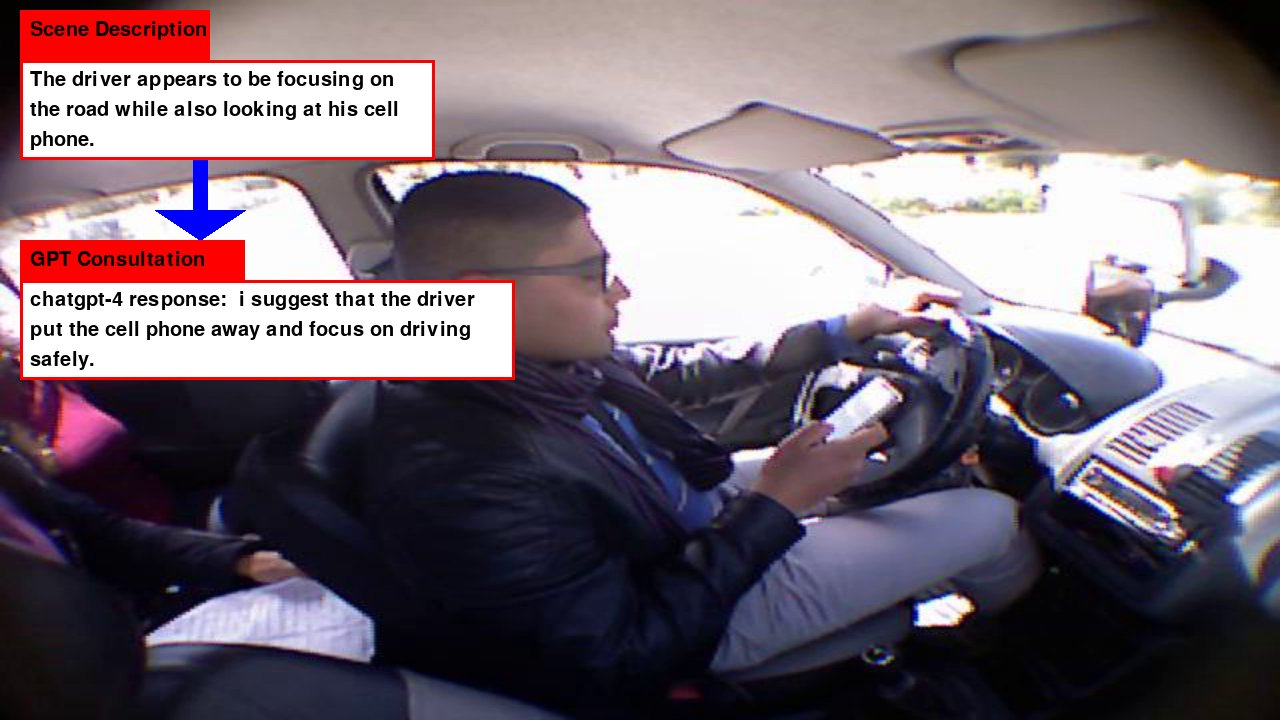}

\subsectionimage{Behavioral Description}{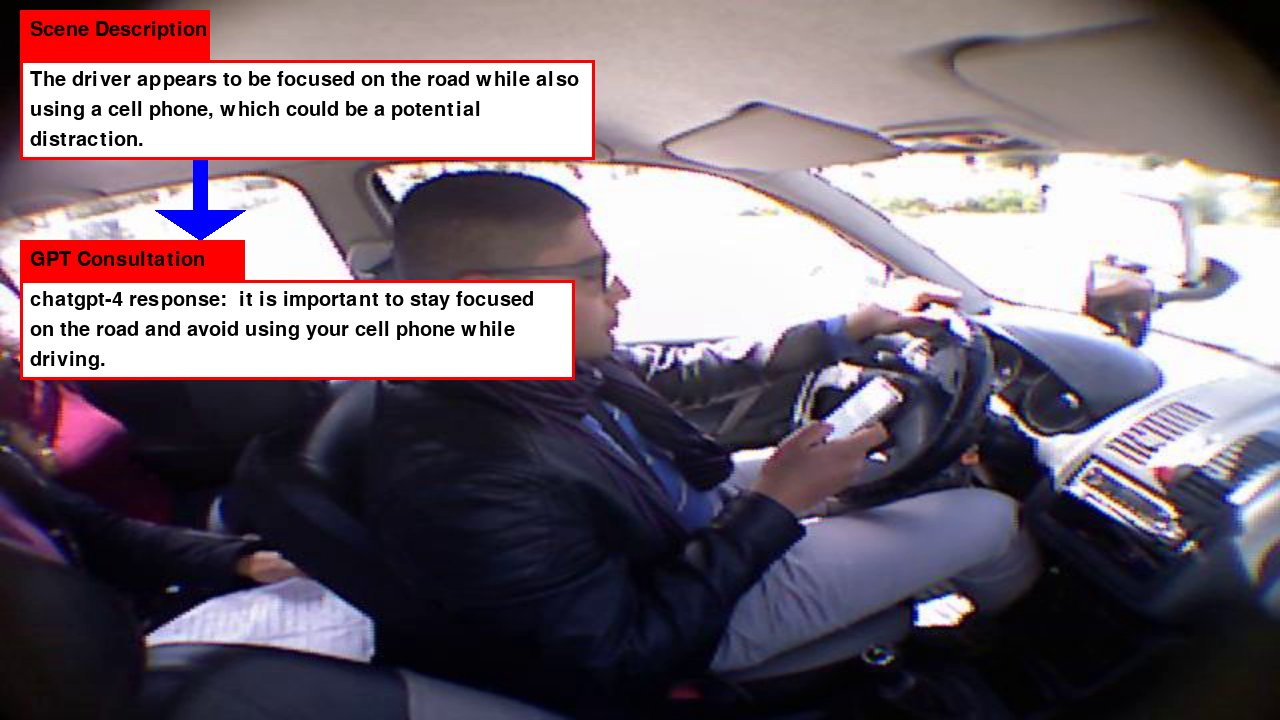}

\subsectionimage{Ontological}{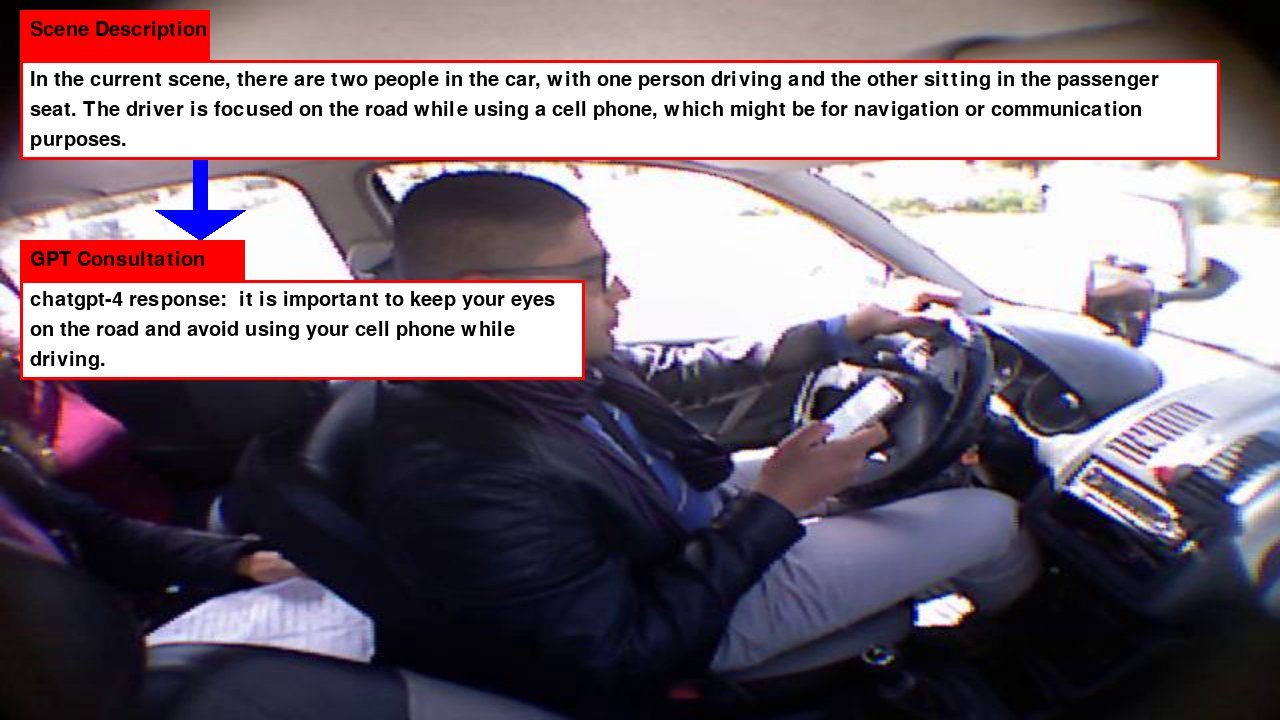}

\paragraph{LLM Prompts}

\subsectionimage{Consultative}{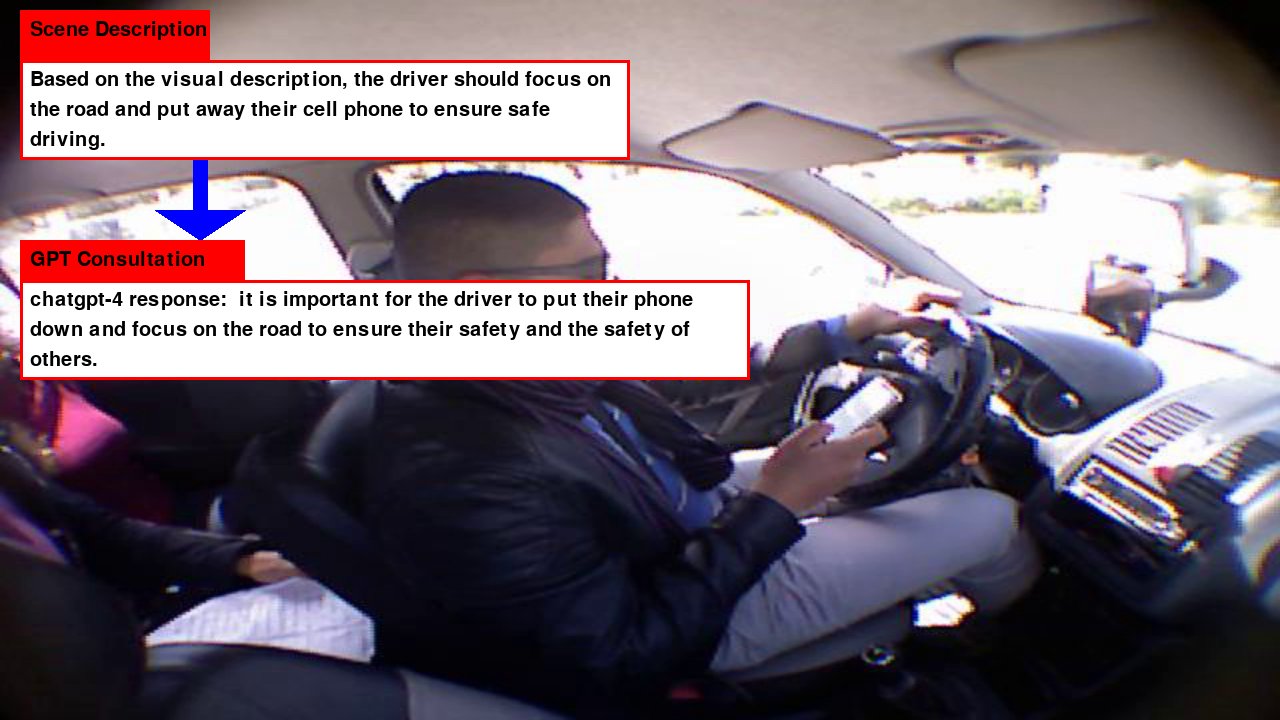}

\subsectionimage{Action-Oriented}{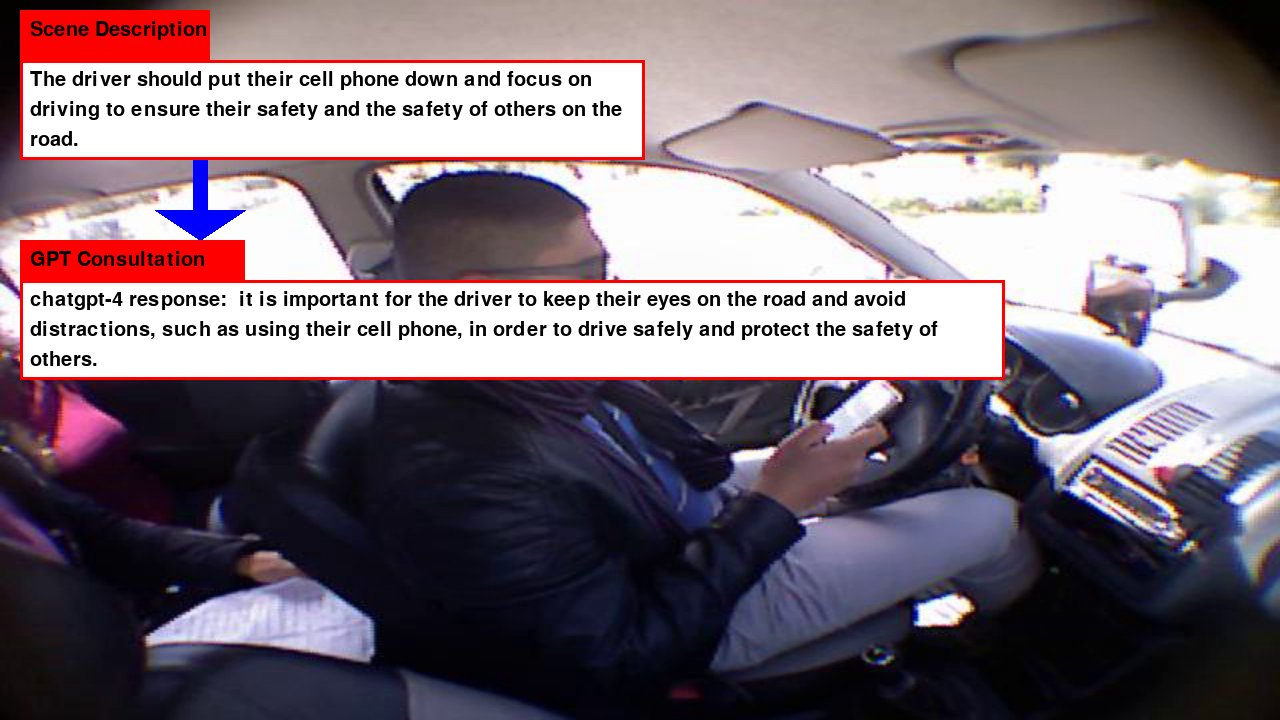}

\subsectionimage{Ontological}{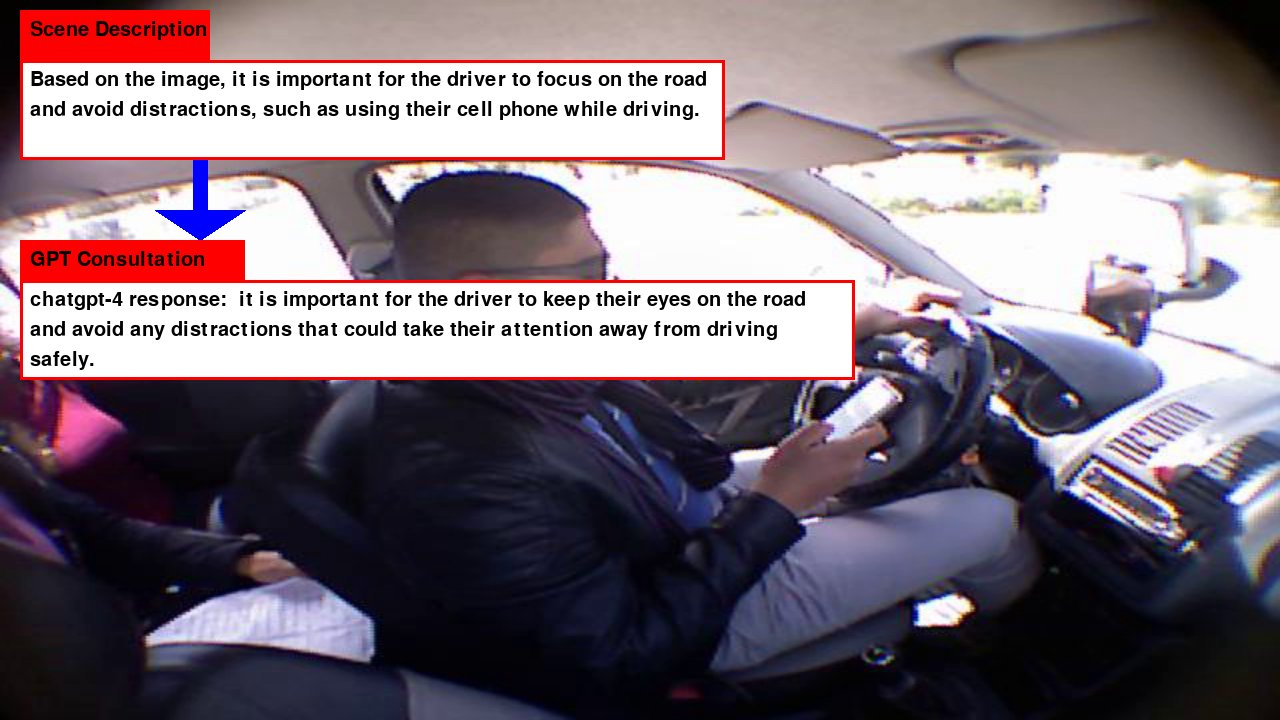}
%%%%%%%%%%%%%%%%%%%%%%%%%%%%%%%%%%%%%%%%%%%%%%%%%
\subsubsection{Case 8: sleeping during driving}

\paragraph{Visual Prompts}

\subsectionimage{Focused Description}{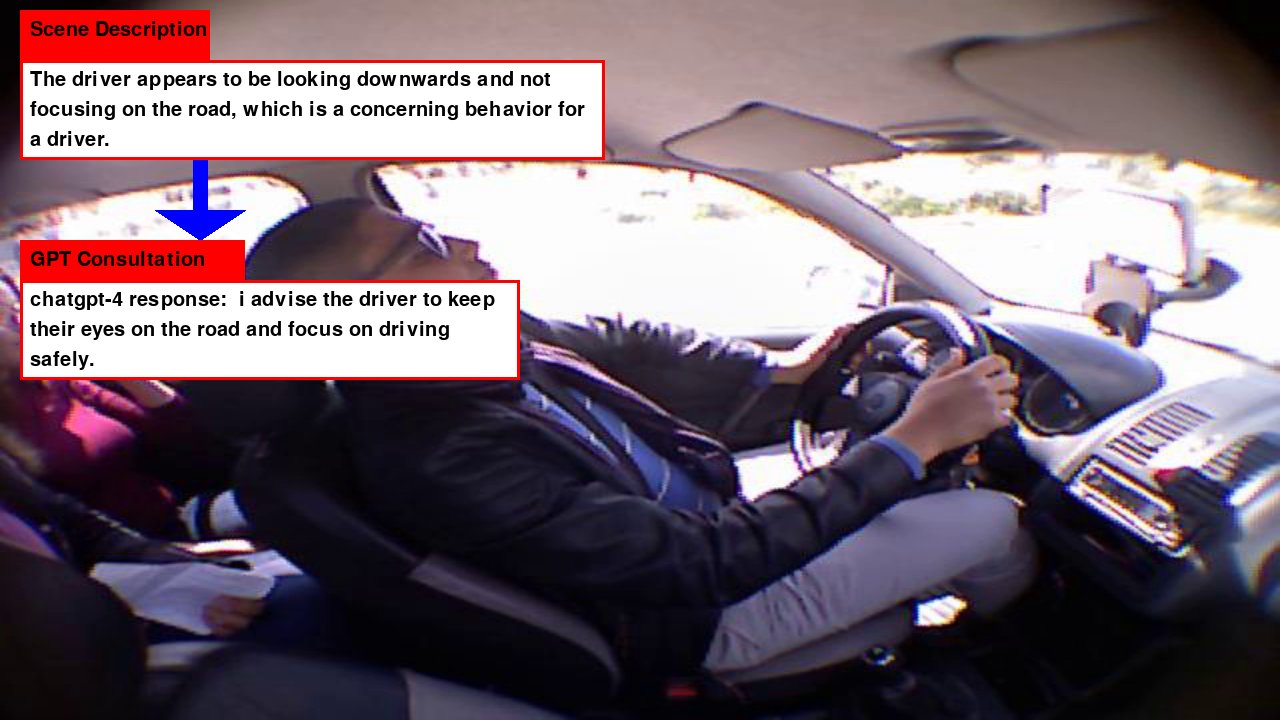}

\subsectionimage{Behavioral Description}{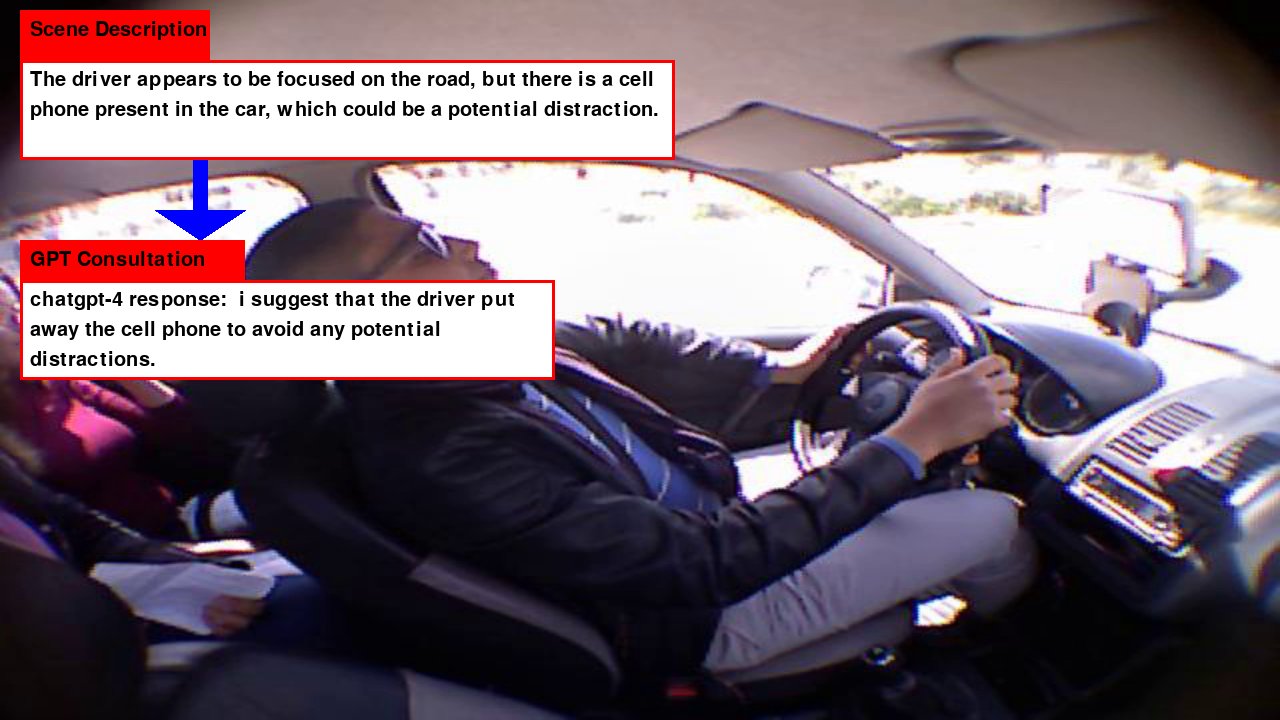}

\subsectionimage{Ontological}{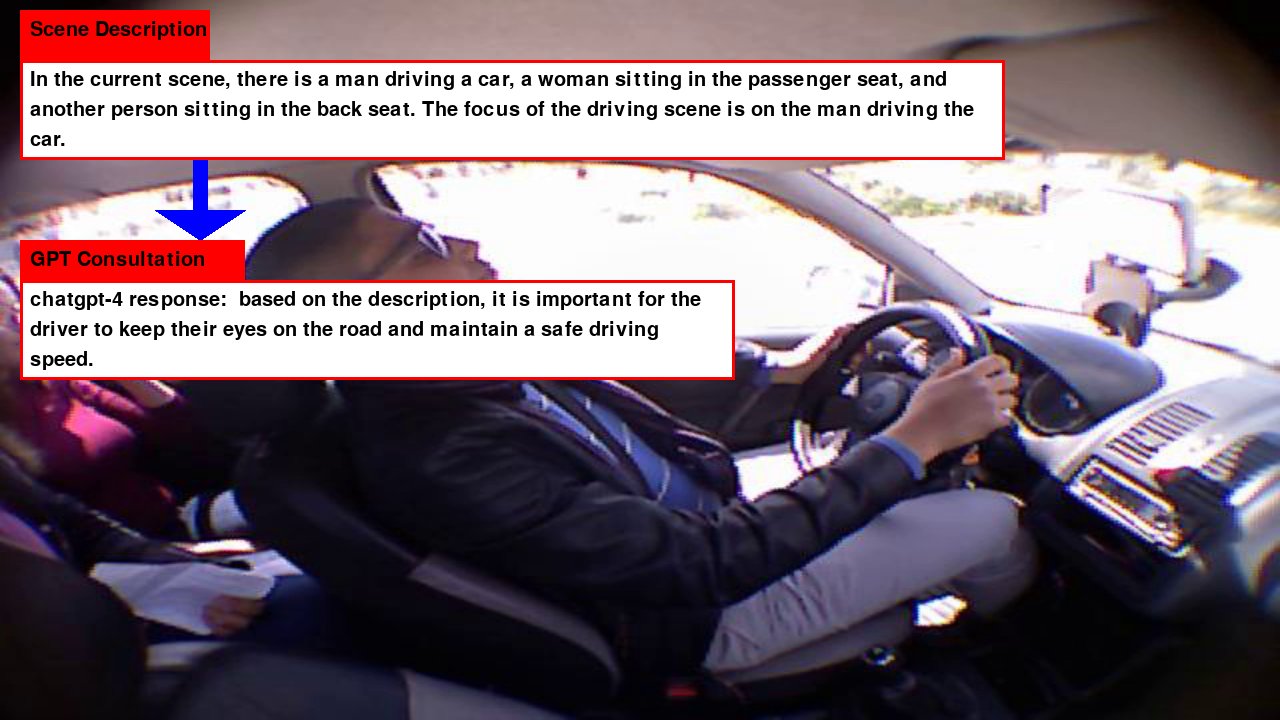}

\paragraph{LLM Prompts}

\subsectionimage{Consultative}{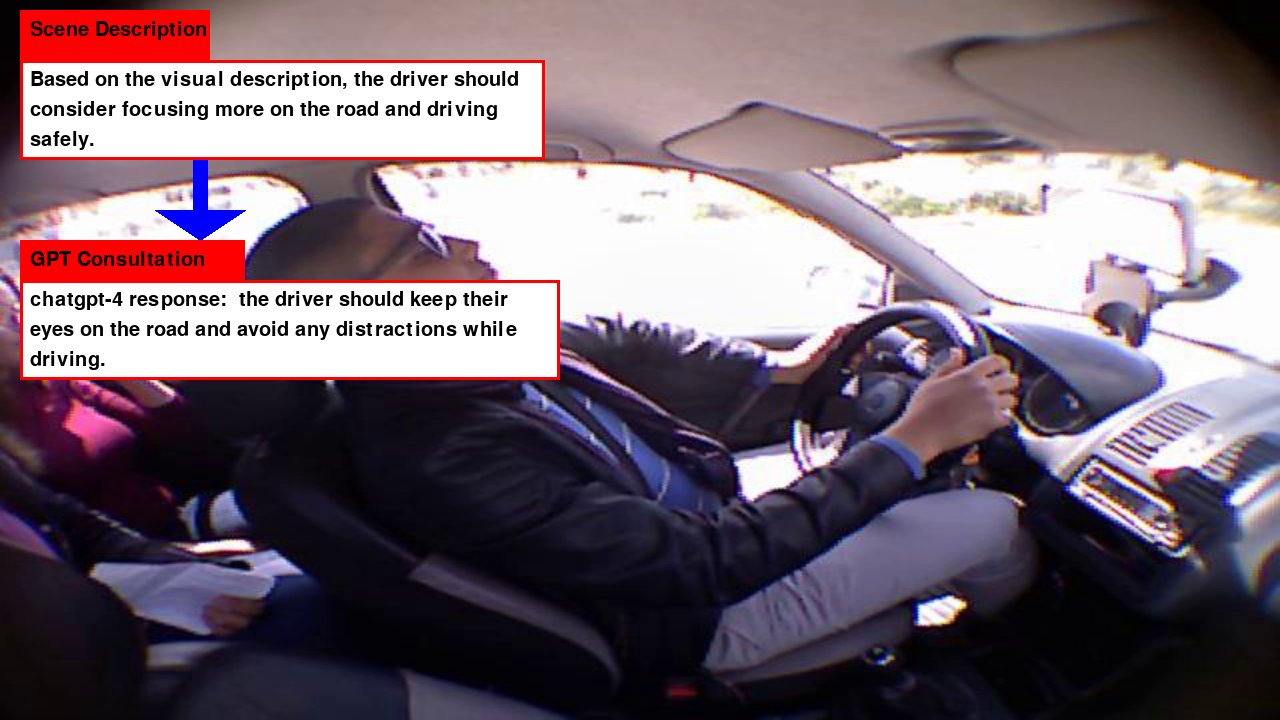}

\subsectionimage{Action-Oriented}{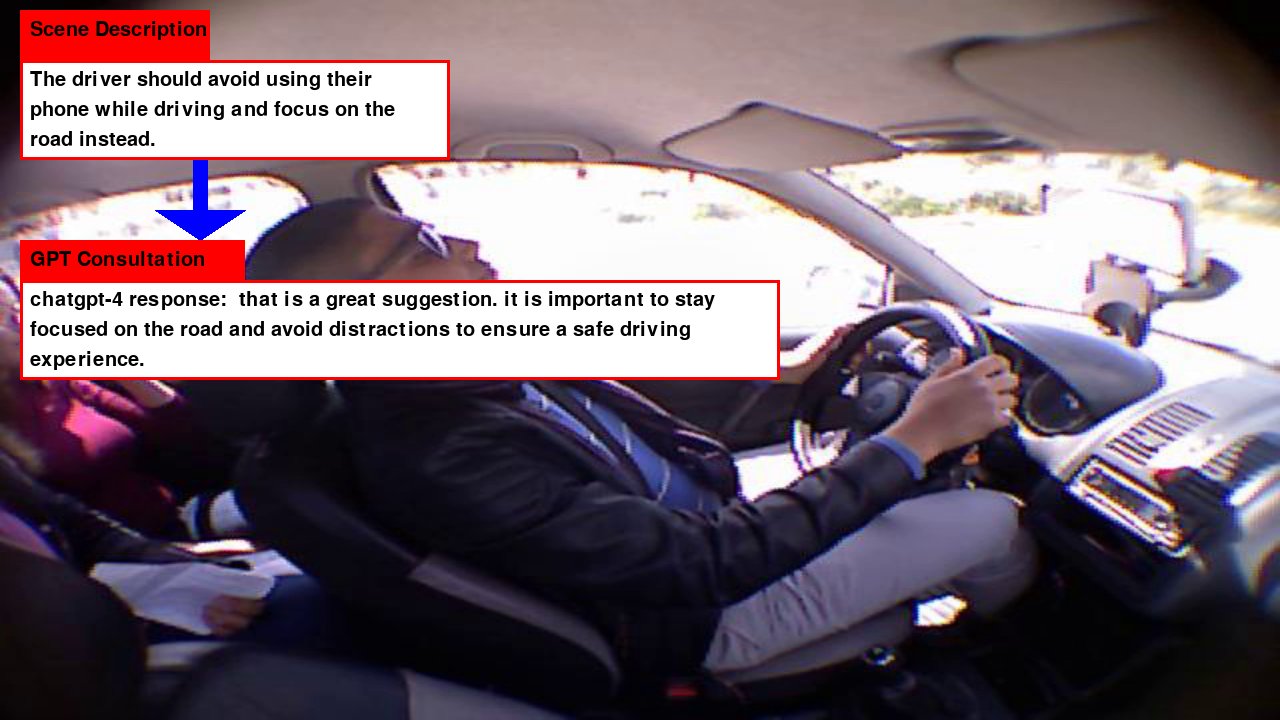}

\subsectionimage{Ontological}{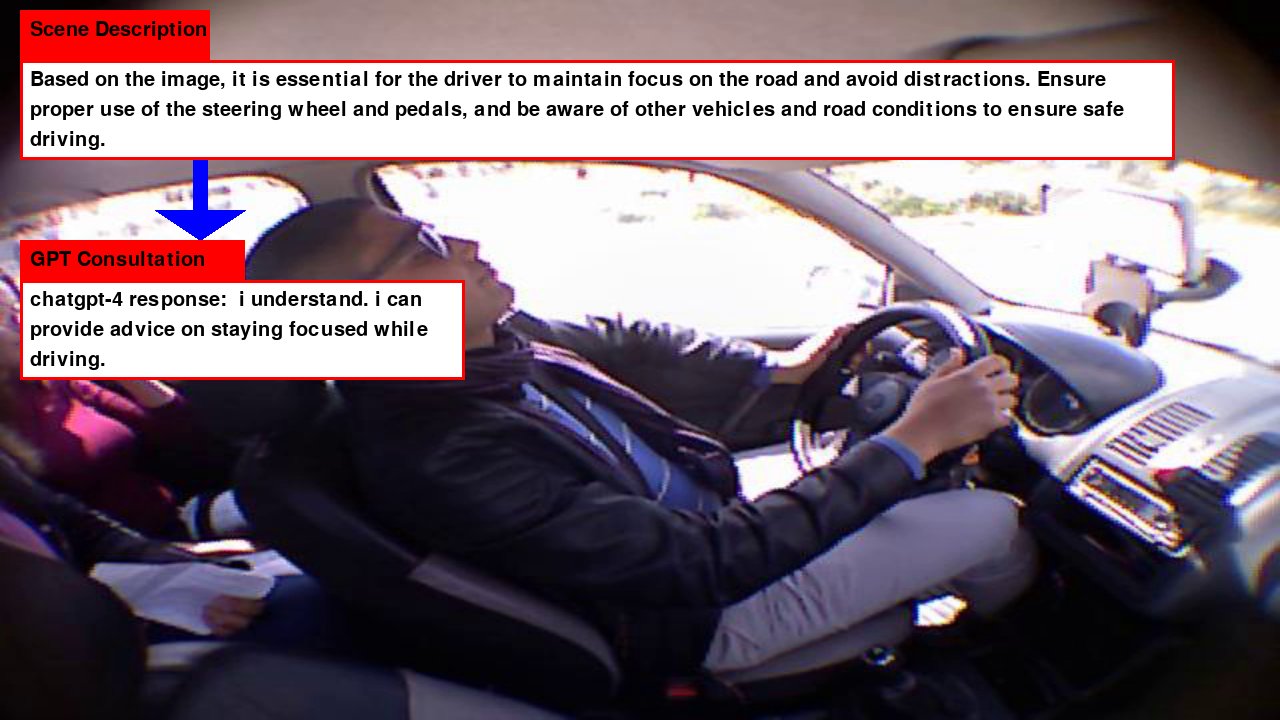}
%%%%%%%%%%%%%%%%%%%%%%%%%%%%%%%%%%%%%%%%%%%%%%%%%
\subsubsection{Case 9: smoking during driving}

\paragraph{Visual Prompts}

\subsectionimage{Focused Description}{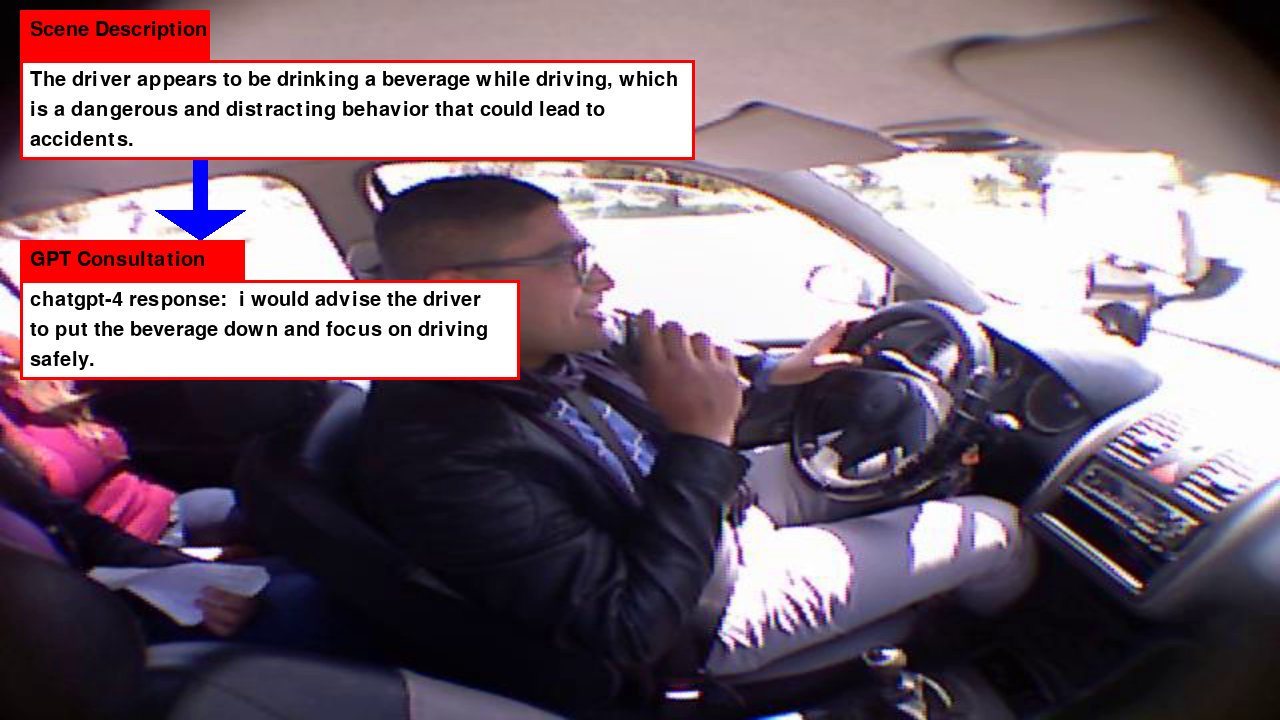}

\subsectionimage{Behavioral Description}{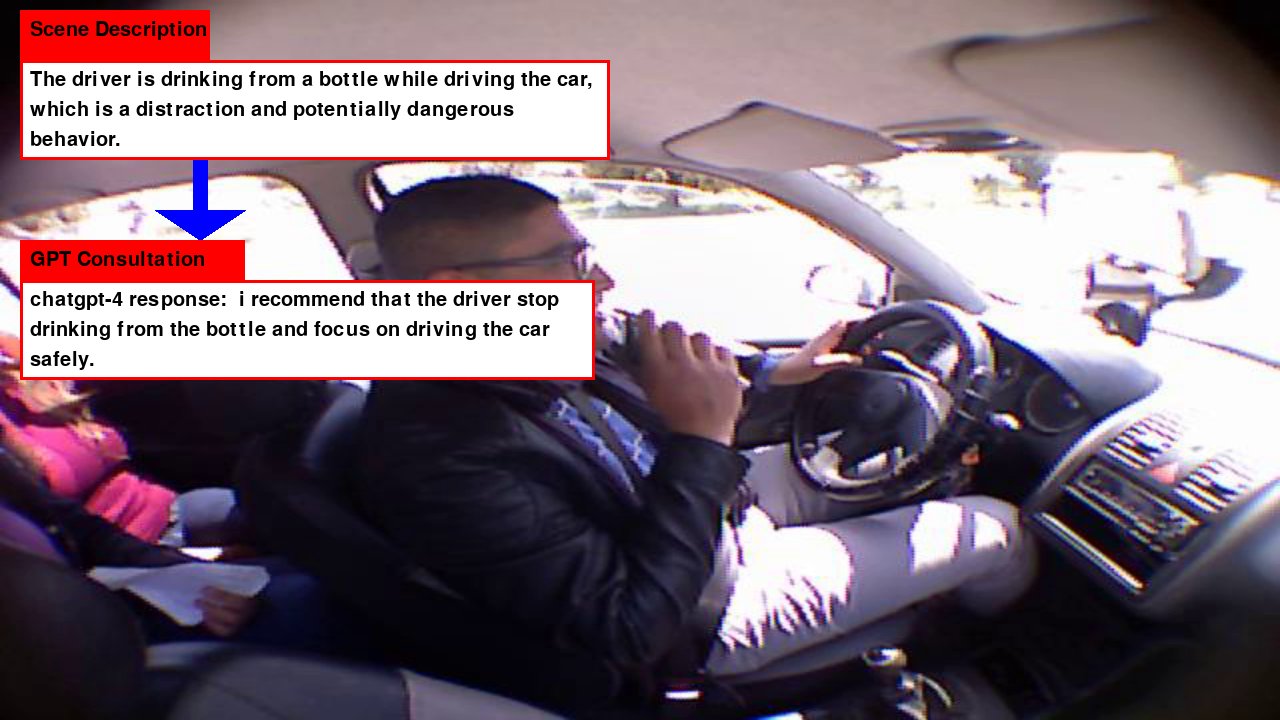}

\subsectionimage{Ontological}{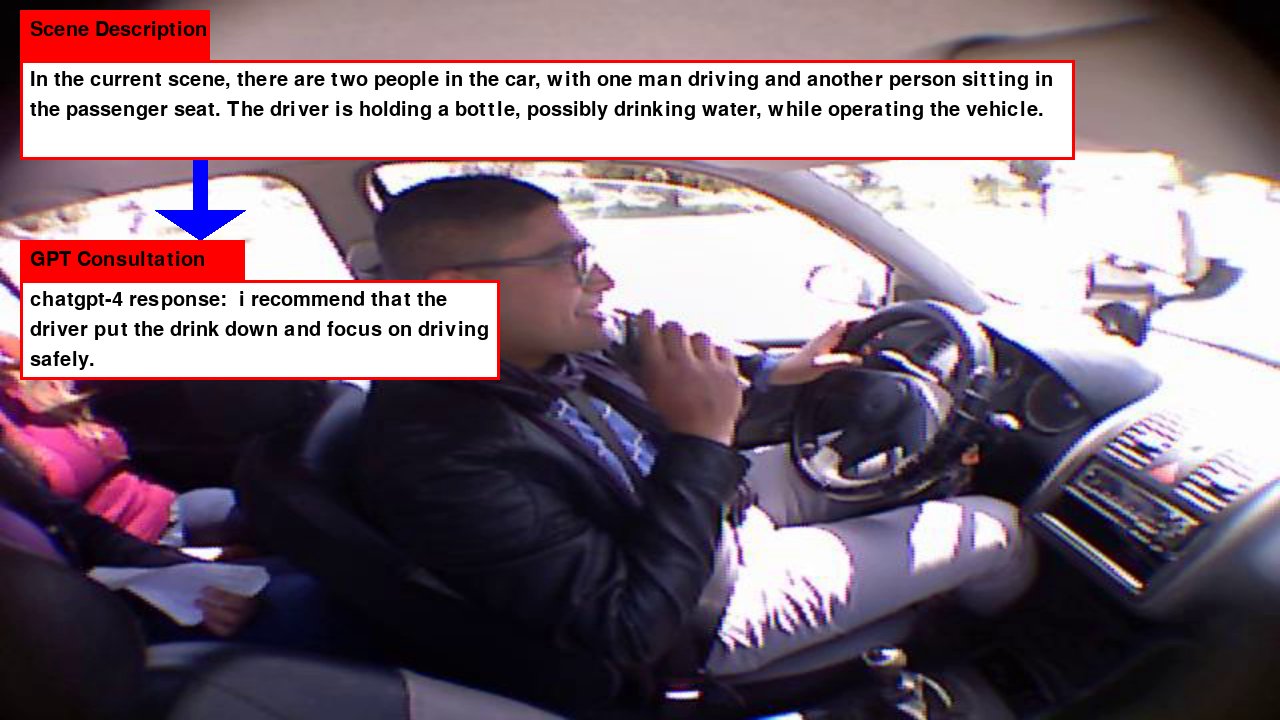}

\paragraph{LLM Prompts}

\subsectionimage{Consultative}{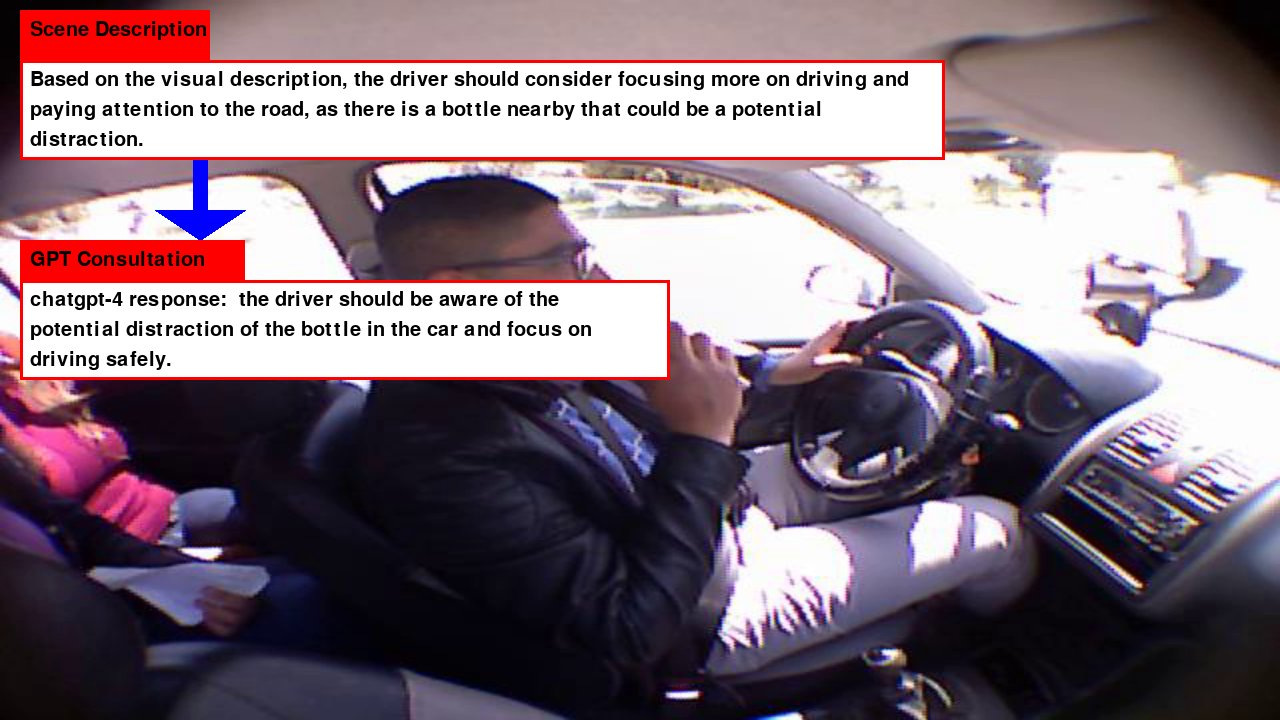}

\subsectionimage{Action-Oriented}{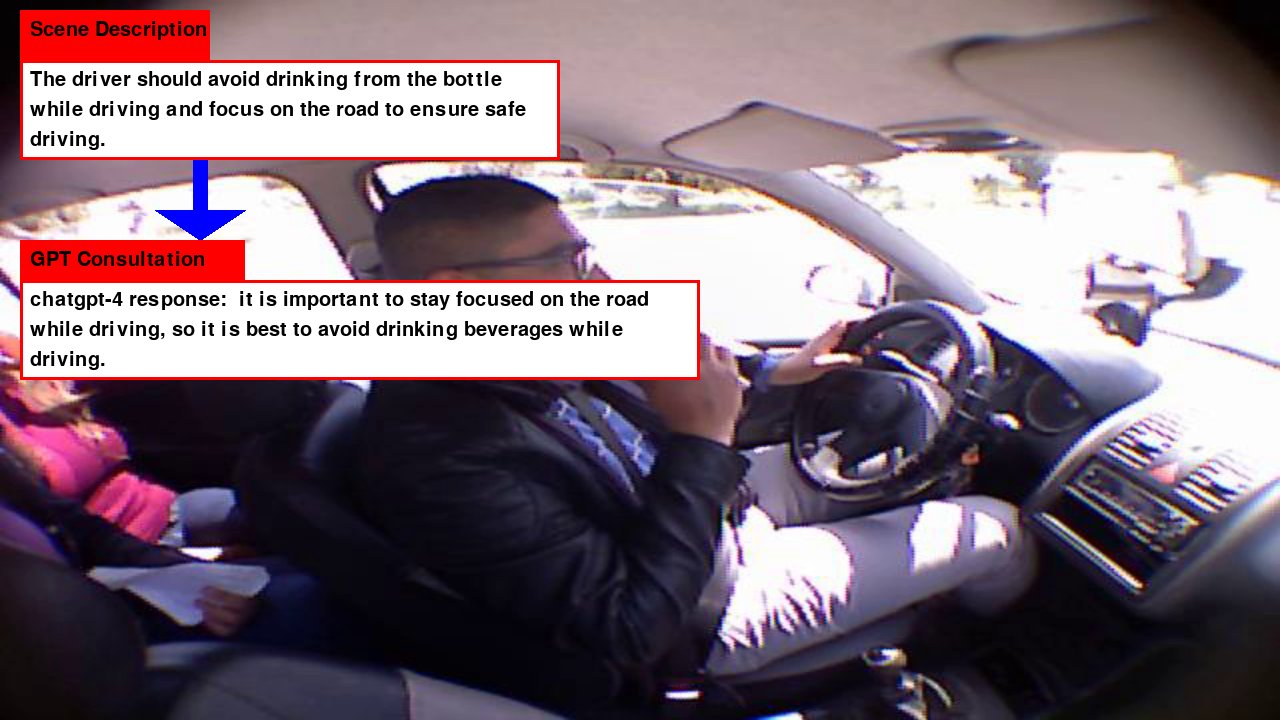}

\subsectionimage{Ontological}{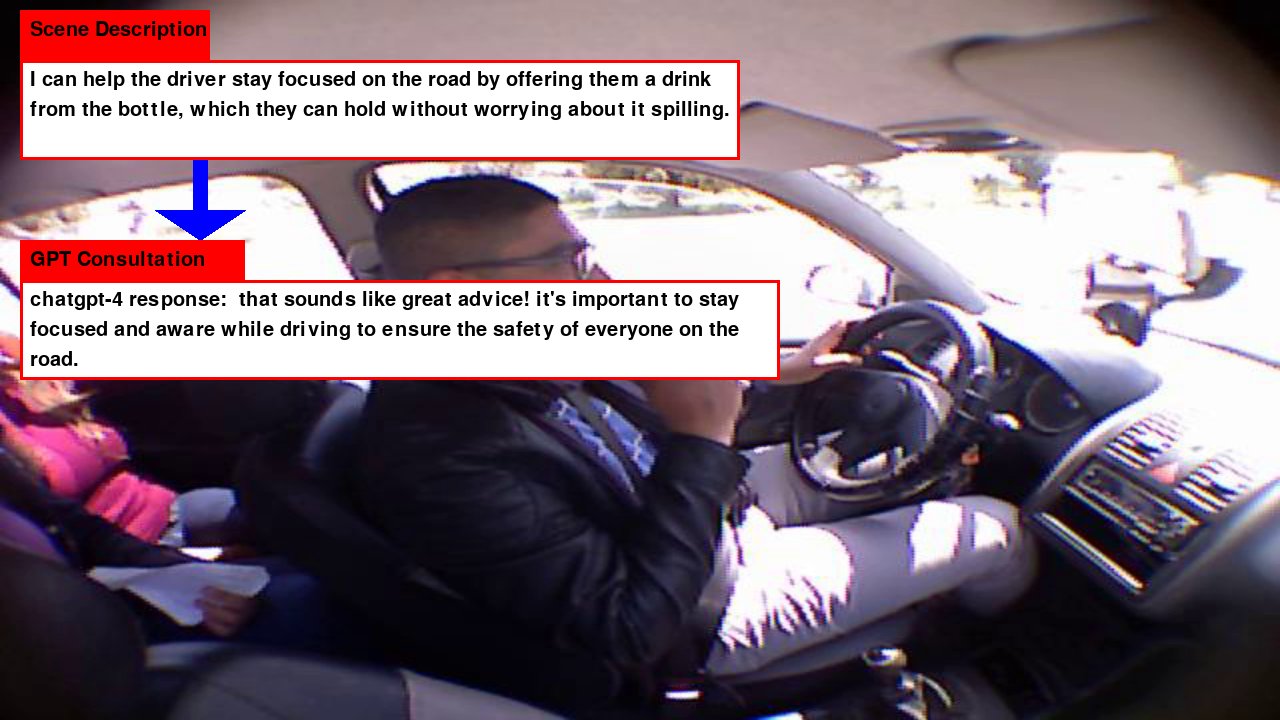}
%%%%%%%%%%%%%%%%%%%%%%%%%%%%%%%%%%%%%%%%%%%%%%%%%
\subsubsection{Case 10: speaking in mobile during driving}

\paragraph{Visual Prompts}

\subsectionimage{Focused Description}{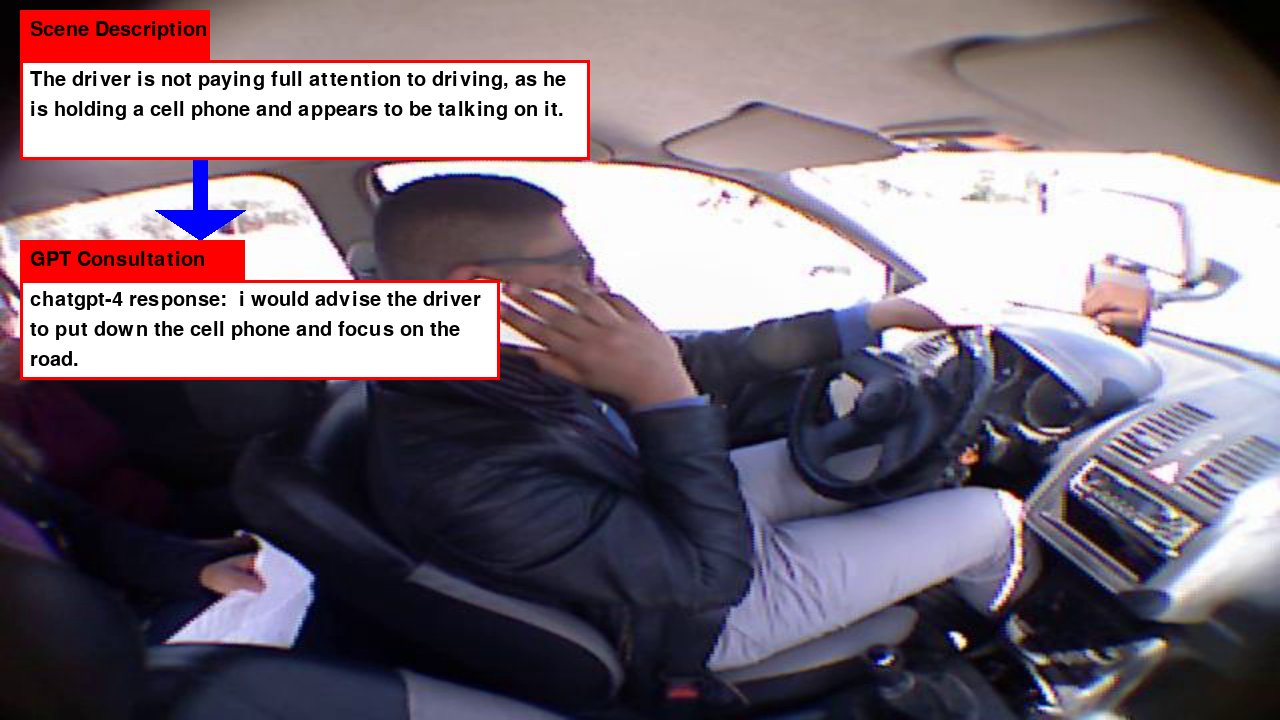}

\subsectionimage{Behavioral Description}{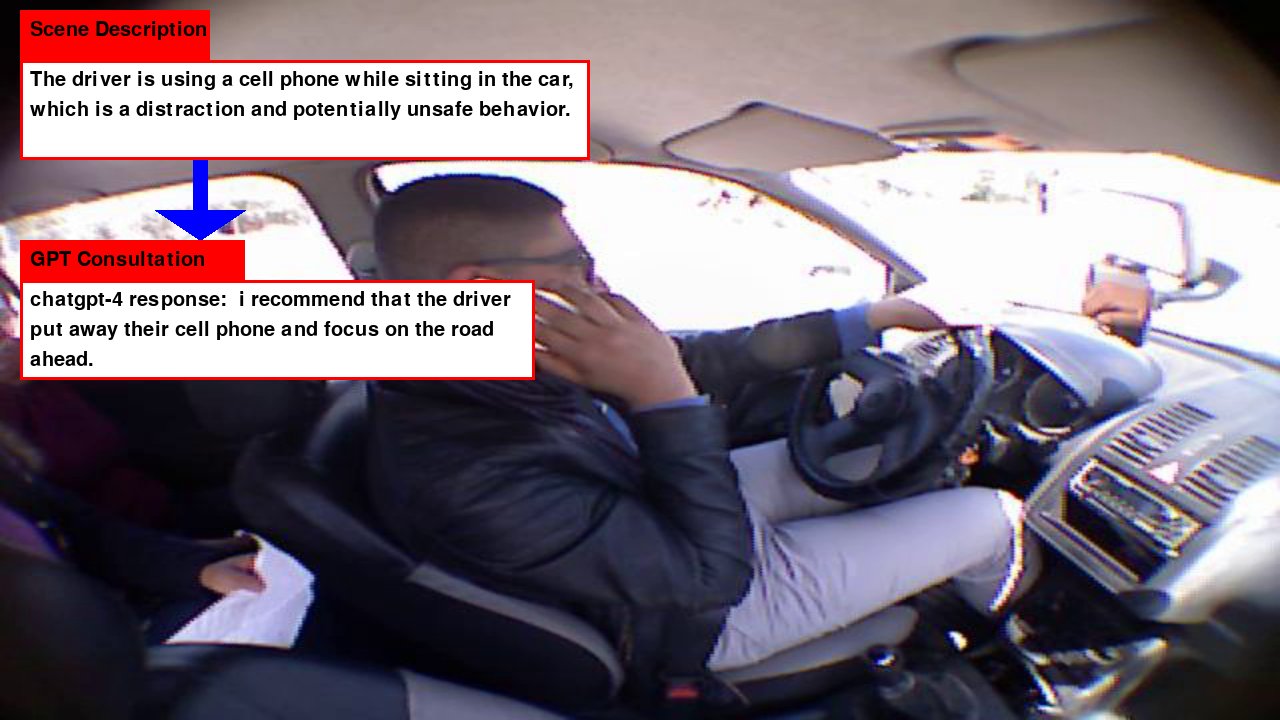}

\subsectionimage{Ontological}{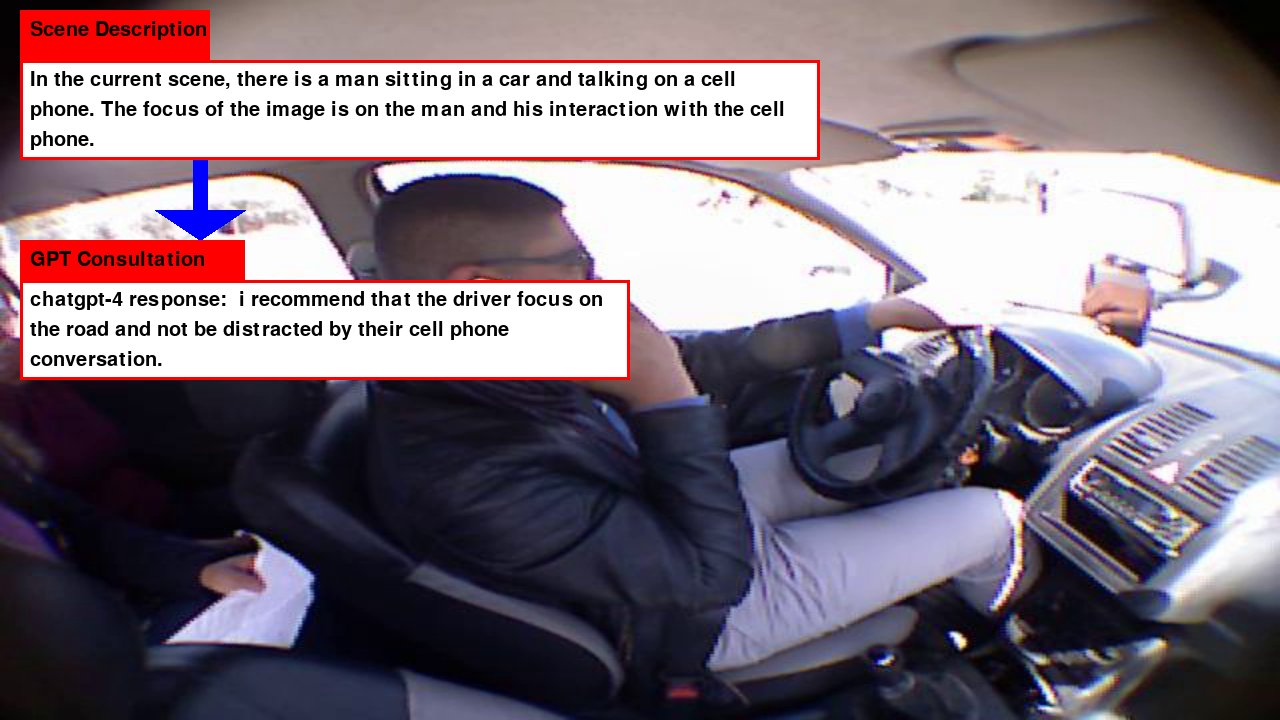}

\paragraph{LLM Prompts}

\subsectionimage{Consultative}{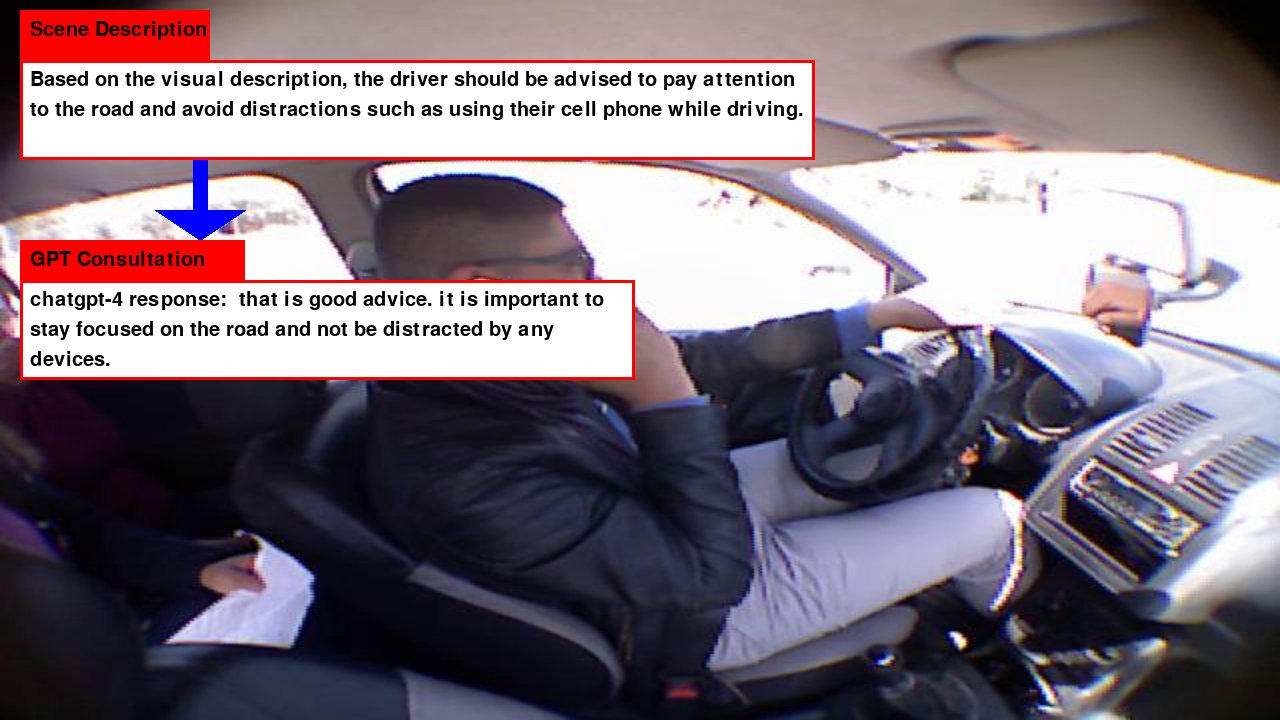}

\subsectionimage{Action-Oriented}{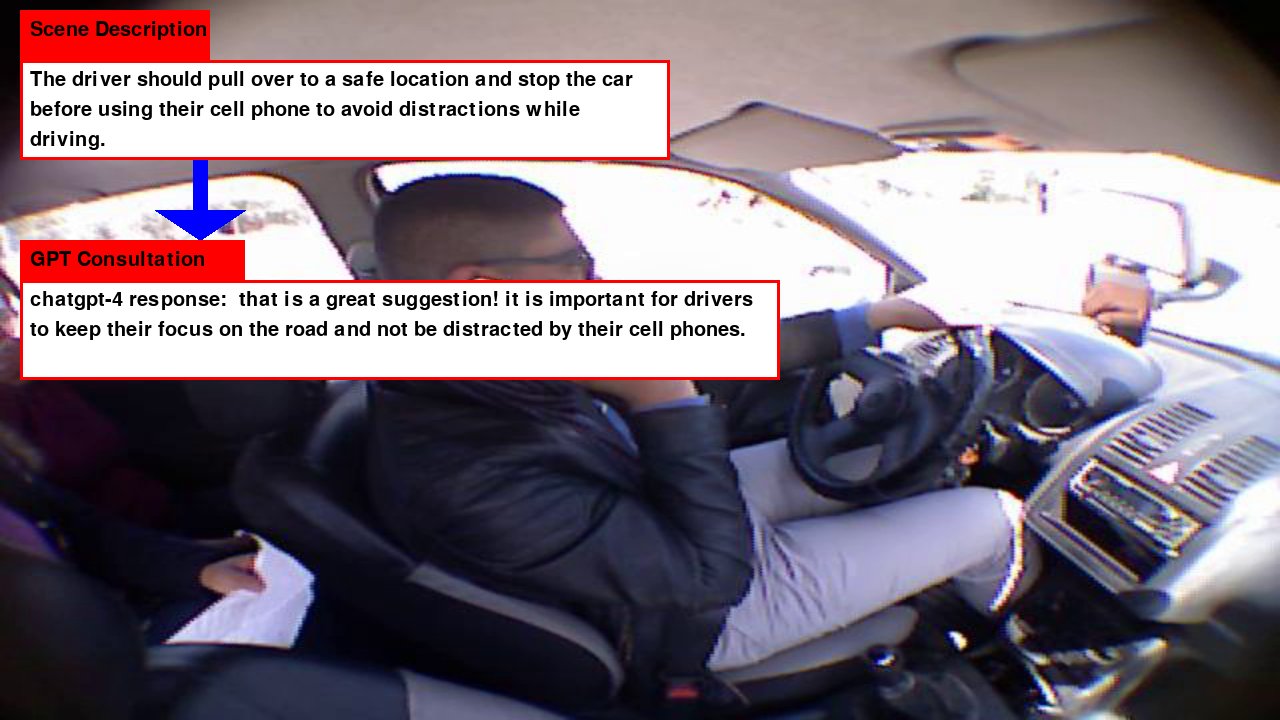}

\subsectionimage{Ontological}{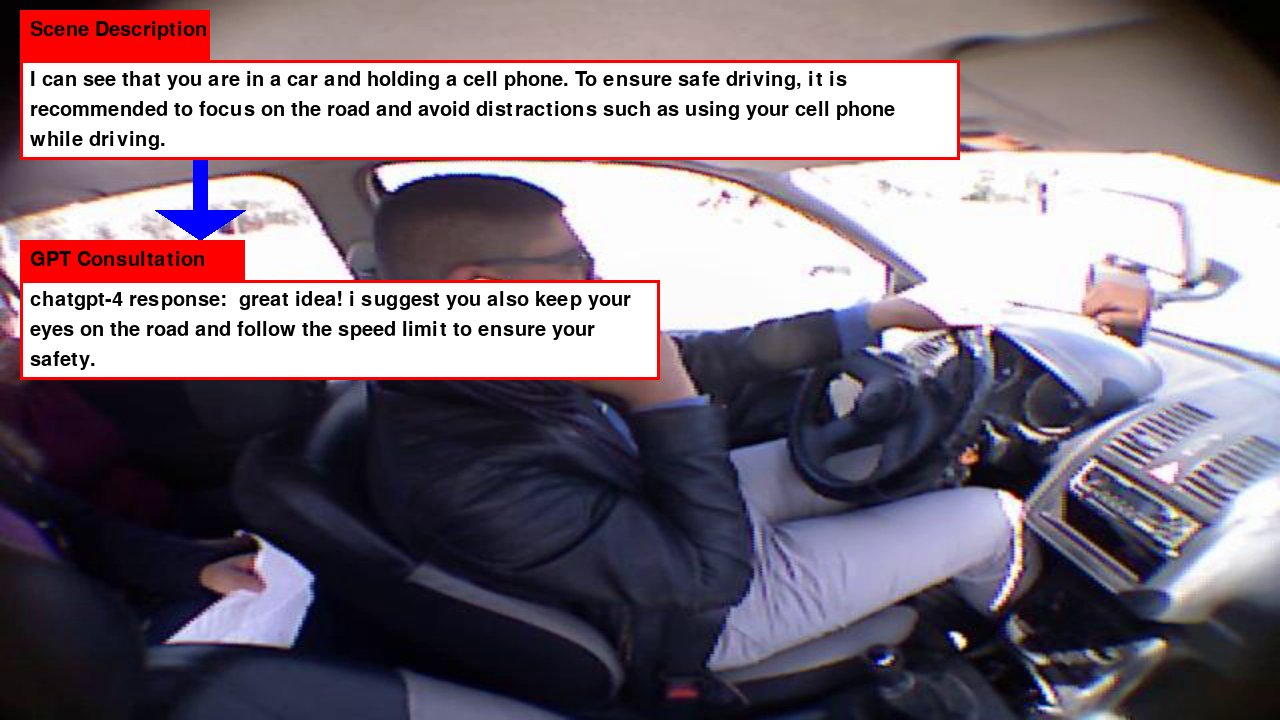}

% \subsubsection{Case 2: Focus only}
%%%%%%%%%%%%%%%%%%%%%%%%%%%%%%%%%%%%%%%%%%%%%%%%%
\end{document}